\documentclass[letterpaper, 10 pt, conference]{ieeeconf}  

\IEEEoverridecommandlockouts                              

\usepackage{times}
\usepackage{epsfig}
\usepackage{graphicx}
\usepackage{amsmath}
\usepackage{amssymb}
\usepackage{cite}
\usepackage{amsmath,amssymb,amsfonts}
\usepackage{textcomp}
\usepackage{subfigure}
\usepackage[linesnumbered,boxed,ruled,commentsnumbered]{algorithm2e}
\usepackage{algpseudocode}  
\usepackage{amsmath}  
\usepackage{multirow}
\usepackage{flushend}
\UseRawInputEncoding

\newtheorem{proposition}{Proposition}
\newtheorem{definition}{Definition}

\newtheorem{prooof}{Proof}
\title{\LARGE \bf
RANSIC: Fast and Highly Robust Estimation for Rotation Search and Point Cloud Registration using Invariant Compatibility
}

\author{Lei Sun$^{1}$
\thanks{*This work was not supported by any organization}
\thanks{$^{1}$Lei Sun is with School of Mechanical and Power Engineering, East China University of Science and Technology, Shanghai 200237, China,
        {\tt\small leisunjames@126.com}}%
}

\begin{document}

\maketitle
\thispagestyle{empty}
\pagestyle{empty}

\begin{abstract}

Correspondence-based rotation search and point cloud registration are two fundamental problems in robotics and computer vision. However, the presence of outliers, sometimes even occupying the great majority of the putative correspondences, can make many existing algorithms either fail or have very high computational cost. In this paper, we present RANSIC (RANdom Sampling with Invariant Compatibility), a fast and highly robust method applicable to both problems based on a new paradigm combining random sampling with invariance and compatibility. Generally, RANSIC starts with randomly selecting small subsets from the correspondence set, then seeks potential inliers as graph vertices from the random subsets through the compatibility tests of invariants established in each problem, and eventually returns the eligible inliers when there exists at least one $K$-degree vertex ($K$ is automatically updated depending on the problem) and the residual errors satisfy a certain termination condition at the same time. In multiple synthetic and real experiments, we demonstrate that RANSIC is fast for use, robust against over 95\% outliers, and also able to recall approximately 100\% inliers, outperforming other state-of-the-art solvers for both the rotation search and the point cloud registration problems.

\end{abstract}

\section{INTRODUCTION}

Rotation search and point cloud registration, broadly applied in 3D reconsturction~\cite{blais1995registering,henry2012rgb,choi2015robust}, object recognition and localization~\cite{drost2010model,papazov2012rigid,guo20143d}, SLAM systems~\cite{zhang2014loam} and attitude estimation~\cite{wahba1965least}, are two fundamental building blocks in robotics, computer vision and aerospace engineering.

Assume that we have two sets of vectors $\mathcal{A}=\{\mathbf{a}_i\}_{i=1}^{N}$ and $\mathcal{B}=\{\mathbf{b}_i\}_{i=1}^{N}$ both of size $N$, where $\mathbf{a}_i,\mathbf{b}_i\in \mathbb{R}^{3}$ constitute a putative vector correspondence. The goal of the rotation search problem is to estimate the best rotation $\boldsymbol{R}\in SO(3)$ aligning the vector sets $\mathcal{A}$ and $\mathcal{B}$. In the presence of isotropic zero-mean Gaussian noise, rotation search is equivalent to the following minimization problem:
\begin{equation}\label{Pro-1}
\underset{\boldsymbol{R}\in SO(3)}{\min} \sum_{i=1}^{N} ||\boldsymbol{R}\mathbf{a}_i-\mathbf{b}_i||,
\end{equation}
where $\boldsymbol{R}\mathbf{a}_i+\boldsymbol{\varepsilon}_i=\mathbf{b}_i$ ($i=1,2,\dots,N$ and $\boldsymbol{\varepsilon}_i$ denotes the noise measurement). Problem \eqref{Pro-1} can be solved by different closed-form methods~\cite{horn1987closed,arun1987least,horn1988closed,markley1988attitude}.

For the point cloud registration problem, given two point sets $\mathcal{P}=\{\mathbf{p}_i\}_{i=1}^{N}$ and $\mathcal{Q}=\{\mathbf{q}_i\}_{i=1}^{N}$ where $\mathbf{p}_i,\mathbf{q}_i\in \mathbb{R}^{3}$ make up a point correspondence, we aim to solve the scale $\mathit{s}>0$, rotation $\boldsymbol{R}\in SO(3)$ and translation $\boldsymbol{t}\in \mathbb{R}^{3}$ between $\mathcal{P}$ and $\mathcal{Q}$. Also, under the perturbance of isotropic zero-mean Gaussian noise, we obtain a minimization problem such that
\begin{equation}\label{Pro-2}
\underset{\boldsymbol{R}\in SO(3)}{\min} \sum_{i=1}^{N} ||\mathit{s}\boldsymbol{R}\mathbf{p}_i+\boldsymbol{t}-\mathbf{q}_i||,
\end{equation}
where $\mathit{s}\boldsymbol{R}\mathbf{p}_i+\boldsymbol{t}+\boldsymbol{\epsilon}_i=\mathbf{q}_i$ ($i=1,2,\dots,N$ and $\boldsymbol{\epsilon}_i$ is the noise measurement).

However, in many real-world applications, the great majority of the putative correspondences could be corrupted or substituted by the outliers (false correspondences), which are usually generated by mismatched 2D or 3D keypoints using feature detecting and matching approaches (e.g. SIFT~\cite{lowe2004distinctive} and SURF~\cite{bay2006surf} for 2D, and FPFH~\cite{rusu2008aligning} for 3D).

\begin{figure}[t]
\centering

\footnotesize{(a) Image Stitching with RANSIC}

\subfigure{
\begin{minipage}[t]{1\linewidth}
\centering
\includegraphics[width=0.385\linewidth]{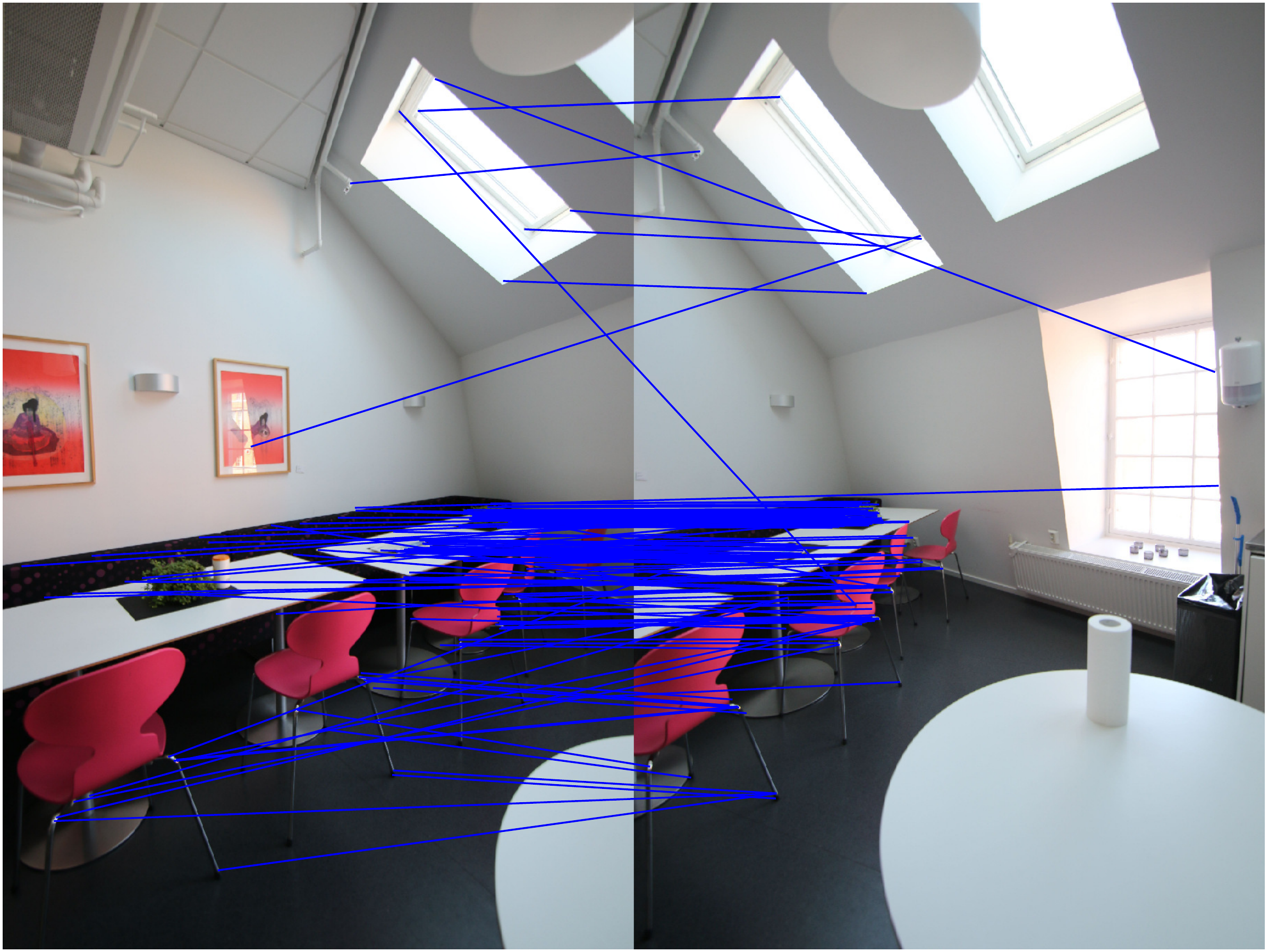}
\includegraphics[width=0.385\linewidth]{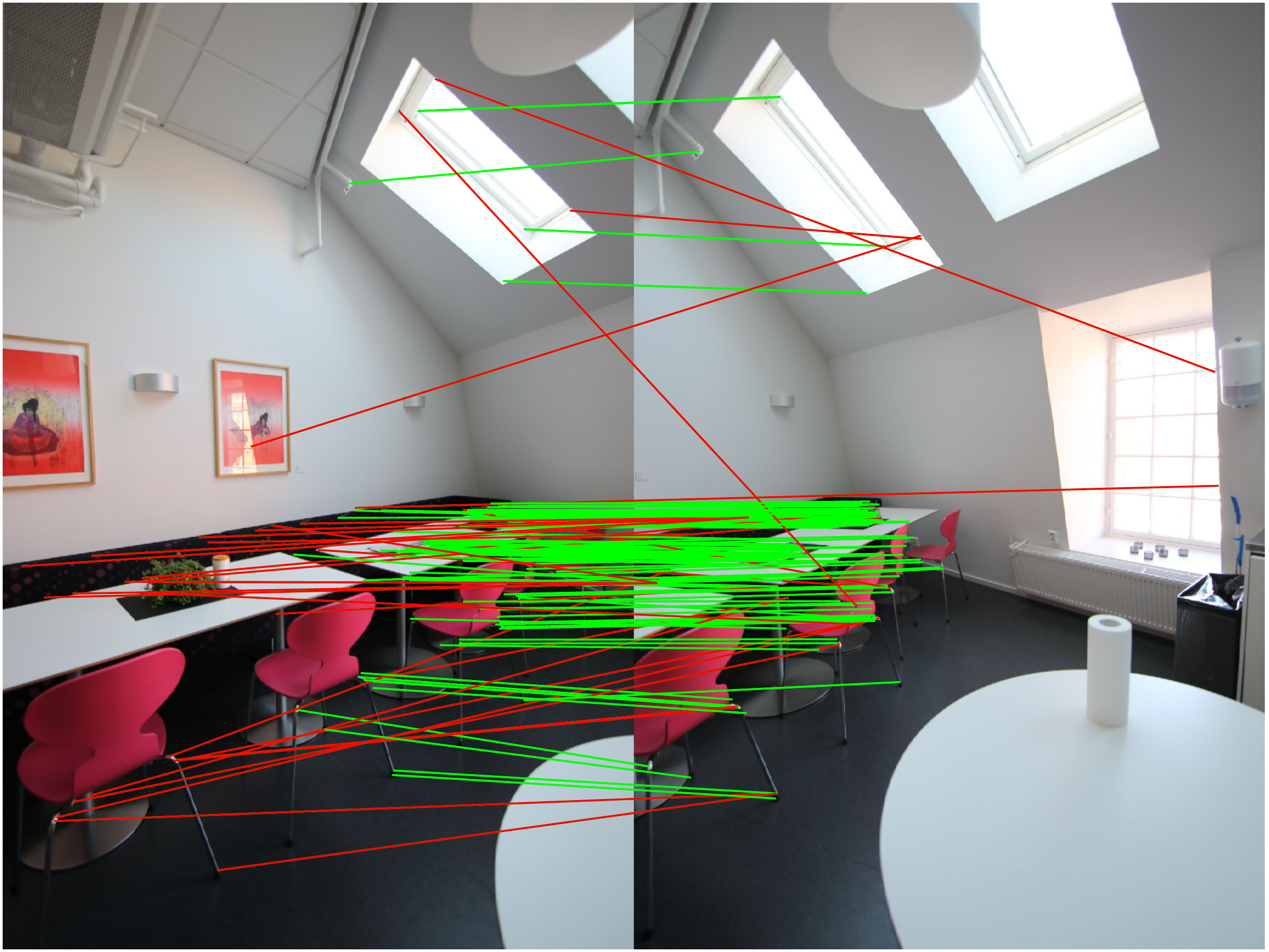}
\includegraphics[width=0.195\linewidth]{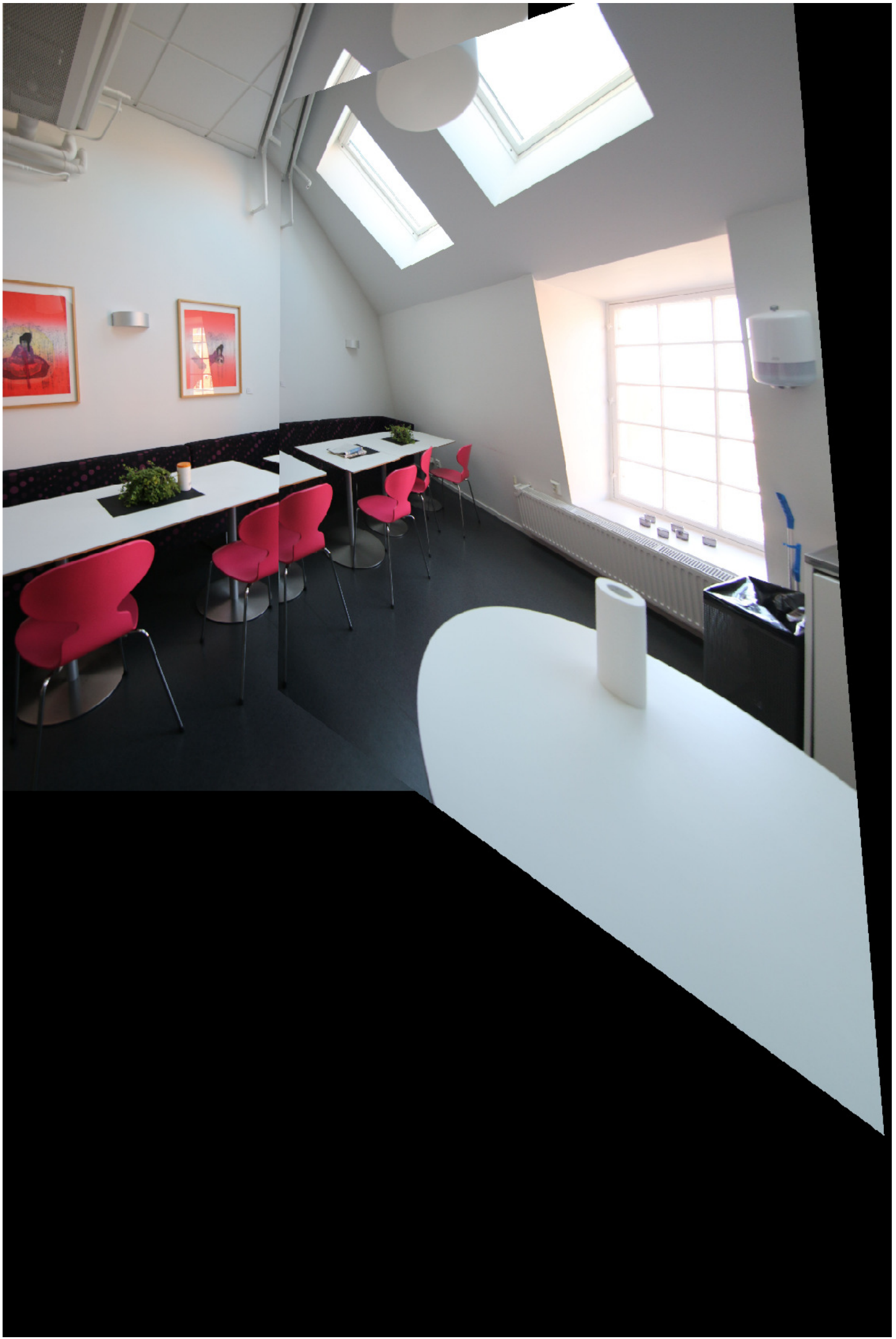}
\end{minipage}
}%

\footnotesize{(b) Unknown-scale Point Cloud Registration with RANSIC}

\subfigure{
\begin{minipage}[t]{1\linewidth}
\centering
\includegraphics[width=0.493\linewidth]{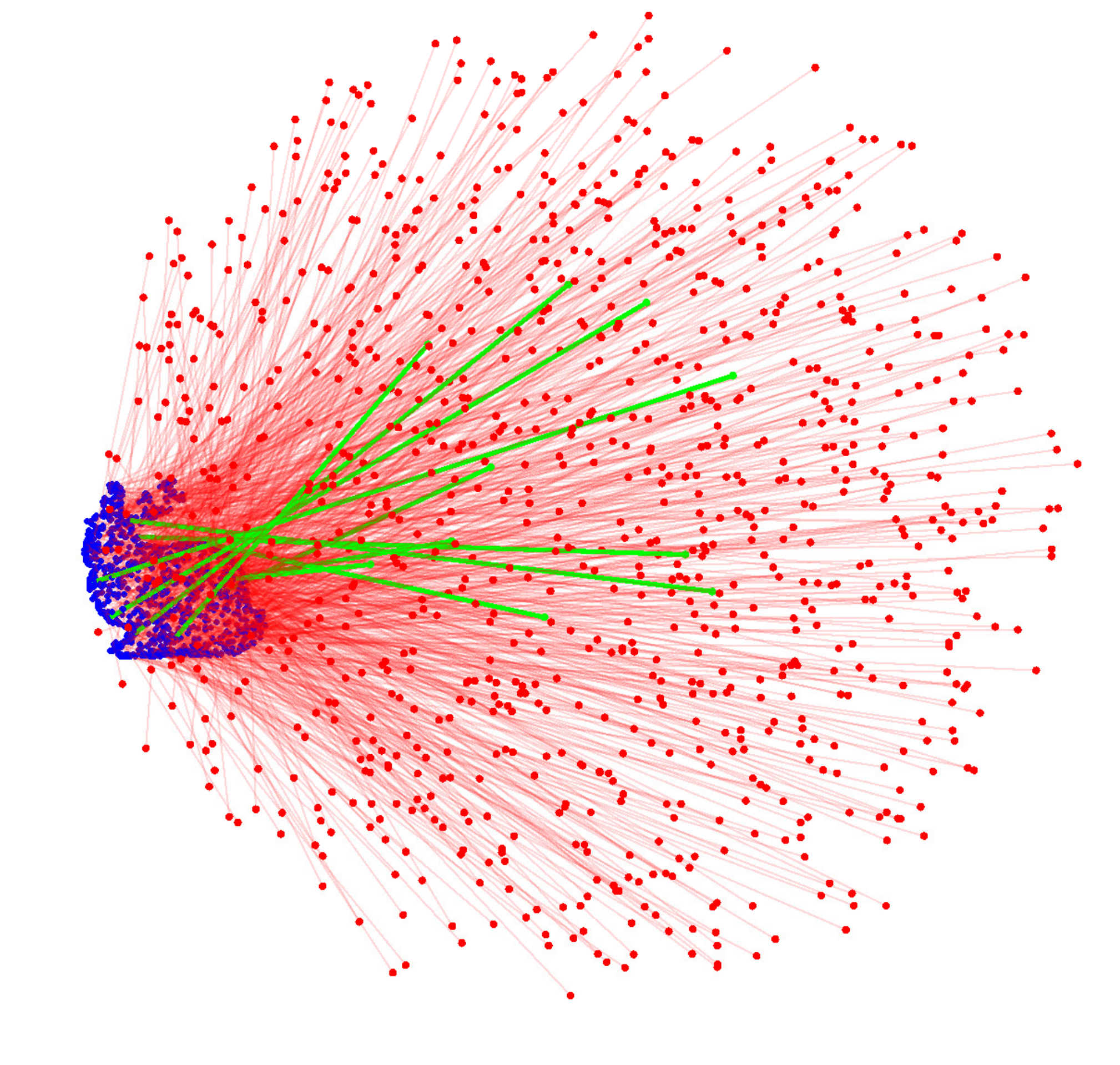}
\includegraphics[width=0.493\linewidth]{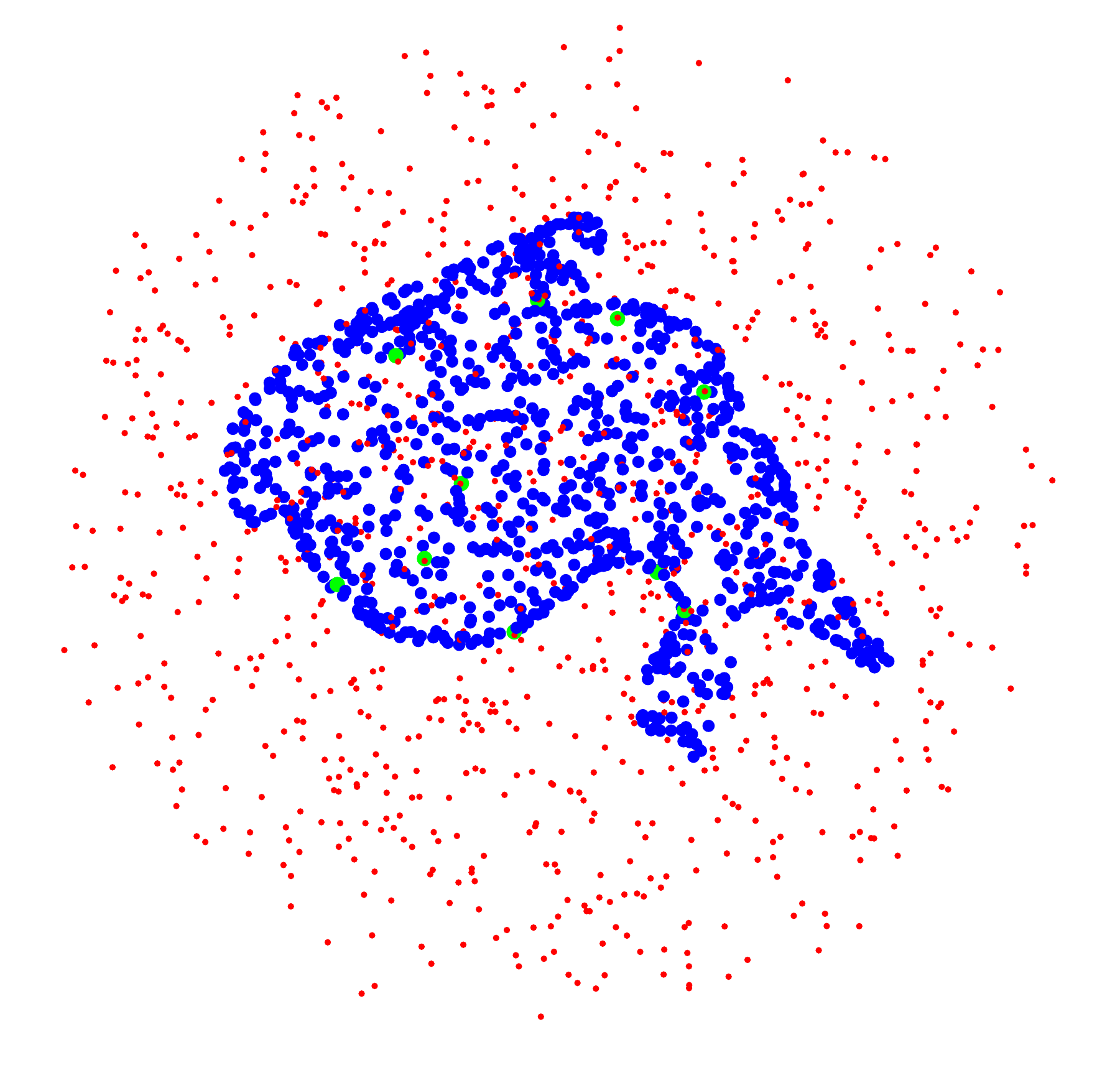}
\end{minipage}
}%

\vspace{-3mm}

\centering
\caption{(a) Image stitching using RANSIC. Left: Raw correspondences matched by SURF~\cite{bay2006surf}. Middle: Inliers sought by RANSIC (in green lines). Right: Image stitching result. (b) Unknown-scale point cloud registration using RANSIC. Left: Correspondences (inliers are in green lines and outliers are in red lines). Right: Registration result. We show that RANSIC can robustly estimate the scale, rotation and translation even when 990 of the 1000 correspondences are outliers.}
\label{demo}
\end{figure}

To deal with outliers, RANSAC~\cite{fischler1981random} and BnB (Branch-and-Bound~\cite{hartley2009global,olsson2008branch,parra2014fast}) are two popular consensus maximization robust solutions~\cite{chin2017maximum}. Unfortunately, RANSAC only works with minimal solvers and is too time-consuming once the outliers are in large numbers, while BnB usually requires significantly higher time cost with increasing correspondences. Moreover, though the more recent Graduated Non-Convexity (GNC) methods (e.g.\cite{yang2020graduated,antonante2020outlier,zhou2016fast}) can cooperate with non-minimal solvers to reject outliers without any initial guess, they become fragile when the outlier ratio is extreme (e.g. $\geq$90\%), which is common in real scenes~\cite{bustos2017guaranteed}.

Our contribution in this paper is to propose a novel robust estimation paradigm that incorporates the compatibility of invariants and graph theory into random sampling, leading to an invariant-based, fast and highly robust solver, named RANSIC (RANdom Sampling with Invariant Compatibility), for the correspondence-based rotation search and (unknown-scale or known-scale) point cloud registration problems.

Specifically, we construct a series of invariant functions customized for the two problems and explore the mutual compatibility between them to establish a sequence of stratified `filters' that can swiftly eliminate outliers layer by layer and preserve inliers simultaneously from the random sample flow. Through this framework, nearly all the inliers can be extracted from the putative correspondences, when the two conditions are both satisfied: (i) we have at least one $K$-vertex in the undirected graph built upon the compatibility tests, and (ii) the distribution of the residual errors reaches a certain regularity. The resulting solver RANSIC is more time-efficient than other tested state-of-the-art solvers in most cases, can tolerate at least 95\% outliers, and also has the highest estimation accuracy since it can recall almost 100\% inliers. We show the distinguished performance of RANSIC in various experiments over rotation search and point cloud registration as well as their typical real-data applications.

Our solver RANSIC is partially inspired by the invariant-based outlier pruner ROBIN~\cite{shi2020robin}, but they are rather different in the following ways: (i) ROBIN is mainly designed to boost the robustness of existing solvers (e.g. GNC), but not as an independent solver, whereas RANSIC itself can solve the two problems efficiently, (ii) ROBIN relies on the maximal clique or k-core solvers adopted, while RANSIC does not require any external algorithm, (iii) ROBIN cannot address rotation search and unknown-scale registration, while RANSIC can, and (iv) the methodologies are quite disparate: ROBIN uses invariant compatibility to prune outliers, while RANSIC uses it to seek inliers. This paper is also an improved, extended and generalized work of our previous preprint~\cite{sun2021iron}.

\section{Related Work}

\textbf{Consensus Maximization Methods.} A traditional way to tackle outliers is to maximize the correspondence set whose errors are below a certain bound. The hypothesize-and-test framework RANSAC~\cite{fischler1981random} is a popular robust heuristic~\cite{andrew2001multiple}, but its runtime increases exponentially with the outlier ratio. BnB\cite{hartley2009global,olsson2008branch,parra2014fast} is another robust heuristic that searches in the $SO(3)$ group of the rotation or the $SE(3)$ group of the rigid transformation to solve the optimization problem globally, but its runtime grows exponentially with the correspondence number. Therefore, both of them have serious limitations in practical applications.

\textbf{M-estimation Methods.} M-estimation methods seek to decrease the effect (weights) of the outliers by using robust cost functions. However, local M-estimation methods (e.g. \cite{kummerle2011g,sunderhauf2012towards}) highly depend on a promising initial guess, thus prone to fail in scenes lacking the initialization information. Graduated Non-Convexity (GNC) is a special M-estimation method that does not require any initial guess and optimizes surrogate functions instead. The GNC framework is found effective in multiple robotics problems~\cite{yang2020graduated}, including point cloud registration, pose graph optimization, shape alignment, etc, but its downside lies in its limited robustness ($\leq$85\%).

\textbf{Graph-theoretic Methods.} Graph theory has found wide applications in diverse robotics and computer vision problems, including human pose estimation~\cite{bray2006posecut}, feature matching~\cite{bailey2000data}, motion segmentation~\cite{perera2012maximal}, 3D-2D/3D registration~\cite{enqvist2009optimal}, etc. It can also be used for improving the traditional heuristics (e.g. RANSAC~\cite{barath2018graph}). Currently, graph theory and invariants have been applied to prune outliers (e.g. TEASER(++)~\cite{yang2019polynomial,yang2020teaser} and ROBIN~\cite{shi2020robin}) with the maximal clique~\cite{eppstein2010listing} or k-core~\cite{dasari2014park} solvers.

\begin{figure}[t]
\centering

\includegraphics[width=1\linewidth]{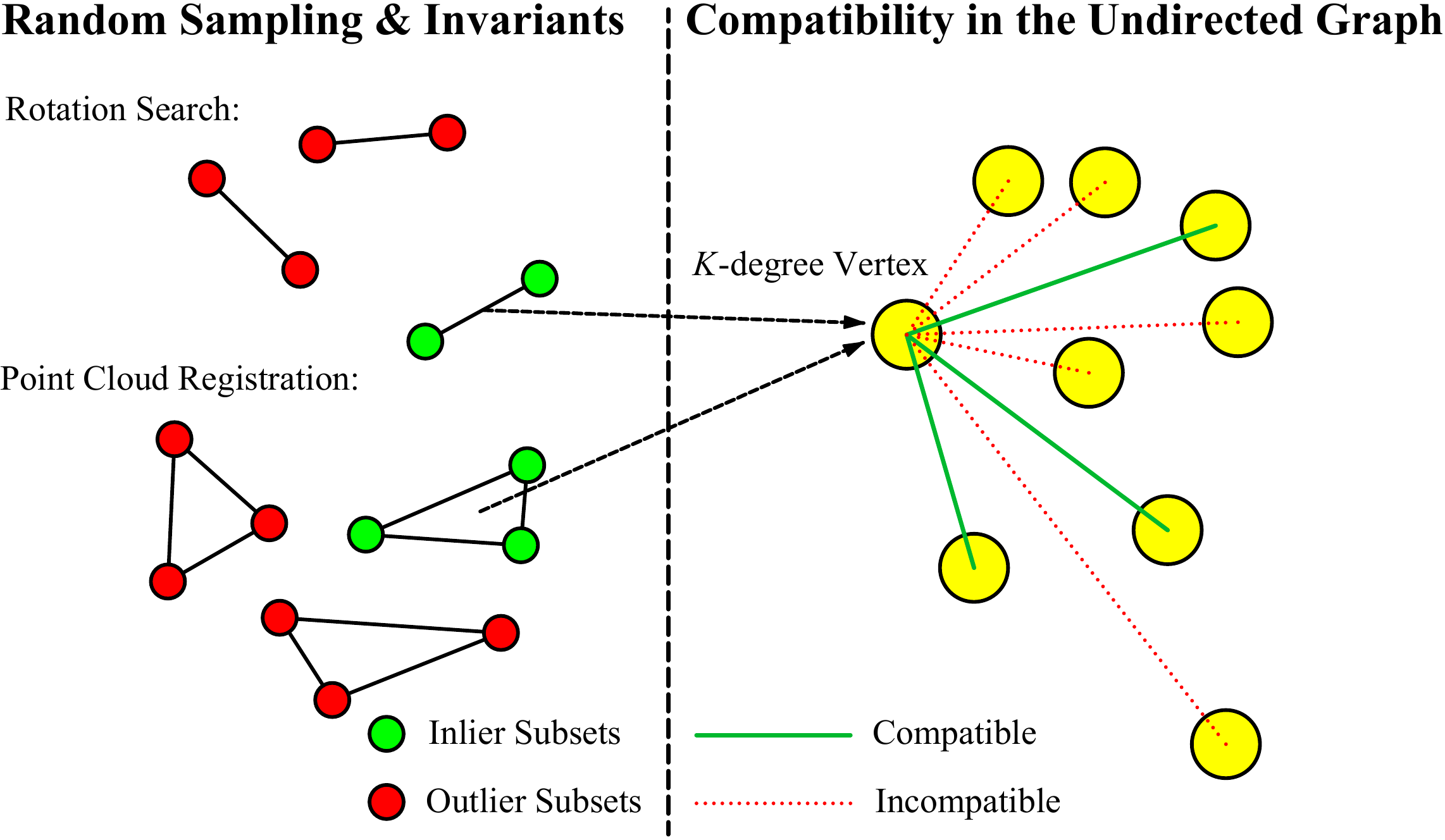}

\vspace{-2mm}

\centering
\caption{A graph-theoretic overview of RANSIC.}
\label{overview}
\end{figure}

\section{Definition and Notation}

To define the invariant functions in the two problems, we adopt the notation and definition in~\cite{shi2020robin}.

\begin{definition}[Invariant Function]
Assume that in one geometric problem, we have $N$ putative correspondences with indices in set $\mathcal{N}=\{1,2,\dots,N\}$. Let $\mathcal{S}\subset\mathcal{N}$ be a size-$k$ subset of $\mathcal{N}$ and $\boldsymbol{x}$ be the variable vector in this problem. Then let $\boldsymbol{y}_{\mathcal{S}}$ represent the measurements w.r.t. set $\mathcal{S}$, $\boldsymbol{\eta}_{\mathcal{S}}$ represent the corresponding noise measurements (independent of $\boldsymbol{x}$), and $\boldsymbol{h}_{\mathcal{S}}(\boldsymbol{x},\boldsymbol{\eta}_{\mathcal{S}})$ represent the corresponding model that relates $\boldsymbol{x}$ to the measurements $\boldsymbol{y}_{\mathcal{S}}$ such that $\boldsymbol{y}_{\mathcal{S}}=\boldsymbol{h}_{\mathcal{S}}(\boldsymbol{x},\boldsymbol{\eta}_{\mathcal{S}})$.

For a function $\boldsymbol{f}(\boldsymbol{y}_{\mathcal{S}})$ where $\boldsymbol{f}(\boldsymbol{y}_{\mathcal{S}})=\boldsymbol{f}\left(\boldsymbol{h}_{\mathcal{S}}(\boldsymbol{x},\boldsymbol{\eta}_{\mathcal{S}})\right)$ is always satisfied and independent of the variables $\boldsymbol{x}$ no matter which $\mathcal{S}\subset\mathcal{N}$ is chosen, we consider function $\boldsymbol{f}$ as a function invariant to $\boldsymbol{x}$, also called as a $k$-invariant function.

\end{definition}

\section{Our Method: RANSIC}

\subsection{Graph-Theoretic Overview of RANSIC}

We first provide a graph-theoretic overview on the general mechanism of our solver RANSIC, as displayed in Fig.~\ref{overview}. 

At the beginning, subsets of the universal correspondence set are randomly selected and subsequently examined by a series of compatibility tests (built upon the mathematical constraints of the problem). If a sample (subset) passes these tests (green circle group), we can take it as a \textit{potential inlier subset}. We then collect a minimum number of such subsets and consider them as the `vertices' in an undirected graph (yellow circles). Since then, during the sampling process, whenever we obtain a new potential inlier subset (as a new `vertex'), we generate a new undirected graph by checking the mutual compatibility between this vertex and every other existing vertex using another series of compatibility tests. If a pair of vertices are compatible with each other, we put an `edge' between them (green line); if not, no edge is placed (red dot line). In this way, if a new vertex has degree over $K$ ($K$ is a known number), the sampling is halted and this vertex and its compatible vertices are very likely to be inliers. 

Before RANSIC breaks from random sampling, we need to check the reliability of the estimated solution, say $\boldsymbol{x}^{\star}$.

\begin{proposition}[Termination Condition]\label{Prop-distrib}
Let $\mathcal{r}=\{r_{i}\}_{i=1}^N$ be the set of residual errors computed from the estimated $\boldsymbol{x}^{\star}$. If $\boldsymbol{x}^{\star}$ is globally optimal and the inlier set candidate contains no outlier, the following inequalities should be satisfied:
\begin{equation}\label{distribution}
\overline{\mathcal{r}}^{\star}\leq \upsilon \cdot \sigma,\, |\mathcal{r}^{\star}|\geq \tau,\, s.t.\,\mathcal{r}^{\star}=\{r_i|r_i\in\mathcal{r}, r_i<=5.2\sigma\},
\end{equation}
where $\overline{\mathcal{r}}^{\star}$ denotes the mean value of all elements in $\mathcal{r}^{\star}$, $\upsilon$ and $\tau$ are two preset number, and $\sigma$ is the standard deviation of Gaussian noise. Intuitively, this condition has two key points: (i) the mean of residual errors is below a certain level depending on noise, and (ii) a least possible number of inlier correspondences have been obtained. The explicit realization of RANSIC is rendered in the following subsections.

\subsection{Invariant Functions for Rotation Search}\label{RS}

Following Problem~\eqref{Pro-1}, our first step is to construct the invariant functions for the rotation search problem. Here we propose the length-based invariants:

\begin{proposition}[Length-based Invariant Function]\label{Prop-1}
If we randomly select two pairs of the vector correspondences $(\mathbf{a}_1,\mathbf{b}_1,\mathbf{a}_2,\mathbf{b}_2)$ from set $\mathcal{A}$ and $\mathcal{B}$, and then normalize them to unit length such that
\begin{equation}
\mathbf{a}^{*}_1=\frac{\mathbf{a}_1}{||\mathbf{a}_1||}, \,
\mathbf{b}^{*}_1=\frac{\mathbf{b}_1}{||\mathbf{b}_1||},\,
\mathbf{a}^{*}_2=\frac{\mathbf{a}_2}{||\mathbf{a}_2||}, \,
\mathbf{b}^{*}_2=\frac{\mathbf{b}_2}{||\mathbf{b}_2||},
\end{equation}
we can define a length-based $2$-invariant function as: 
\begin{equation}
\boldsymbol{f}(\mathbf{a}_1,\mathbf{b}_1,\mathbf{a}_2,\mathbf{b}_2)=\left| \,||\mathbf{b}^{*}_1-\mathbf{b}^{*}_2||-||\mathbf{a}^{*}_1-\mathbf{a}^{*}_2||\, \right|, 
\end{equation}
which is invariant to rotation $\boldsymbol{R}$. In addition, we can derive a compatibility condition on function $\boldsymbol{f}(\mathbf{a}_1,\mathbf{b}_1,\mathbf{a}_2,\mathbf{b}_2)$ as:
\begin{equation}\label{length-comp}
\boldsymbol{f}(\mathbf{a}_1,\mathbf{b}_1,\mathbf{a}_2,\mathbf{b}_2)\leq X,
\end{equation}
where $X$ is a finite constant independent of $\boldsymbol{R}$.
\end{proposition}
\begin{prooof}
Based on triangular inequality and the properties of $\boldsymbol{R}$, we can always have:
\begin{equation}
\begin{gathered}
\boldsymbol{f}(\mathbf{a}_1,\mathbf{b}_1,\mathbf{a}_2,\mathbf{b}_2)= \left| \,||\mathbf{b}^{*}_1-\mathbf{b}^{*}_2||-||\mathbf{a}^{*}_1-\mathbf{a}^{*}_2||\, \right| \\
=\left| \|\boldsymbol{R}\| \cdot \left\| \frac{\mathbf{a}_1+\boldsymbol{R}^{\top}\boldsymbol{\varepsilon}_1}{||\mathbf{b}_1||}-\frac{\mathbf{a}_2+\boldsymbol{R}^{\top}\boldsymbol{\varepsilon}_2}{||\mathbf{b}_2||}\right\| -\left\| \frac{\mathbf{a}_1}{\|\mathbf{a}_1\|}-\frac{\mathbf{a}_2}{\|\mathbf{a}_2\|}\right\|  \right|,\\
\leq \, X^{*} +\|\boldsymbol{R}^{\top} \|  \cdot\left\| \frac{\boldsymbol{\varepsilon}_1}{||\mathbf{b}_1||}- \frac{\boldsymbol{\varepsilon}_2}{||\mathbf{b}_2||} \right\| \leq X^{*}+\zeta=X,
\end{gathered}
\end{equation}
where scalar $X^{*} = \left\| \frac{\mathbf{a}_1}{||\mathbf{b}_1||}-\frac{\mathbf{a}_2}{||\mathbf{b}_2||}-\frac{\mathbf{a}_1}{||\mathbf{a}_1||}+\frac{\mathbf{a}_2}{||\mathbf{a}_2||}\right\| $ and $\zeta$ is a constant satisfying: $\left\| \frac{\boldsymbol{\varepsilon}_1}{||\mathbf{b}_1||}-\frac{\boldsymbol{\varepsilon}_2}{||\mathbf{b}_2||} \right\|\leq ||\frac{\boldsymbol{\varepsilon}_1}{\mathbf{b}_1}||+||\frac{\boldsymbol{\varepsilon}_2}{\mathbf{b}_2}||\leq \zeta$.
\end{prooof}

Proposition~\ref{Prop-1} is the first compatibility test that serves to filter out the outliers and preserve the vector pairs that are likely to be inliers, called \textit{potential vector pairs}. Then we compute the raw rotation matrix for each potential vector pair, where we adopt the fast non-minimal SVD (Singular Value Decomposition) rotation estimator in \cite{arun1987least}.

As for any two potential vector pairs and their estimated rotations, say $\boldsymbol{R}^*_1$ and $\boldsymbol{R}^*_2$, we employ another compatibility test to determine if they are likely to be both true inliers.

Assume that now we have $H$ potential vector pairs {$\mathcal{H}=\{\{(\mathbf{a}_i, \mathbf{b}_i)_c\}_{i=1}^2\}_{c=1}^H$} in total. We are able to define a new rotation-based invariant function.

\begin{proposition}[Rotation-based Invariant Function]\label{Prop-2}
If we select two rotations $\boldsymbol{R}^*_1$ and $\boldsymbol{R}^*_2$ w.r.t. any two potential vector pairs, we can define a $4$-invariant function $\boldsymbol{E}(\boldsymbol{R}^*_1, \boldsymbol{R}^*_2)$ with its compatibility condition such that
\begin{equation}\label{E-comp}
\begin{gathered}
\boldsymbol{E}(\boldsymbol{R}^*_1, \boldsymbol{R}^*_2)=\angle(\boldsymbol{R}^*_1, \boldsymbol{R}^*_2)\\
=\angle\left({Exp}(\boldsymbol{\pi}^{\boldsymbol{R}^*_1}), {Exp}(\boldsymbol{\pi}^{\boldsymbol{R}^*_2})\right) \leq \theta,
\end{gathered}
\end{equation}
where $\boldsymbol{E}(\boldsymbol{R}^*_1, \boldsymbol{R}^*_2)$ is invariant to $\boldsymbol{R}$, $\angle$ denotes the geodesic distance (error) between the two rotations~\cite{hartley2013rotation}, and ${Exp}$ denotes the exponential map \cite{barfoot2017state} that describes the noise perturbance $\boldsymbol{\pi}^{\boldsymbol{R}^*}$ on the rotation. 
\end{proposition}
\begin{prooof}
According to the properties of the geodesic distance in~\cite{hartley2013rotation}, and similar to~\cite{shi2020robin}, we can derive that
\begin{equation}\label{geo-error}
\begin{gathered}
\angle(\boldsymbol{R}^*_1, \boldsymbol{R}^*_2)=\angle\left(\boldsymbol{{R}}{Exp}(\boldsymbol{\pi}^{\boldsymbol{R}^*_1}), \boldsymbol{{R}} {Exp}(\boldsymbol{\pi}^{\boldsymbol{R}^*_2})\right)\\
=\left| arccos\left(\frac{tr({Exp}(\boldsymbol{\pi}^{\boldsymbol{R}^*_1})^{\top}\boldsymbol{R}^{\top}\boldsymbol{R}{Exp}(\boldsymbol{\pi}^{\boldsymbol{R}^*_2}))-1}{2}\right)\right|\\
=\angle \left({Exp}(\boldsymbol{\pi}^{\boldsymbol{R}^*_1}), {Exp}(\boldsymbol{\pi}^{\boldsymbol{R}^*_2}) \right) \\ 
\leq \angle \left({Exp}(\boldsymbol{\pi}^{\boldsymbol{R}^*_1}), \mathbf{I}_3 \right) + 
\angle \left({Exp}(\boldsymbol{\pi}^{\boldsymbol{R}^*_2}), \mathbf{I}_3 \right) \leq 2\theta^{*}=\theta,
\end{gathered}
\end{equation}
where $\mathbf{I}_3\in\mathbb{R}^{3\times3}$ is the identity matrix. As the noise measurements $\boldsymbol{\pi}^{\boldsymbol{R}^*_1}$ and $\boldsymbol{\pi}^{\boldsymbol{R}^*_2}$ here are always bounded (say $\angle \left({Exp}(\boldsymbol{\pi}^{\boldsymbol{R}^*}), \mathbf{I}_3 \right) \leq \theta^{*}$), the inequality regarding the geodesic distance~\eqref{geo-error} can be always satisfied.
\end{prooof}

\end{proposition}

\subsection{RANSIC for Rotation Search}

In Algorithm~\ref{Algo1-RS}, we provide the pseudocode of RANSIC for the rotation search problem.

\textbf{Description of Algorithm~\ref{Algo1-RS}.} We set an empty set $\mathcal{X}$ to contain potential inliers and $K=1$ (\textit{line 1}). We take random samples and check their invariant compatibility \eqref{length-comp}, and if a sample (vector pair) $\mathcal{S}$ passes, we add it to set $\mathcal{X}$ (\textit{lines 3-5}). Then we consider all the vector pairs in $\mathcal{X}$ as the vertices of an undirected graph, and put all the vertices that are in the compatibility \eqref{E-comp} with the new vertex ${V}_{\mathcal{S}}$ into set $\mathcal{Y}$ (a subset of $\mathcal{X}$), including ${V}_{\mathcal{S}}$ itself (\textit{lines 5-6}). If the degree of the new vertex is no smaller than $K$, we solve the rotation with $\mathcal{Y}$ and compute the residual errors for all correspondences in $\mathcal{C}$ (\textit{lines 7-9}). If the residual errors satisfy condition~\eqref{distribution}, we stop sampling and consider $\mathcal{Y}$ as the true inliers; if not, we increase $K$ by 1 (\textit{lines 10-13}). Finally, we can find all the inliers by checking the residual errors (only keeping correspondences with residual error no larger than $5.2\sigma$) to form the ultimate inlier set $\mathcal{C}^{\star}$, and then solve the optimal rotation $\boldsymbol{R}^{\star}$ with the inliers in $\mathcal{C}^{\star}$ (\textit{lines 17-18}).

\begin{algorithm}[t]
\caption{RANSIC for Rotation Search}
\label{Algo1-RS}
\SetKwInOut{Input}{\textbf{Input}}
\SetKwInOut{Output}{\textbf{Output}}
\Input{correspondences $\mathcal{C}=\{(\mathbf{p}_i,\mathbf{q}_i)\}_{i=1}^N$; \\ initial degree $K$ to break sampling\;}
\Output{optimal rotation $\boldsymbol{R}^{\star}$; inlier set $\mathcal{C}^{\star}$ \;}
\BlankLine
Set empty sets $\mathcal{X}=\emptyset$, $\mathcal{C}^{\star}=\emptyset$, and $K=1$\;
\While {true}{
{Select a random subset $\mathcal{S}=\{(\mathbf{p}_j,\mathbf{q}_j)\}_{j=1}^2\subset\mathcal{C}$}\;
\If {set $\mathcal{S}$ can pass test \eqref{length-comp}}{
$\mathcal{X}=\mathcal{X}\cup\{\mathcal{S}\}$ as the vertices\;
Use test \eqref{E-comp} to find all the vertices $\mathcal{Y}\subset\mathcal{X}$ that have edges meeting vertex ${V}_{\mathcal{S}}$\;
\If {vertex ${V}_{\mathcal{S}}$ has degree at least $K$}{
Solve $\boldsymbol{R}$ with the correspondences in $\mathcal{Y}$\;
Compute the residuals $\mathcal{r}=\{r_i\}_{i=1}^N$ with $\boldsymbol{R}$ for all the correspondences in $\mathcal{C}$\;
\If {$\mathcal{r}$ fulfills condition \eqref{distribution}}{
\textbf{break}
}
$K=K+1$\;
}
}
}
Add all correspondences whose $r_i\leq5.2\sigma$ to set $\mathcal{C}^{\star}$\;
Solve $\boldsymbol{R}^{\star}$ with the inlier set $\mathcal{C}^{\star}$\;
\Return $\boldsymbol{R}^{\star}$ and $\mathcal{C}^{\star}$\;
\end{algorithm}

\subsection{Invariant Functions for Point Cloud Registration}\label{PCR}

In terms of Problem \ref{Pro-2}, we adopt both 3-invariant and $6$-invariant functions over different types of variables.
Assume that we have 3 random non-colinear point correspondences {$\{(\mathbf{p}_i, \mathbf{q}_i)\}_{i=1}^3$} $(i=1,2,3)$, here defined as a \textit{3-point set}. The centroids w.r.t. $\{\mathbf{p}_i\}_{i=1}^3$ and $\{\mathbf{q}_i\}_{i=1}^3$ can be computed as
\begin{equation}\label{RANSIC-1}
\mathbf{\bar{p}}=\frac{1}{3} \sum^{3}_{i=1}\mathbf{p}_i, \,\,\,\mathbf{\bar{q}}=\frac{1}{3}\sum^{3}_{i=1}\mathbf{q}_i.
\end{equation}
Then we define two sets of vectors directed from the centroid to the 3D points, which can be written as
\begin{equation}\label{RANSIC-2}
\mathbf{\widetilde{p}}_i=\mathbf{p}_i-\mathbf{\bar{p}}, \,\,\,\mathbf{\widetilde{q}}_i=\mathbf{q}_i-\mathbf{\bar{q}}.
\end{equation}
Here, we introduce a translation-free technique (first appeared in~\cite{arun1987least}) based on the centroids \eqref{RANSIC-1} such that
\begin{equation}\label{RANSIC-3}
\boldsymbol{t}=\mathbf{\bar{q}}-\mathit{s}\boldsymbol{R}\mathbf{\bar{p}},
\end{equation}
where since noise $\boldsymbol{\epsilon}_i$ is isotropic, we can approximately omit the noise measurement in this formulation.
\begin{proposition}[Scale-based Invariant Function] \label{Prop2} We can define a set of 3-invariant functions $\{{s}_i(\mathbf{p}_i, \mathbf{q}_i)\}^{3}_{i=1}$ such that
\begin{equation}
{s}_i(\mathbf{p}_i, \mathbf{q}_i)=\frac{||\mathbf{\widetilde{q}}_i||}{||\mathbf{\widetilde{p}}_i||}=\mathit{s}+\epsilon_i^{\mathit{s}},
\end{equation}
where each ${s}_i(\mathbf{p}_i, \mathbf{q}_i)$ is invariant to $\boldsymbol{R}$ and $\boldsymbol{t}$, and $\epsilon_i^{\mathit{s}}$ is the noise error describing the difference between the computed scale $\mathit{s}_i$ and the ground-truth scale $\mathit{s}$. Moreover, functions $\{{s}_i(\mathbf{p}_i, \mathbf{q}_i)\}^{3}_{i=1}$ have the following compatibility condition:
\begin{equation}\label{s-comp}
|{s}_i(\mathbf{p}_i, \mathbf{q}_i)-{s}_j(\mathbf{p}_j, \mathbf{q}_j)|\leq \alpha \left(\frac{1}{||\mathbf{\widetilde{p}}_i||}+\frac{1}{||\mathbf{\widetilde{p}}_j||}\right),
\end{equation}
where $(i,j)\in\{(1,2),(1,3),(2,3)\}$ and noise $||\boldsymbol{\epsilon}|| \leq\alpha$. For known $\mathit{s}=1$, we add the condition: $|\mathit{s}_i(\mathbf{p}_i, \mathbf{q}_i)-1|\leq \frac{\alpha}{||\mathbf{\widetilde{p}}_i||}$.
\end{proposition}
\begin{prooof}
According to Problem~\ref{Pro-2} and~\eqref{RANSIC-3}, we can obtain
\begin{equation}
\mathit{s}\boldsymbol{R}\mathbf{\widetilde{p}}_i+\boldsymbol{\epsilon}_i=\mathbf{\widetilde{q}}_i, \Rightarrow \, ||\mathit{s}\boldsymbol{R}\mathbf{\widetilde{p}}_i||=||\mathbf{\widetilde{q}}_i-\boldsymbol{\epsilon}_i||.
\end{equation}
Using the triangle inequality, we can derive that
\begin{equation}
\begin{gathered}
||\mathbf{\widetilde{q}}_i||-||\boldsymbol{\epsilon}_i|| \leq ||\mathit{s}\boldsymbol{R}\mathbf{\widetilde{p}}_i||= \mathit{s}||\mathbf{\widetilde{p}}_i|| \leq ||\mathbf{\widetilde{q}}_i||+||\boldsymbol{\epsilon}_i||,
\end{gathered}
\end{equation}
and if we let $\alpha$ be the noise bound $(||\boldsymbol{\epsilon}_i|| \leq\alpha)$, we have
\begin{equation}
\begin{gathered}
\frac{||\mathbf{\widetilde{q}}_i||-\boldsymbol{\epsilon}_i}{||\mathbf{\widetilde{p}}_i||} \leq \mathit{s} \leq \frac{||\mathbf{\widetilde{q}}_i||+\boldsymbol{\epsilon}_i}{||\mathbf{\widetilde{p}}_i||},
\end{gathered}
\end{equation}
which leads to the compatibility condition in~\eqref{s-comp}.
\end{prooof}

Proposition~\ref{Prop2} indicates that as long as the correspondences in the 3-point set are inliers, they should satisfy inequalities \eqref{s-comp}, which makes up our first compatibility condition.

Then we roughly compute the raw scale and rotation with the 3-point set. We use the least-squares solver~\cite{yang2019polynomial} for scale:
\begin{equation}\label{RANSIC-9}
\mathit{s}^{*}=\frac{1}{\sum_{i=1}^3 w_i}\cdot \sum_{i=1}^3 w_i\boldsymbol{s}_i(\mathbf{p}_i, \mathbf{q}_i),
\end{equation}
where $w_i=\frac{||\mathbf{\widetilde{p}}_i||^2}{\alpha^2}$.
To compute rotation, we sill use SVD~\cite{arun1987least} as in rotation search. With the computed $\boldsymbol{R}^{*}$ and $\mathit{s}^{*}$, we can derive our second compatibility condition as follows.

\begin{proposition}[Translation-based Invariant Function] \label{Prop3} We then define a set of functions $\{\boldsymbol{t}_i(\mathbf{p}_i, \mathbf{q}_i)\}^{3}_{i=1}$ satisfying:
\begin{equation}
\boldsymbol{t}_i(\mathbf{p}_i, \mathbf{q}_i)=\mathbf{q}_i-\mathit{{s}}^{*}\boldsymbol{{R}}^{*}\mathbf{p}_i=\boldsymbol{t}+\boldsymbol{\epsilon}^{\boldsymbol{t}}_i,
\end{equation}
where $\boldsymbol{t}_i(\mathbf{p}_i, \mathbf{q}_i)$ is invariant to $\mathit{s}$ and $\boldsymbol{R}$, and $\boldsymbol{\epsilon}^{\boldsymbol{t}}_i$ denotes the noise measurement. Any two $\boldsymbol{t}_i(\mathbf{p}_i, \mathbf{q}_i)$ should satisfy the compatibility condition:
\begin{equation}\label{t-comp}
||\boldsymbol{t}_i(\mathbf{P}_i, \mathbf{Q}_i)-\boldsymbol{t}_j(\mathbf{P}_j, \mathbf{Q}_j)||\leq 2\beta,
\end{equation}
where $(i,j)\in\{(1,2),(1,3),(2,3)\}$ and $||\boldsymbol{\epsilon}^{\boldsymbol{t}}_i||\leq\beta$.
\end{proposition}

Similar to Section~\eqref{RS}, if there exist $M$ 3-point sets {$\mathcal{M}=\{\{(\mathbf{p}_i, \mathbf{q}_i)_d\}_{i=1}^3\}_{d=1}^M$} that pass the first two compatibility conditions, called the \textit{potential 3-point sets}, we can now define a new set of rotation-based invariant functions.

\begin{proposition}[Final 6-Invariant Function] \label{Prop5} We define a set of 6-invariant functions $\{\boldsymbol{r}_d(\mathbf{p}_i, \mathbf{q}_i)\}_{d=1}^2$ satisfying that
\begin{equation}\label{Ul-comp1}
\boldsymbol{r}_d(\mathbf{p}_i, \mathbf{q}_i)=\boldsymbol{{R}}^{*}_d=\boldsymbol{{R}}\cdot {Exp}(\boldsymbol{\rho}^{\boldsymbol{R}}_d)
\end{equation}
where $\boldsymbol{r}_d(\mathbf{p}_i, \mathbf{q}_i)$ is invariant to $\mathit{s}$ and $\boldsymbol{t}$, and $\boldsymbol{\rho}^{\boldsymbol{R}}_d$ is the noise measurement of rotations. For any two potential 3-point sets {$\{\{(\mathbf{p}_i, \mathbf{q}_i)_d\}_{i=1}^3\}_{d=1}^2$}, in the presence of noise, we have the following compatibility conditions: \\ (i) functions $\{\boldsymbol{r}_d(\mathbf{p}_i, \mathbf{q}_i)\}^{2}_{d=1}$ should satisfy that
\begin{equation}\label{Ul-comp2}
\angle\left(\boldsymbol{r}_1(\mathbf{p}_i, \mathbf{q}_i), \boldsymbol{r}_2(\mathbf{p}_j, \mathbf{q}_j)\right) \leq \gamma,
\end{equation}
where $\gamma$ represents the bound of noise (The proof here is analogous to that of Proposition~\ref{Prop-2});\\
(ii) merging the two 3-point sets as a single 6-point set: {$\{(\mathbf{p}_i, \mathbf{q}_i)_l\}_{l=1}^6$}, we can also have 
\begin{align}\label{Ul-comp3}
||{s}_a(\mathbf{p}_a, \mathbf{q}_a)-{s}_b(\mathbf{p}_b, \mathbf{q}_b)||&\leq \alpha\left(\frac{1}{||\mathbf{\widetilde{p}}_a||}+\frac{1}{||\mathbf{\widetilde{p}}_b||}\right), \\
||\boldsymbol{t}_a(\mathbf{p}_a, \mathbf{q}_a)-\boldsymbol{t}_b(\mathbf{p}_b, \mathbf{q}_b)||&\leq 2\beta, \label{Ul-comp4}
\end{align}
where $a,b\in\{1,2,\dots,6\}\,\,(a\neq b)$.
\end{proposition}

Proposition \ref{Prop5} forms our last compatibility condition for extracting the qualified inliers from the potential 3-point sets.

\subsection{RANSIC for Point Cloud Registration}

The pseudocode of RANSIC for point cloud registration with unknown scale is given in Algorithm~\ref{Algo2-PCR}.

\textbf{Description of Algorithm~\ref{Algo2-PCR}.} The procedures of RANSIC in \textit{lines 4-15} are similar to Algorithm~\ref{Algo1-RS}. What is different is that here we design a second iteration to handle very extreme outlier ratios (e.g. 99\%). (Usually, iteration 2 is not required for 0-98\% outlier ratios because $label=1$ (\textit{lines 12-15}) can be already attained in iteration 1.) Once a $3$-degree vertex or set $\mathcal{X}$ of size 500 is obtained (\textit{lines 16-19}) but condition ~\eqref{distribution} is still unsatisfied, RANSIC just breaks from iteration 1 and enters iteration 2, in which the compatibility tests become stricter, in order to enhance time-efficiency.

\section{Experiments}

In this section, we validate the proposed solver RANSIC in a variety of experiments and also compare it with the existing state-of-the-art solvers. All the experiments are implemented in Matlab on a standard laptop with a Core i7-7700HQ CPU and 16GB RAM. And NO parallelism programming is used.

\begin{algorithm}[t]
\caption{RANSIC for Point Cloud Registration}
\label{Algo2-PCR}
\SetKwInOut{Input}{\textbf{Input}}
\SetKwInOut{Output}{\textbf{Output}}
\Input{correspondences $\mathcal{C}=\{(\mathbf{p}_i,\mathbf{q}_i)\}_{i=1}^N$; \\ initial $K$ to break sampling\;}
\Output{optimal $(\mathit{s}^{\star}, \boldsymbol{R}^{\star}, \boldsymbol{t}^{\star})$; inlier set $\mathcal{C}^{\star}$ \;}
\BlankLine
Set $\mathcal{X}=\emptyset$, $\mathcal{C}^{\star}=\emptyset$, $K=1$ and $label=0$\;
\For {$itr=1:2$}{
Set noise-bounds $\alpha, \beta, \gamma$ for the compatibility tests (noise-bounds are stricter when $itr=2$)\;
\While {true}{
{Randomly select $\mathcal{S}=\{(\mathbf{p}_j,\mathbf{q}_j)\}_{j=1}^3\subset\mathcal{C}$}\;
\If {set $\mathcal{S}$ can pass tests \eqref{s-comp} and \eqref{t-comp}}{
$\mathcal{X}=\mathcal{X}\cup\{\mathcal{S}\}$ as the vertices\;
Find all vertices $\mathcal{Y}\subset \mathcal{X}$ having edges meeting vertex ${V}_{\mathcal{S}}$ with tests \eqref{Ul-comp1}-\eqref{Ul-comp4}\;
\If {${V}_{\mathcal{S}}$ has degree at least $K$}{
Solve $\mathit{s}$, $\boldsymbol{R}$ and $\boldsymbol{t}$ with $\mathcal{Y}$\;
Compute residuals $\mathcal{r}=\{r_i\}_{i=1}^N$\;
\If {$\mathcal{r}$ fulfills condition \eqref{distribution}}{
$label=1$\;
\textbf{break}
}
\If {itr=1}{
\If {$K\geq3$ or $|G|\geq500$}{
\textbf{break}
}
}
$K=K+1$\;
}
}
}
\If {$label=1$}{
\textbf{break}
}
}
Add all correspondences whose $r_i\leq5.2\sigma$ to set $\mathcal{C}^{\star}$\;
Solve $\mathit{s}^{\star}$, $\boldsymbol{R}^{\star}$ and $\boldsymbol{t}^{\star}$ with the inlier set $\mathcal{C}^{\star}$\;
\Return $(\mathit{s}^{\star}, \boldsymbol{R}^{\star}, \boldsymbol{t}^{\star})$ and $\mathcal{C}^{\star}$\;
\end{algorithm}

\subsection{Synthetic Experiments on Rotation Search}
\label{Ex-A}

We first evaluate RANSIC on rotation search in a synthetic environment. We generate $N=\{100,500,1000\}$ random unit-norm vectors $\mathcal{A}=\{\mathbf{a}_i\}_{i=1}^{N}$, and rotate set $\mathcal{A}$ with a random rotation $\boldsymbol{R}\in SO(3)$ to achieve its corresponding vector set $\mathcal{B}=\{\mathbf{b}_i\}_{i=1}^{N}$. We add random zero-mean Gaussian noise with standard deviation $\sigma=0.01$ to the vectors in $\mathcal{B}$. To create outliers, a portion of the vectors in $\mathcal{B}$ (0-99\%) are substituted by randomly-generated unit-norm vectors instead. All the results are computed over 50 Monte Carlo runs.

\begin{figure*}[htp]
\centering

\footnotesize{(a) $N=100$}

\subfigure{
\begin{minipage}[t]{1\linewidth}
\centering
\includegraphics[width=0.246\linewidth]{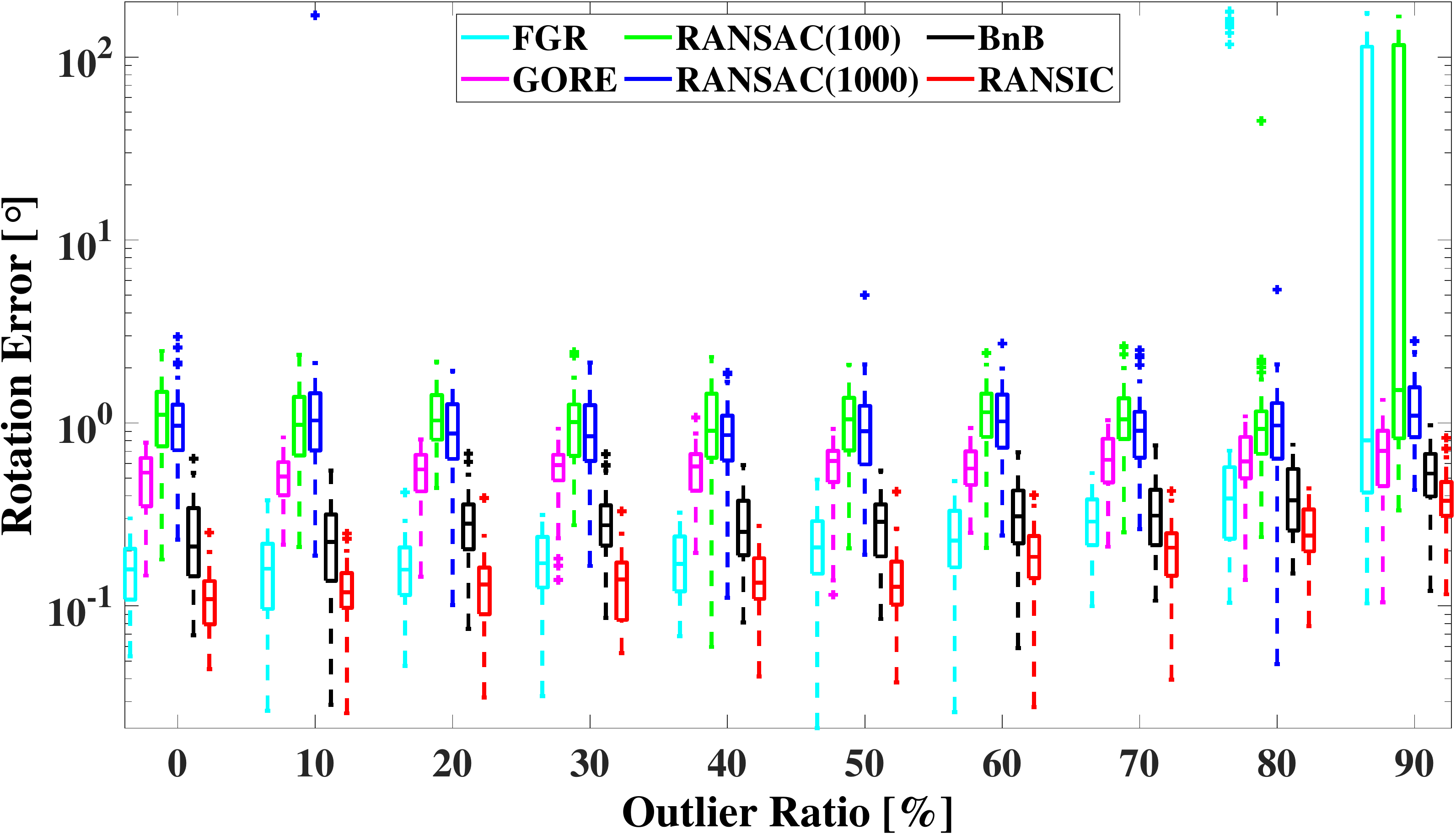}
\includegraphics[width=0.246\linewidth]{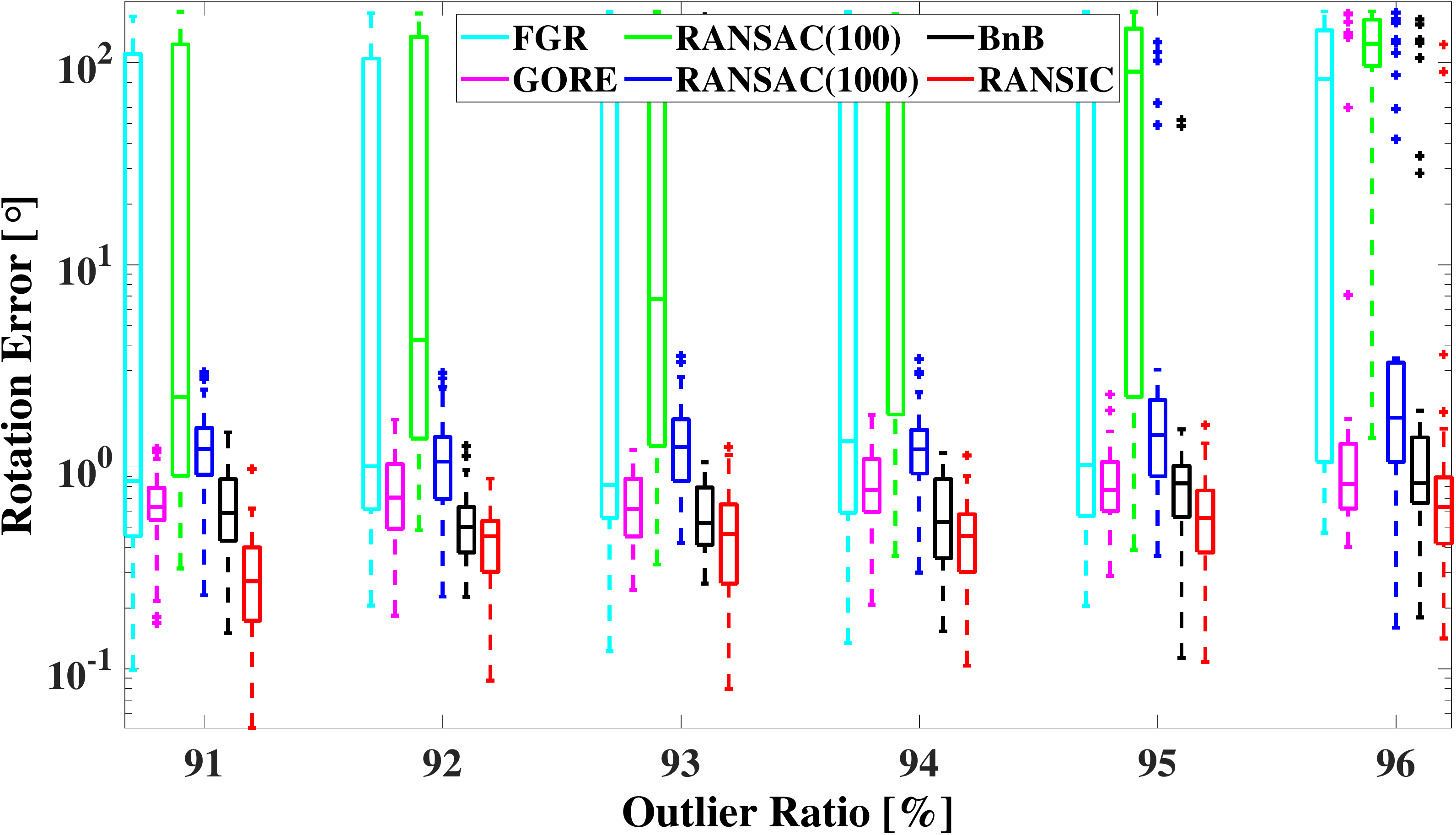}
\includegraphics[width=0.246\linewidth]{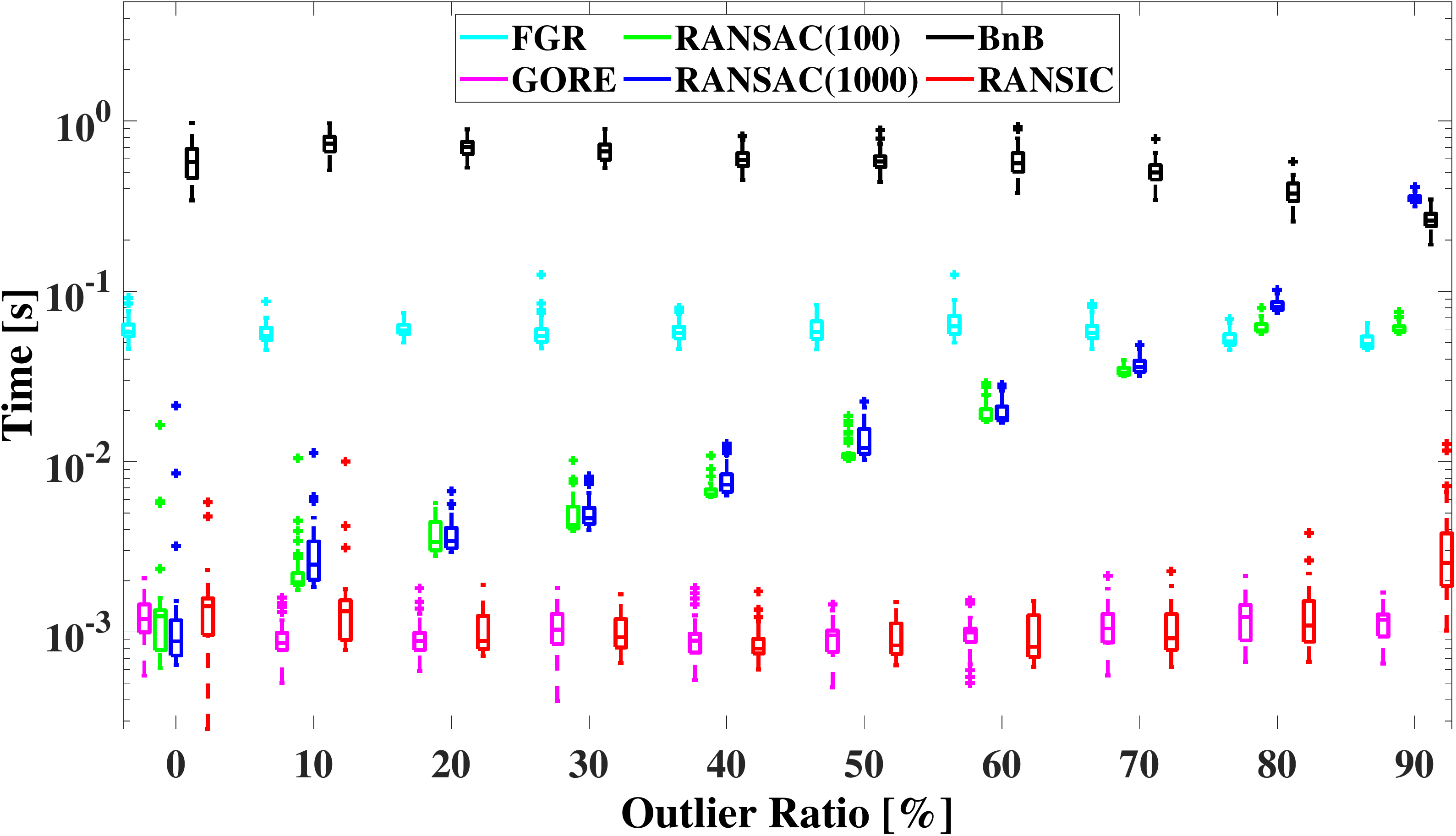}
\includegraphics[width=0.246\linewidth]{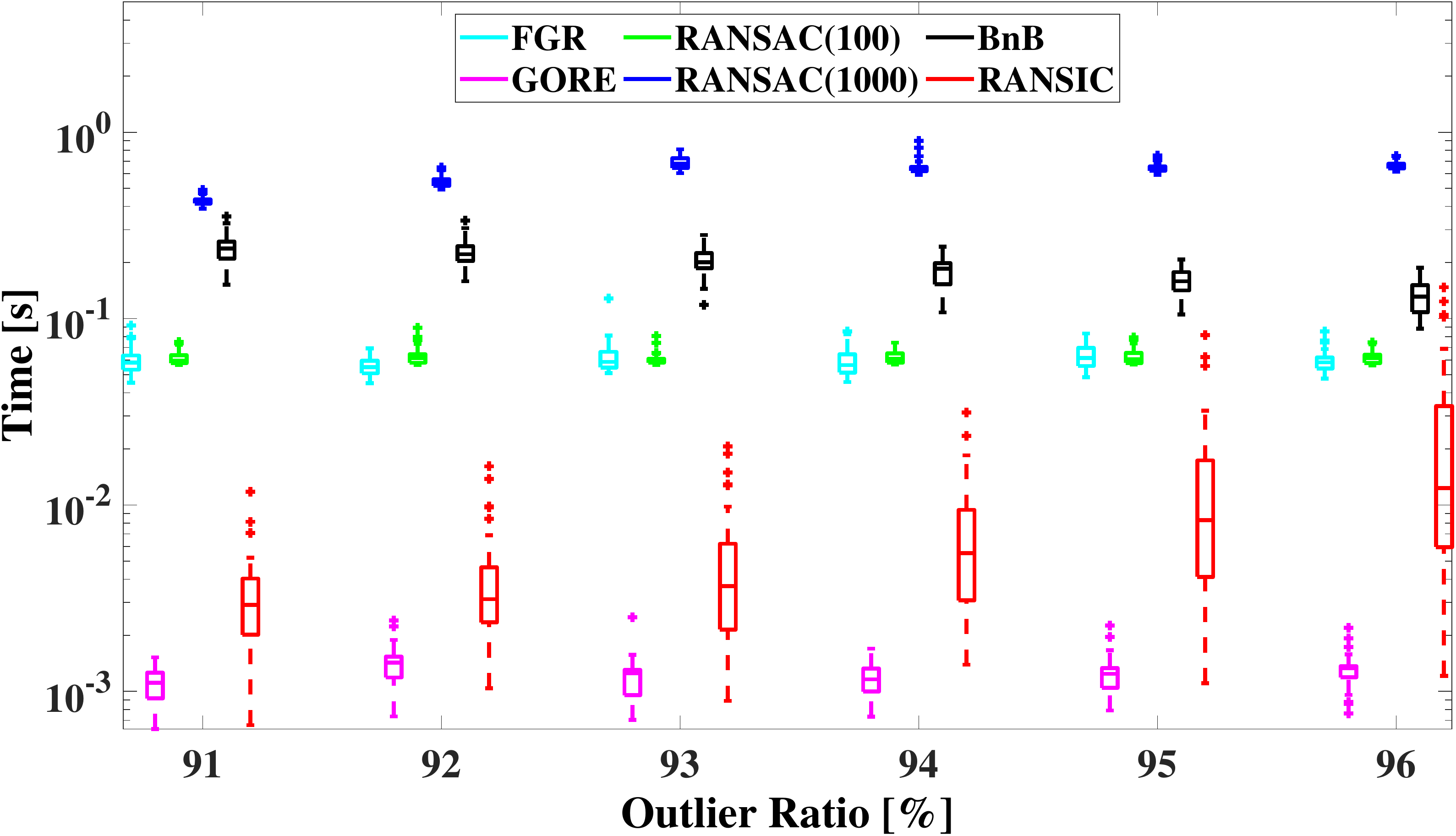}
\end{minipage}
}%

\footnotesize{(b) $N=500$}

\subfigure{
\begin{minipage}[t]{1\linewidth}
\centering
\includegraphics[width=0.246\linewidth]{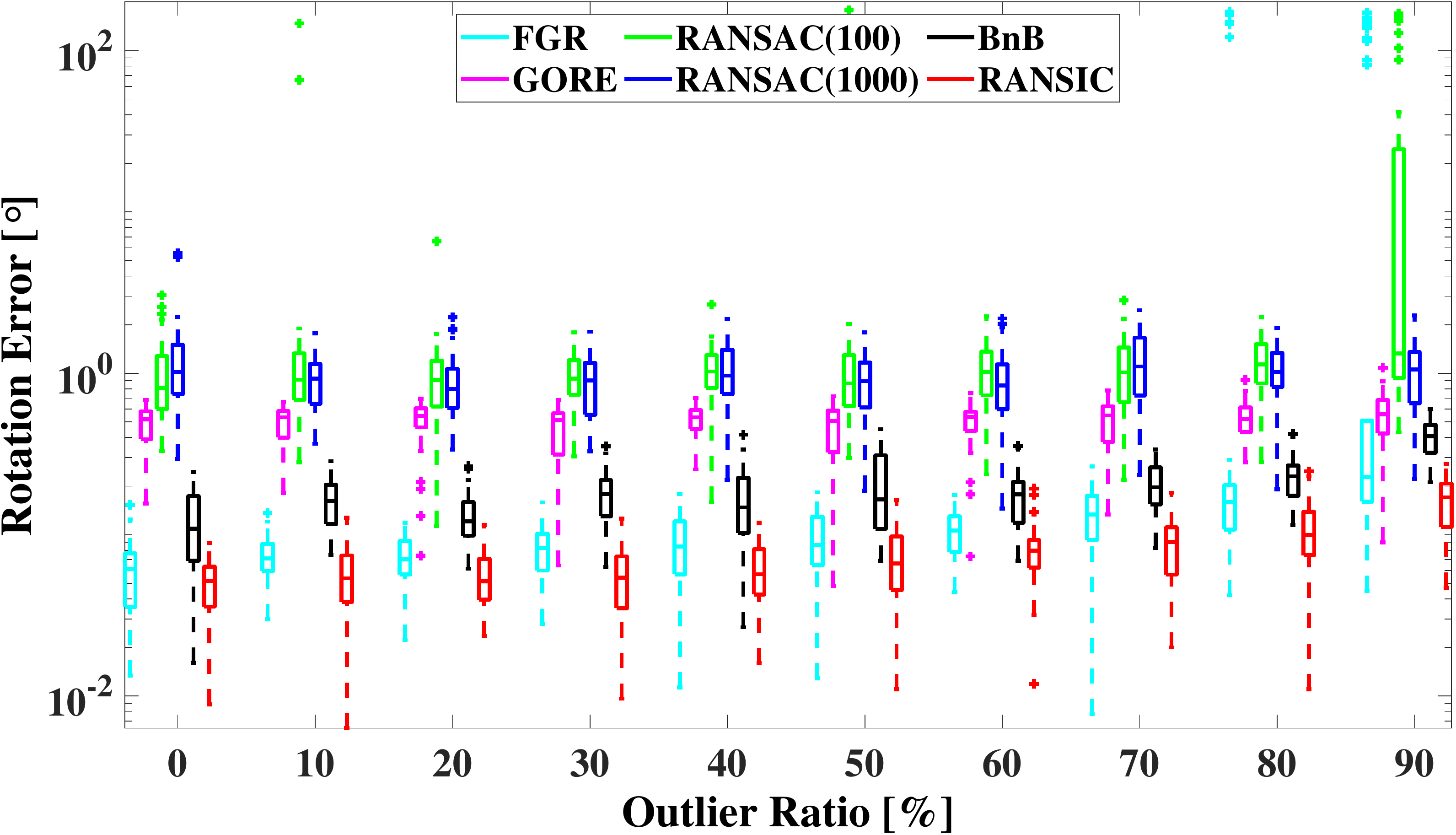}
\includegraphics[width=0.246\linewidth]{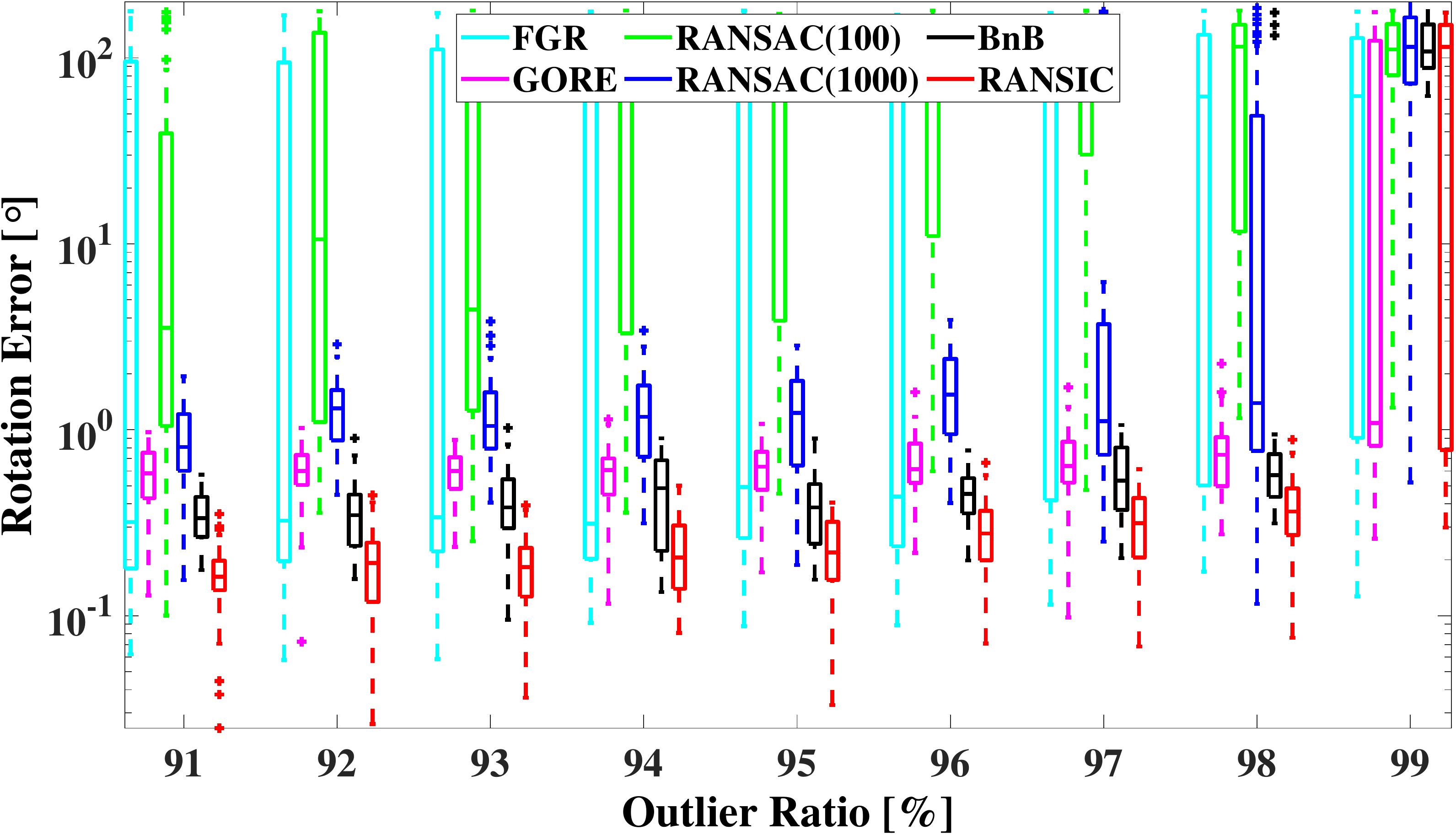}
\includegraphics[width=0.246\linewidth]{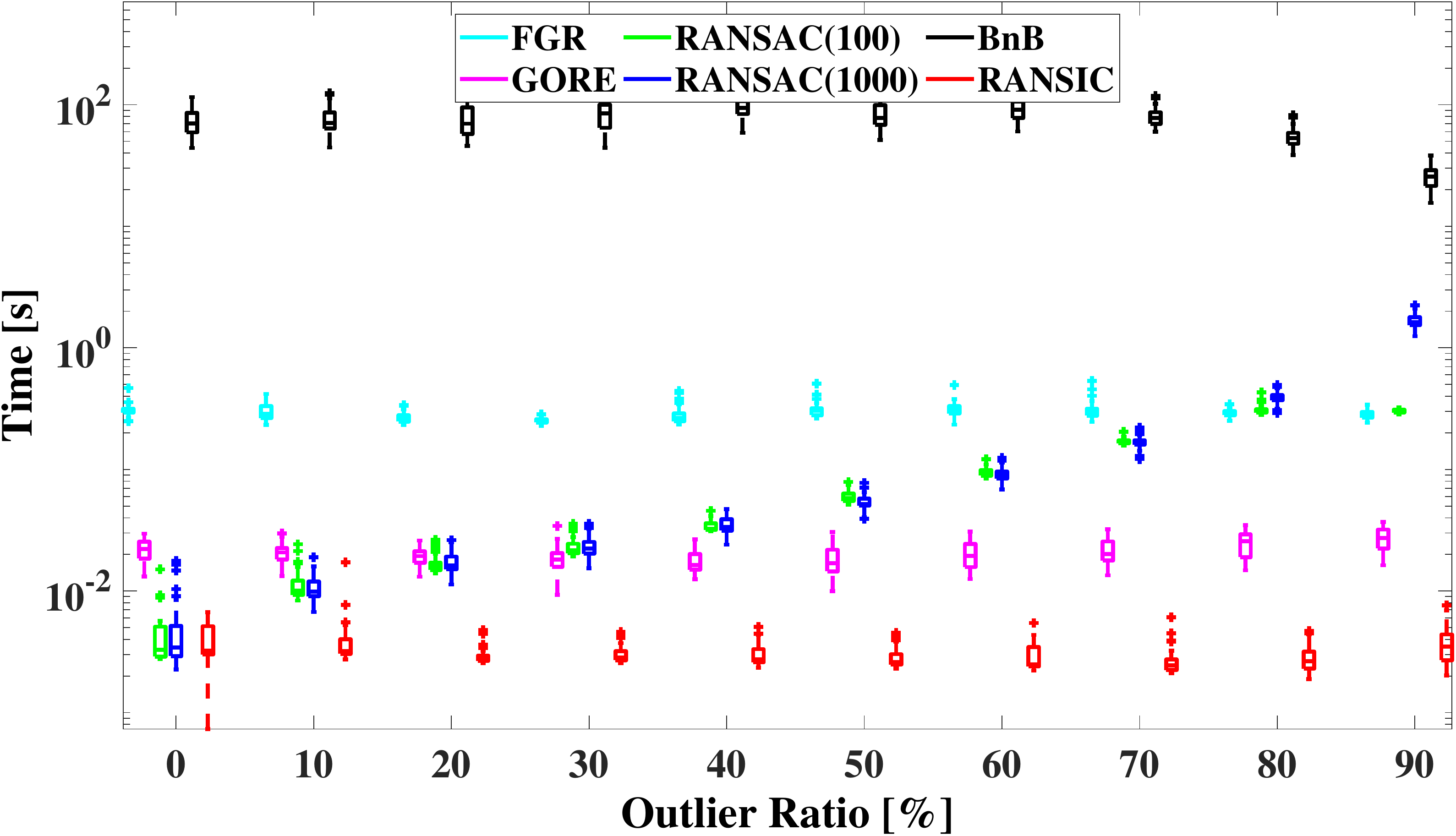}
\includegraphics[width=0.246\linewidth]{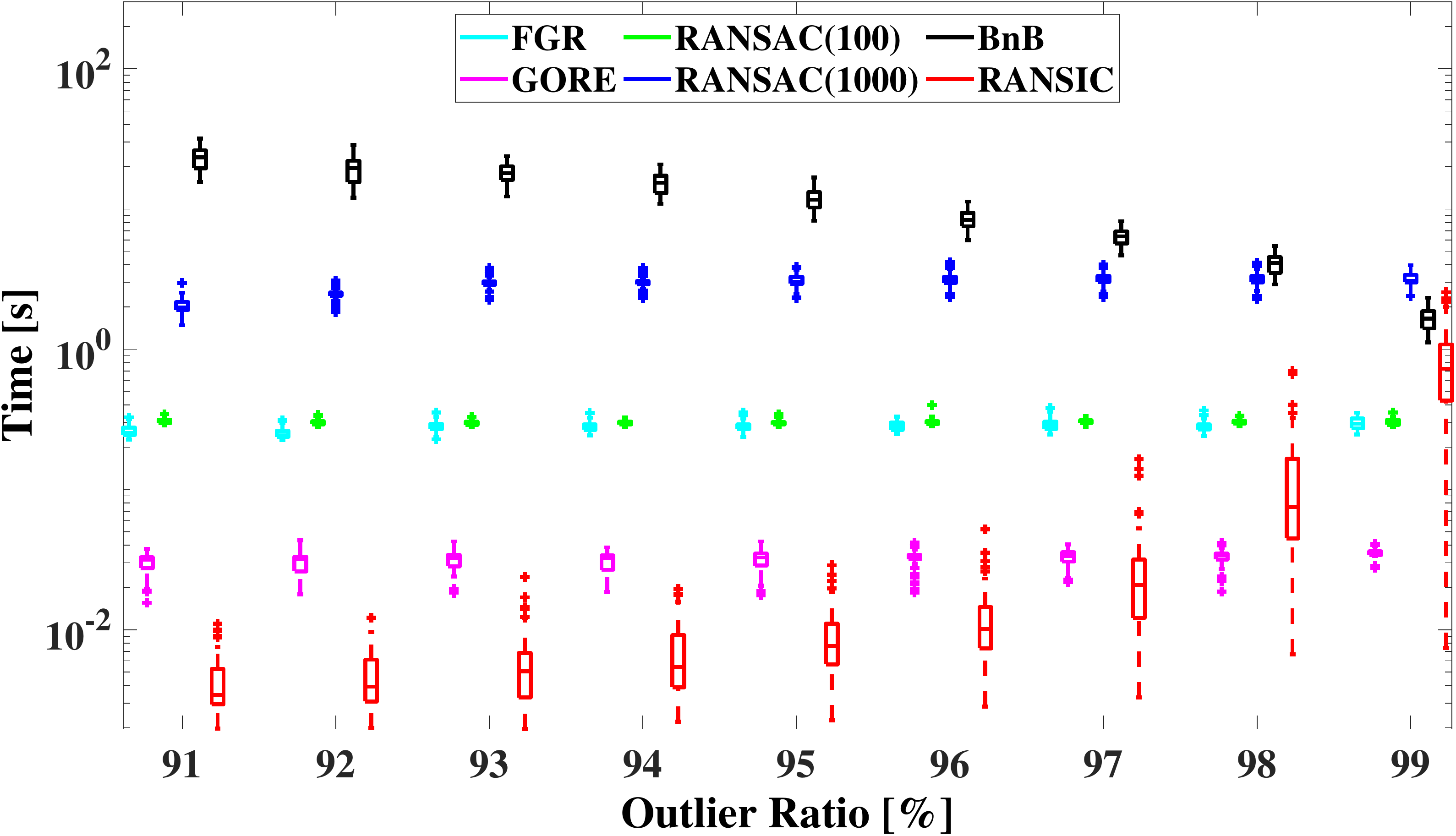}
\end{minipage}
}%

\footnotesize{(c) $N=1000$}

\subfigure{
\begin{minipage}[t]{1\linewidth}
\centering
\includegraphics[width=0.246\linewidth]{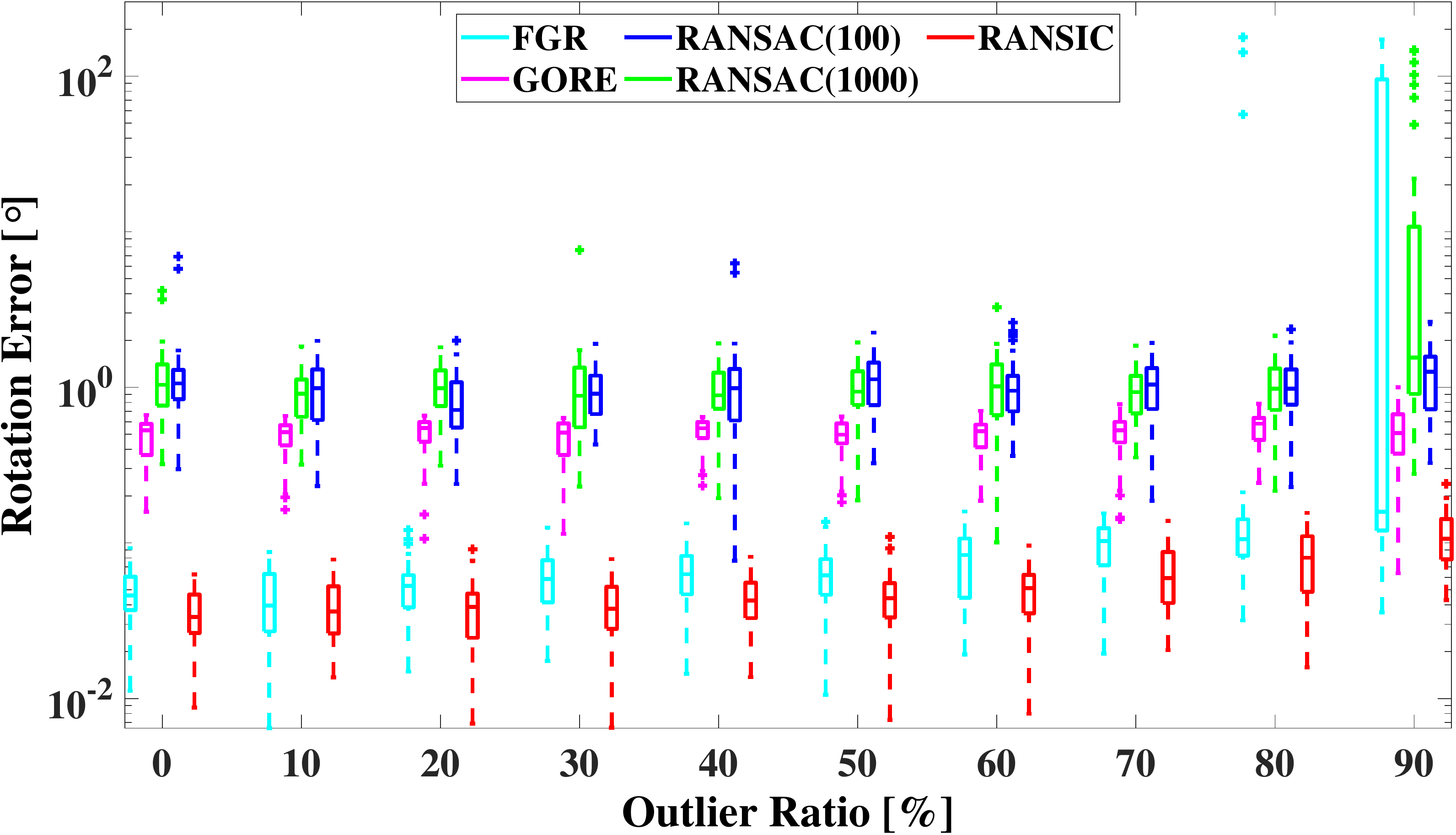}
\includegraphics[width=0.246\linewidth]{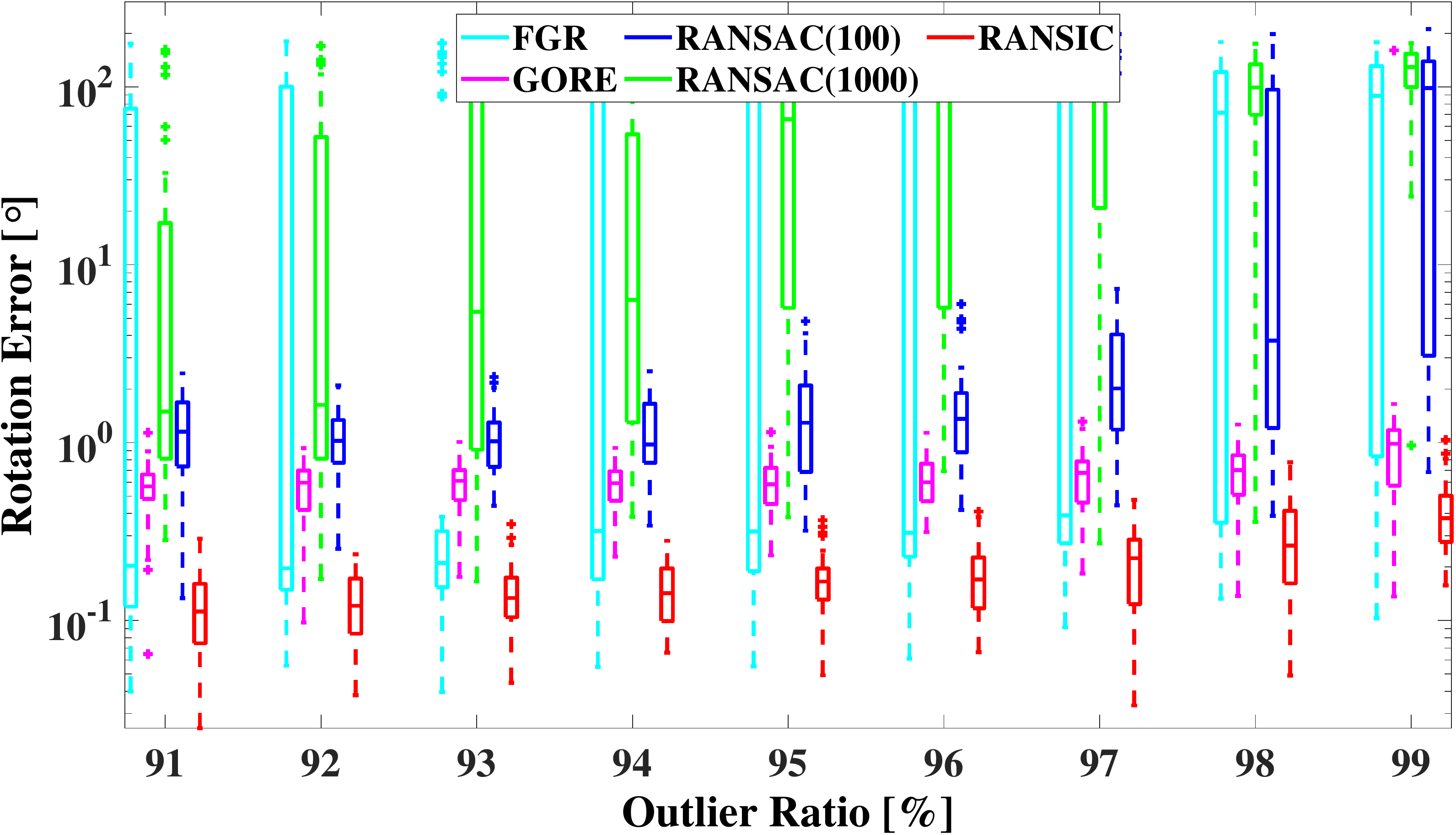}
\includegraphics[width=0.246\linewidth]{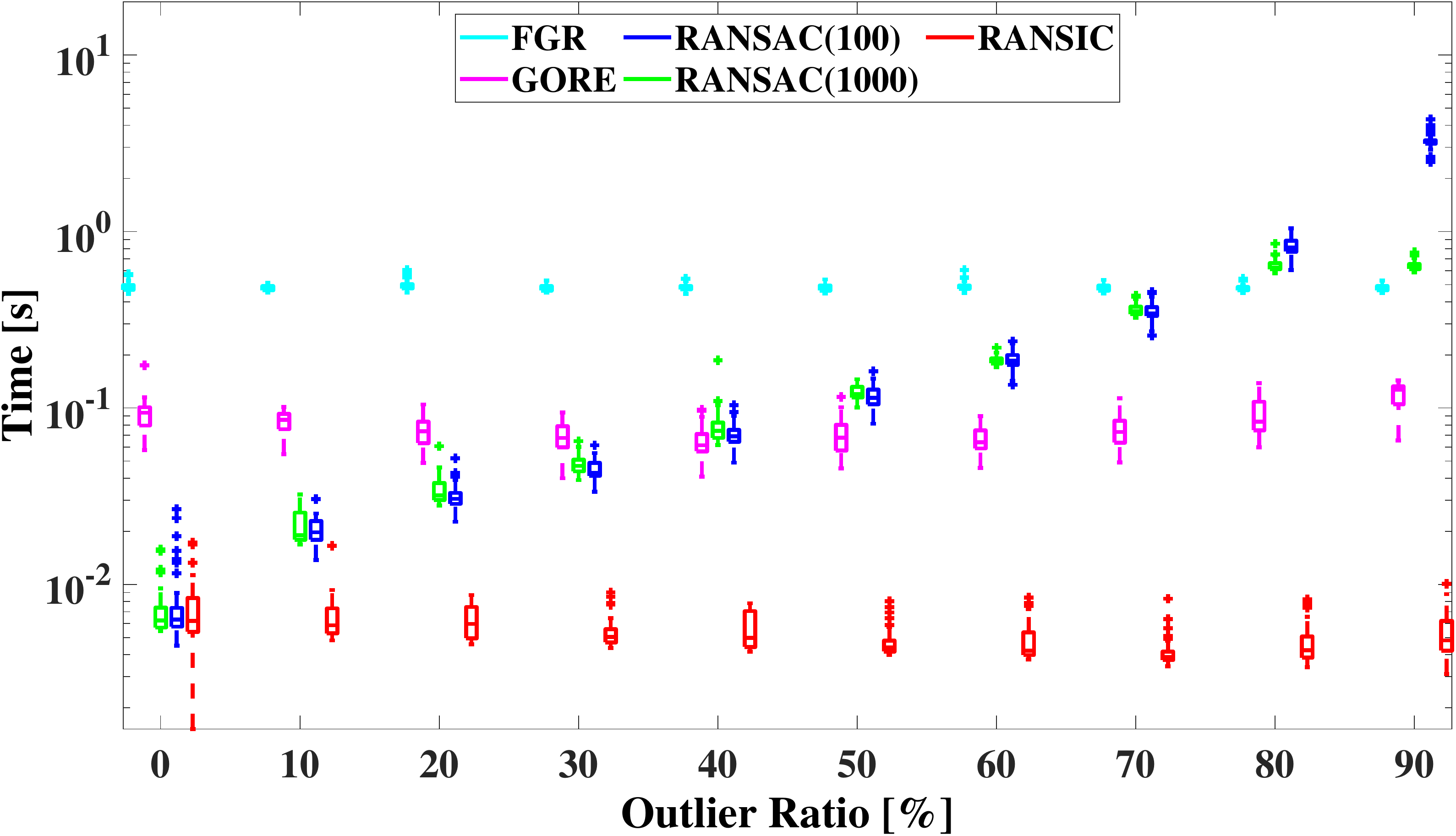}
\includegraphics[width=0.246\linewidth]{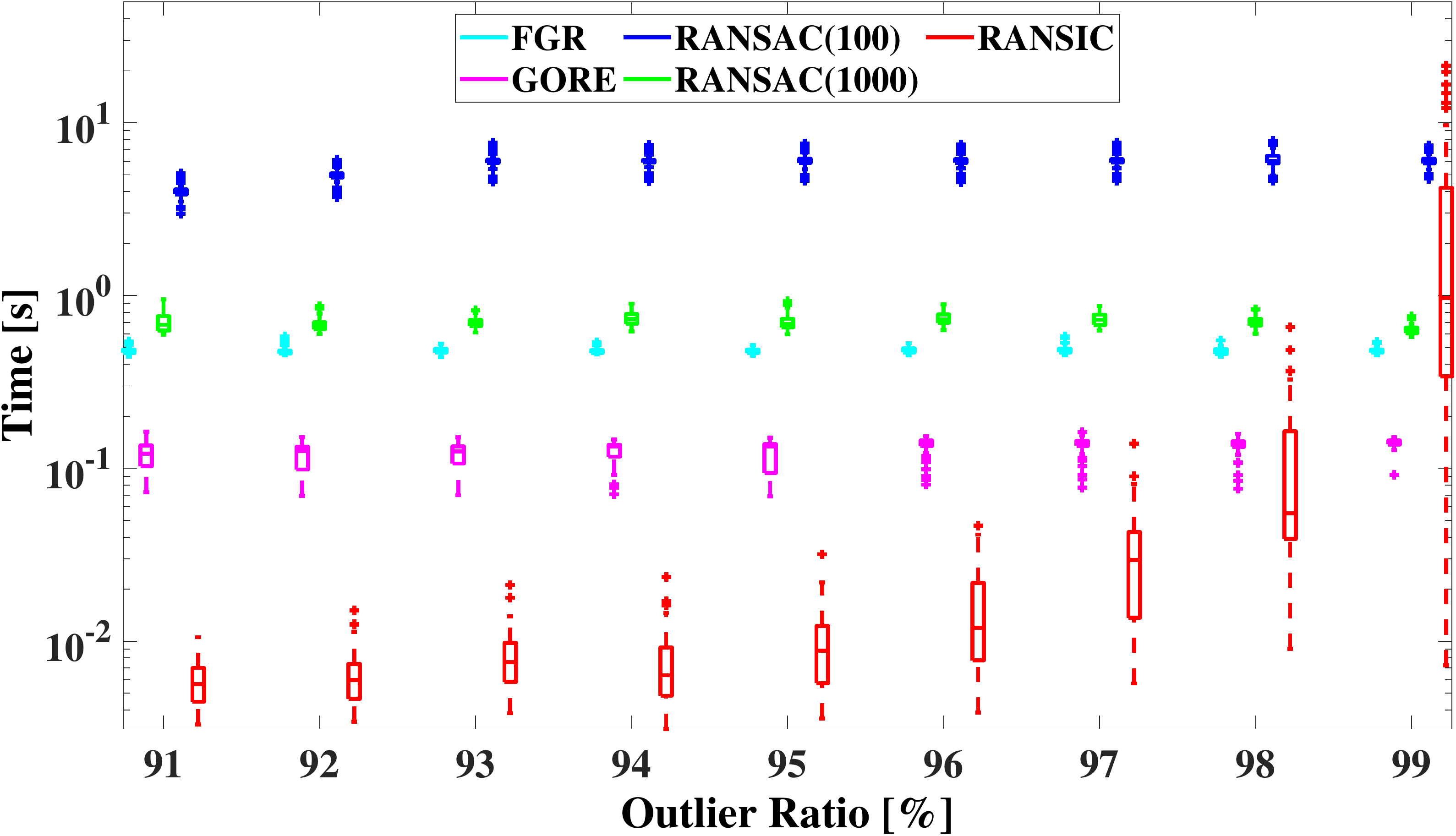}
\end{minipage}
}%
\vspace{-3mm}
\centering
\caption{Experimental results on rotation search over synthetic data. (a) Results with $N=100$. Parameter setup: $\zeta=0.012$, $\upsilon=2.6$, and for 0-95\%, $\theta=5^{\circ}$ and $\tau=5$, while for 96\% (RANSIC may fail), $\theta=6^{\circ}$ and $\tau=4$. (b) Results with $N=500$. Parameter setup: $\zeta=0.008$, $\theta=5^{\circ}$, $\upsilon=2.6$, and for 0-98\%, $\tau=10$, while for 99\% (RANSIC fails), $\tau=5$. (c) Results with $N=1000$. Parameter setup: $\zeta=0.008$, $\theta=5^{\circ}$, $\upsilon=2.6$, $\tau=10$.}
\label{Syn-RS}
\end{figure*}

We benchmark RANSIC against FGR~\cite{zhou2016fast} (rotation only), GORE~\cite{dasari2014park}, RANSAC (with 0.995 confidence)+closed-form solver~\cite{arun1987least} set with two maximum iteration numbers: 100 for RANSAC(100) and 1000 for RANSAC(1000), and BnB~\cite{parra2014fast}. For $N=1000$, we exclude BnB as it takes tens of minutes per run. QUASAR~\cite{yang2019quaternion} is also not used due to its long runtime. Fig.~\ref{Syn-RS} shows the results on accuracy and runtime. To represent the errors of $\boldsymbol{R}$, we adopt the geodesic errors~\eqref{geo-error}.

We can observe that (i) both FGR and RANSAC(100) fail at 90\% and RANSAC(1000) breaks at 98\%, (ii) RANSIC has the highest robustness: it is robust against 95-96\% with $N=100$, 98\% with $N=500$, and 99\% with $N=1000$, (iii) RANSIC is accurate: it has the lowest rotation errors generally, and (iv) RANSIC is time-efficient: it is the fastest solver when the outlier ratio is within 0-96\% with $N=\{500,1000\}$ correspondences and is almost as fast as GORE when the outlier ratio is within 0-80\% with $N=100$.

\begin{figure*}[htp]
\centering

\footnotesize{(a) $N=1000$, Known Scale: $\mathit{s}=1$}

\subfigure{
\begin{minipage}[t]{0.492\linewidth}
\centering
\includegraphics[width=0.493\linewidth]{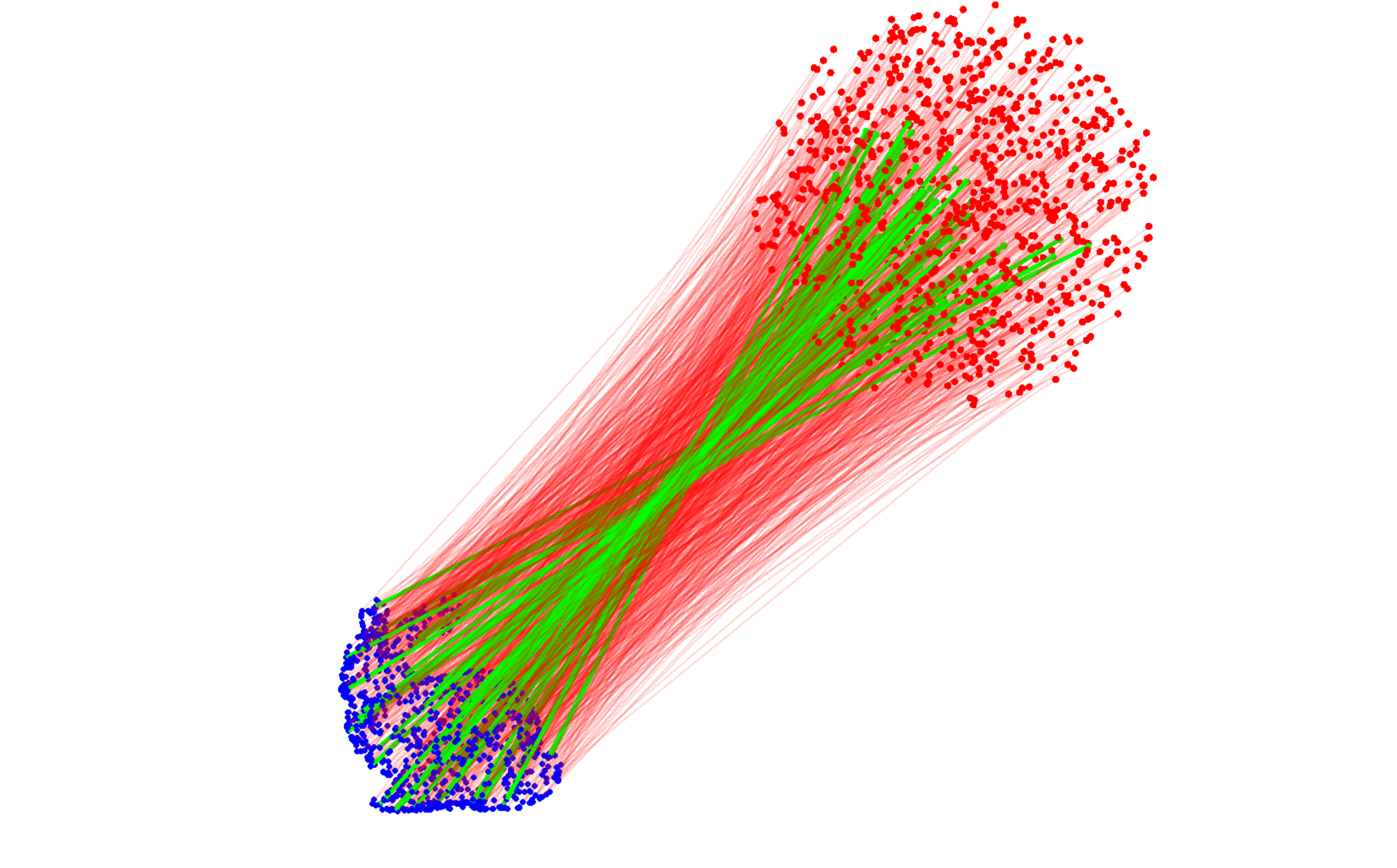}
\includegraphics[width=0.493\linewidth]{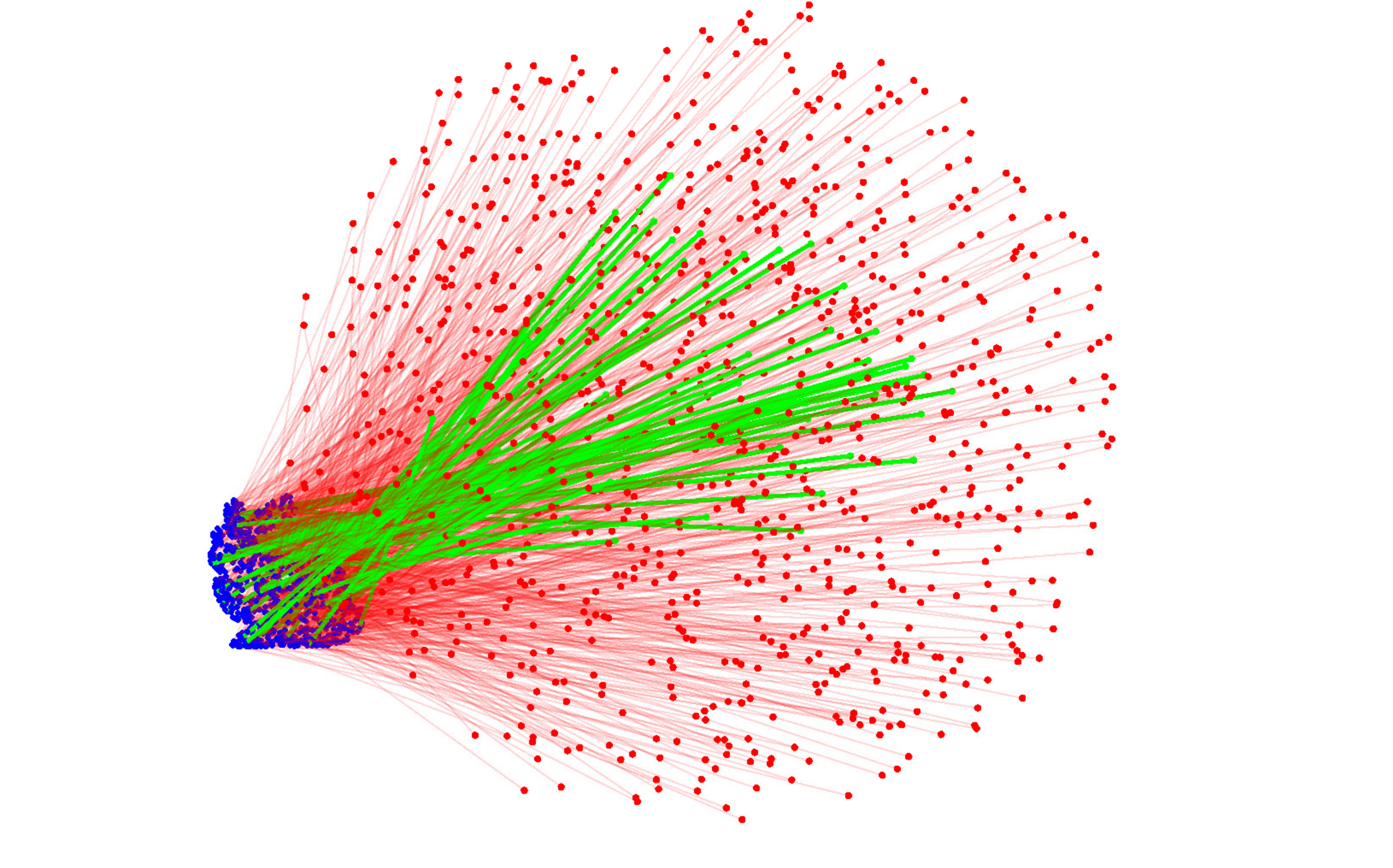}
\end{minipage}
}%
\subfigure{
\begin{minipage}[t]{0.495\linewidth}
\centering
\includegraphics[width=0.495\linewidth]{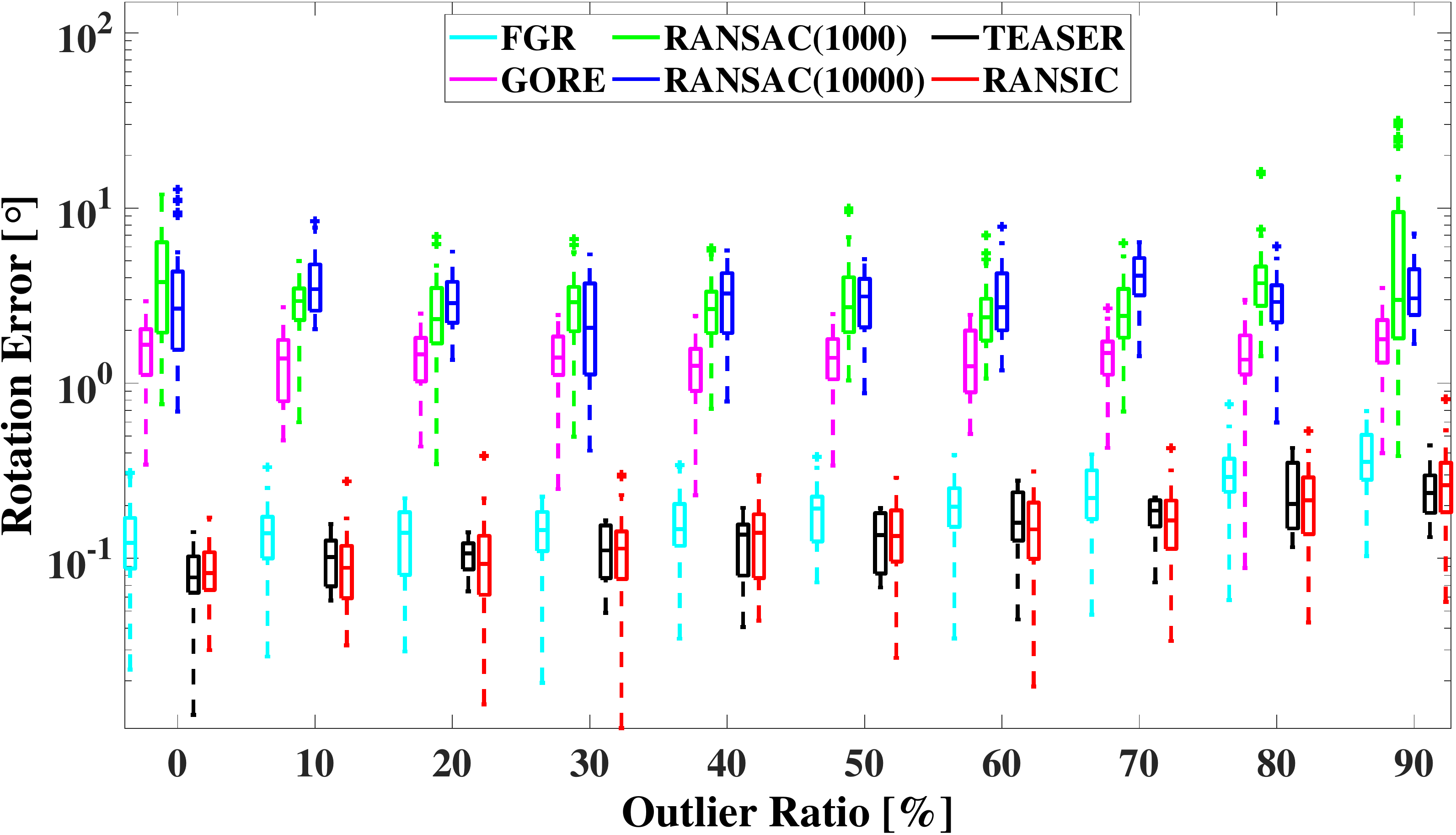}
\includegraphics[width=0.495\linewidth]{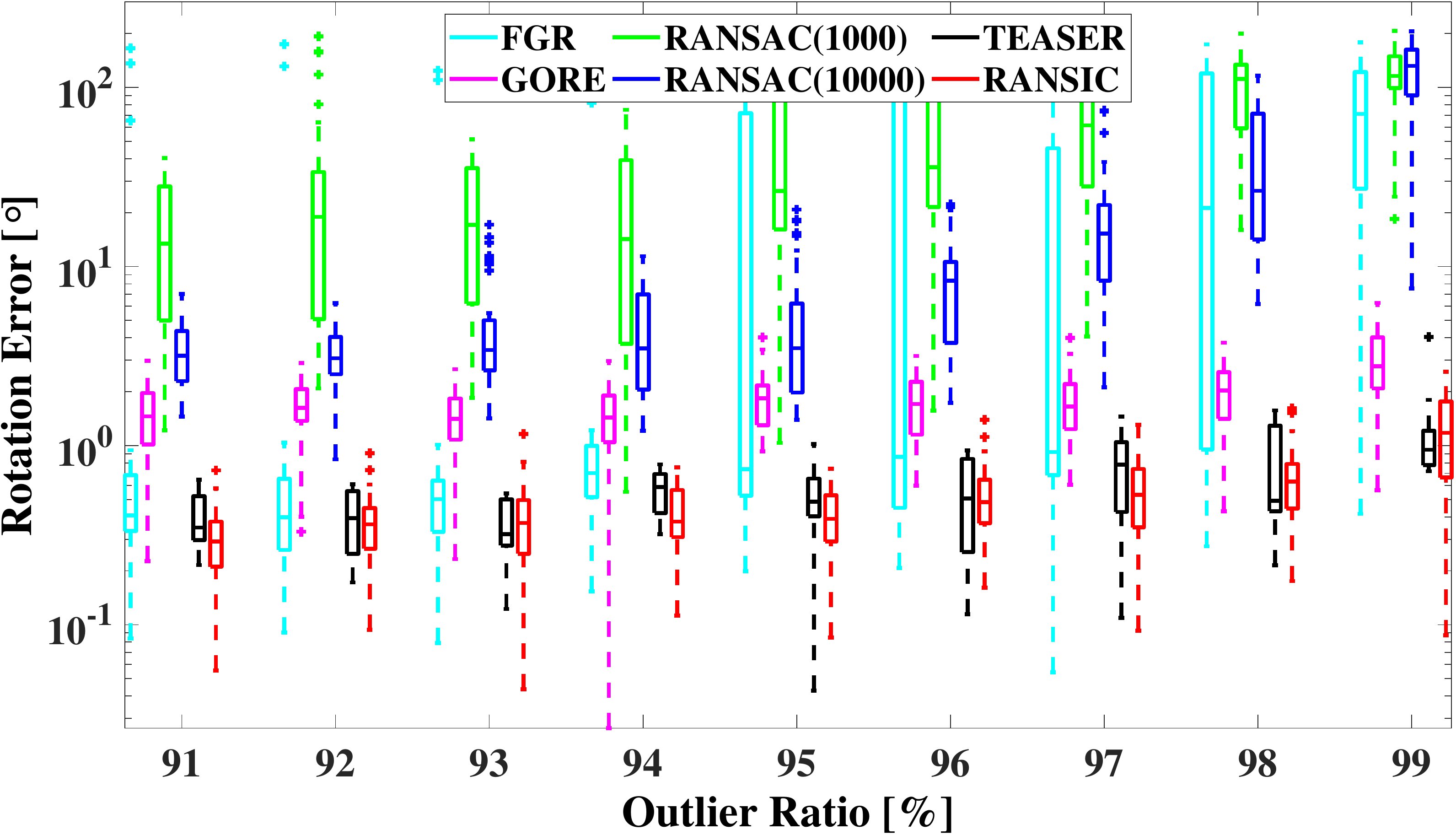}
\end{minipage}
}%

\subfigure{
\begin{minipage}[t]{1\linewidth}
\centering
\includegraphics[width=0.246\linewidth]{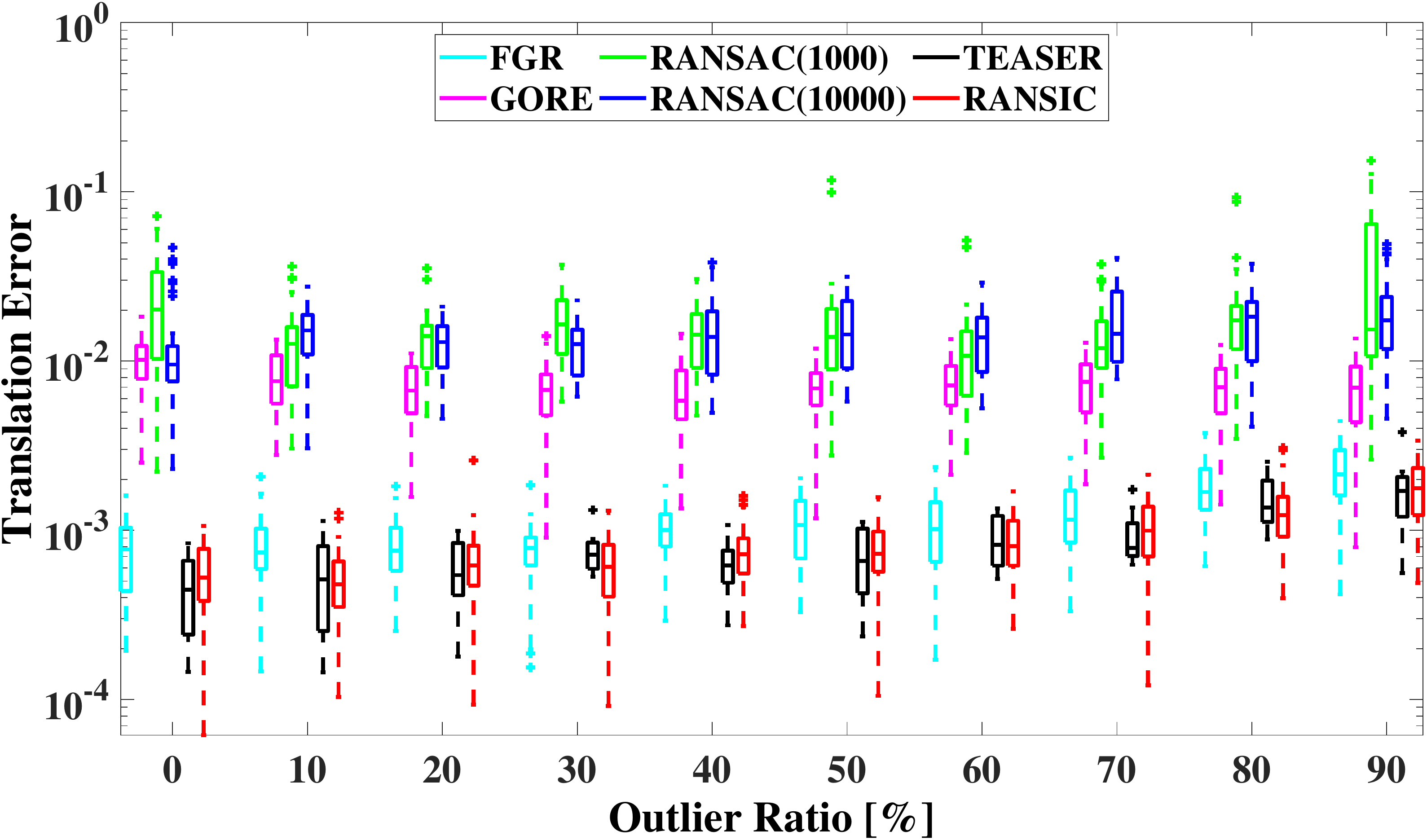}
\includegraphics[width=0.246\linewidth]{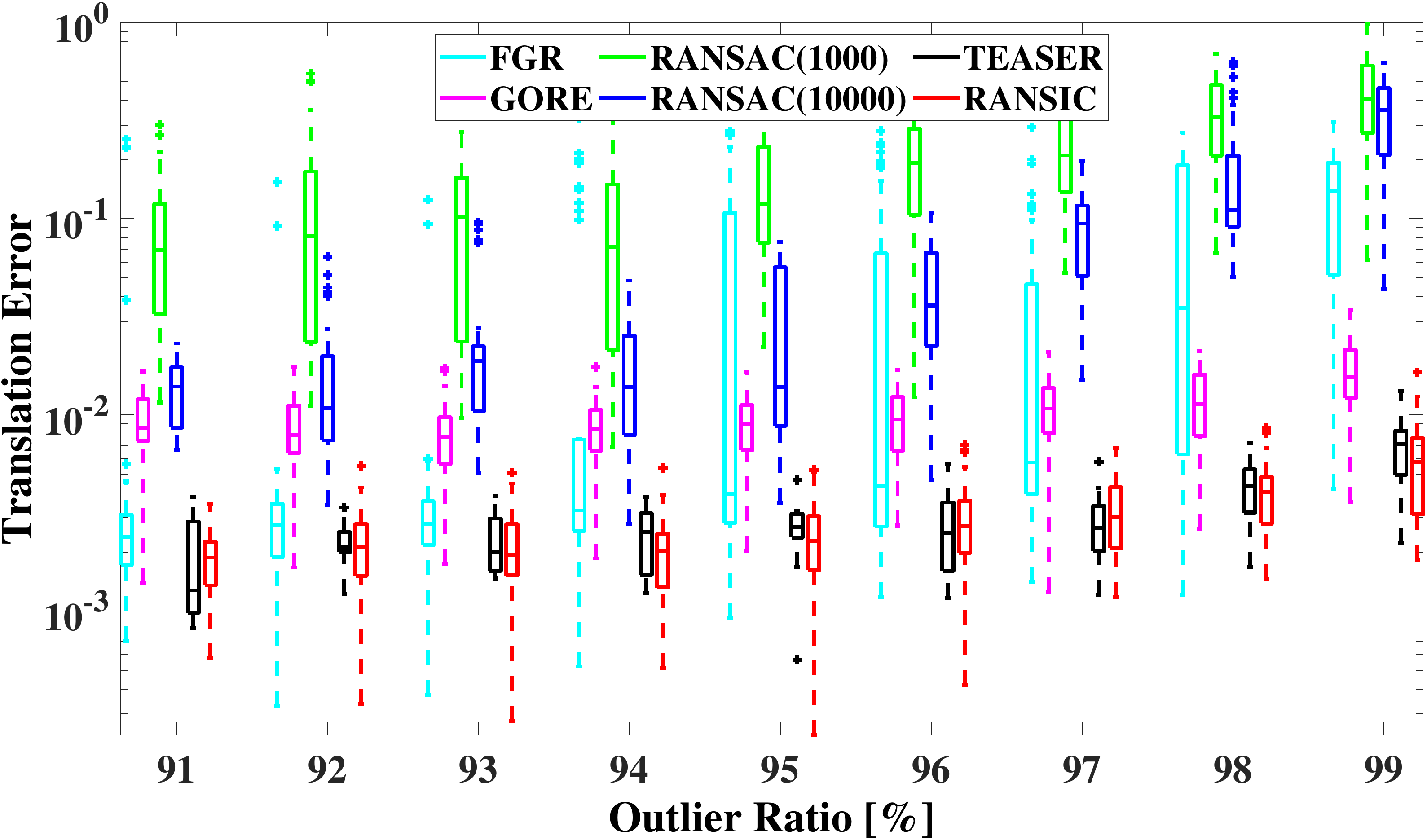}
\includegraphics[width=0.246\linewidth]{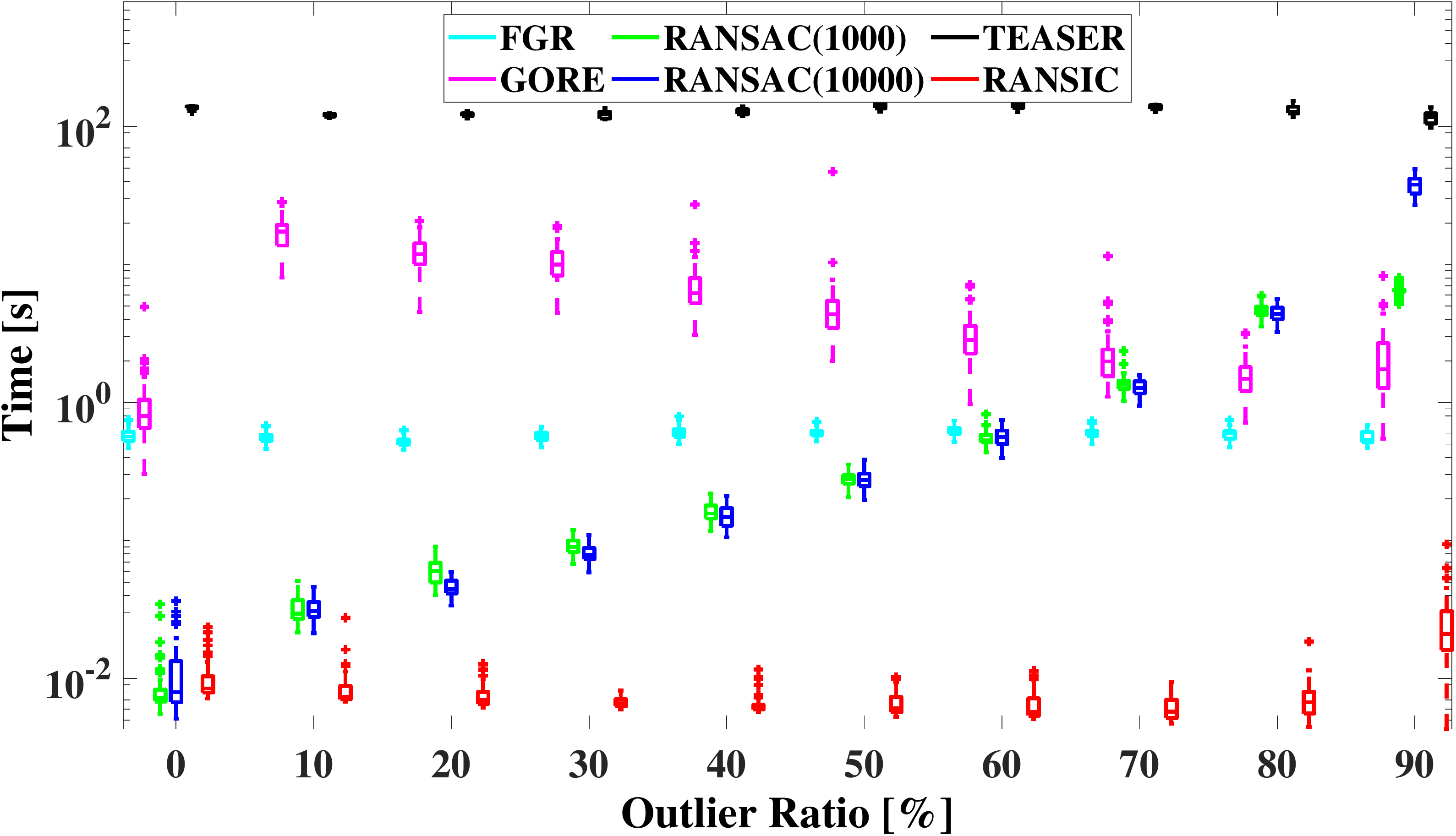}
\includegraphics[width=0.246\linewidth]{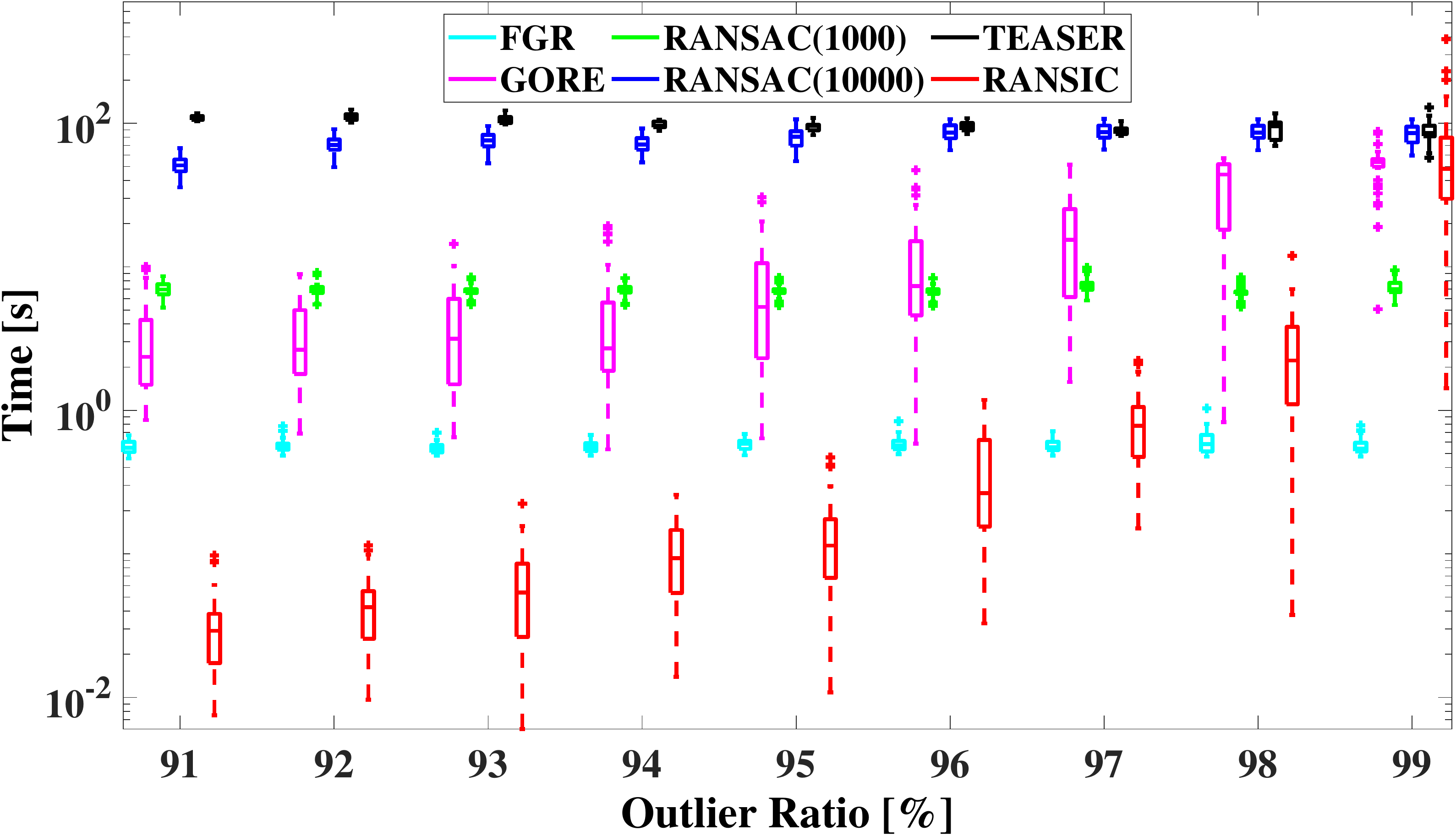}
\end{minipage}
}%

\footnotesize{(b) $N=1000$, Unknown Scale: $\mathit{s}\in(1,5)$}

\subfigure{
\begin{minipage}[t]{1\linewidth}
\centering
\includegraphics[width=0.246\linewidth]{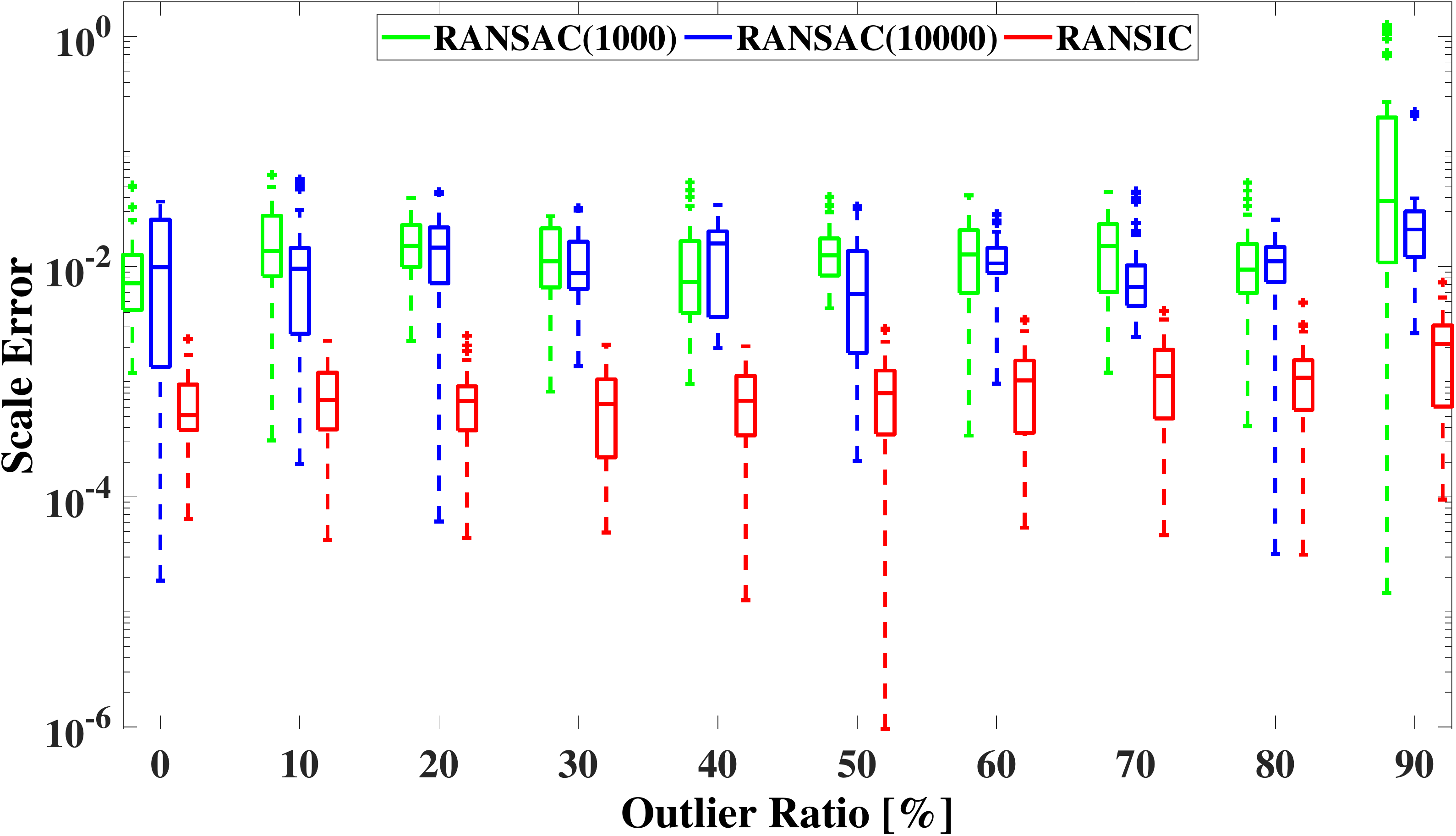}
\includegraphics[width=0.246\linewidth]{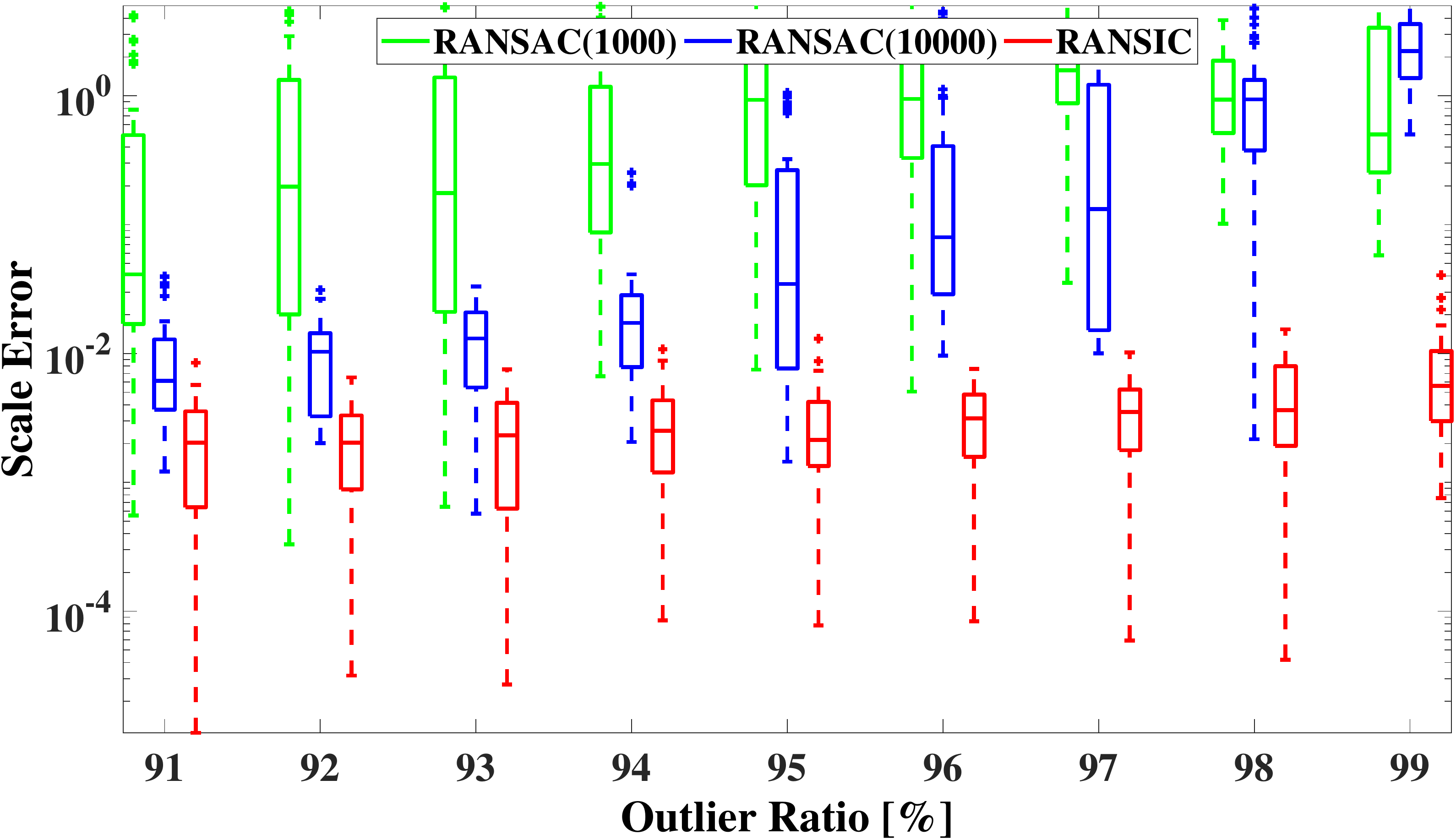}
\includegraphics[width=0.246\linewidth]{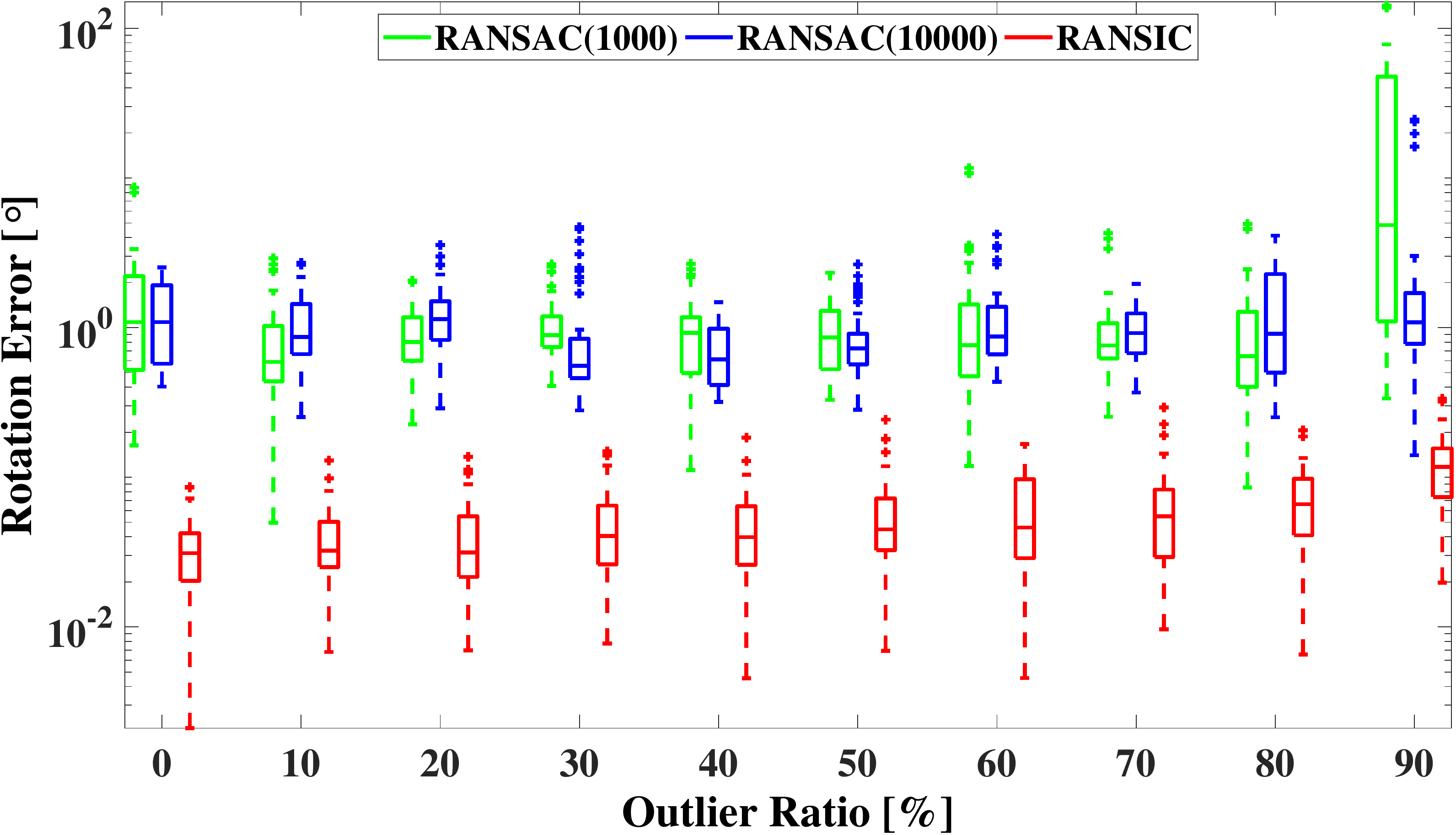}
\includegraphics[width=0.246\linewidth]{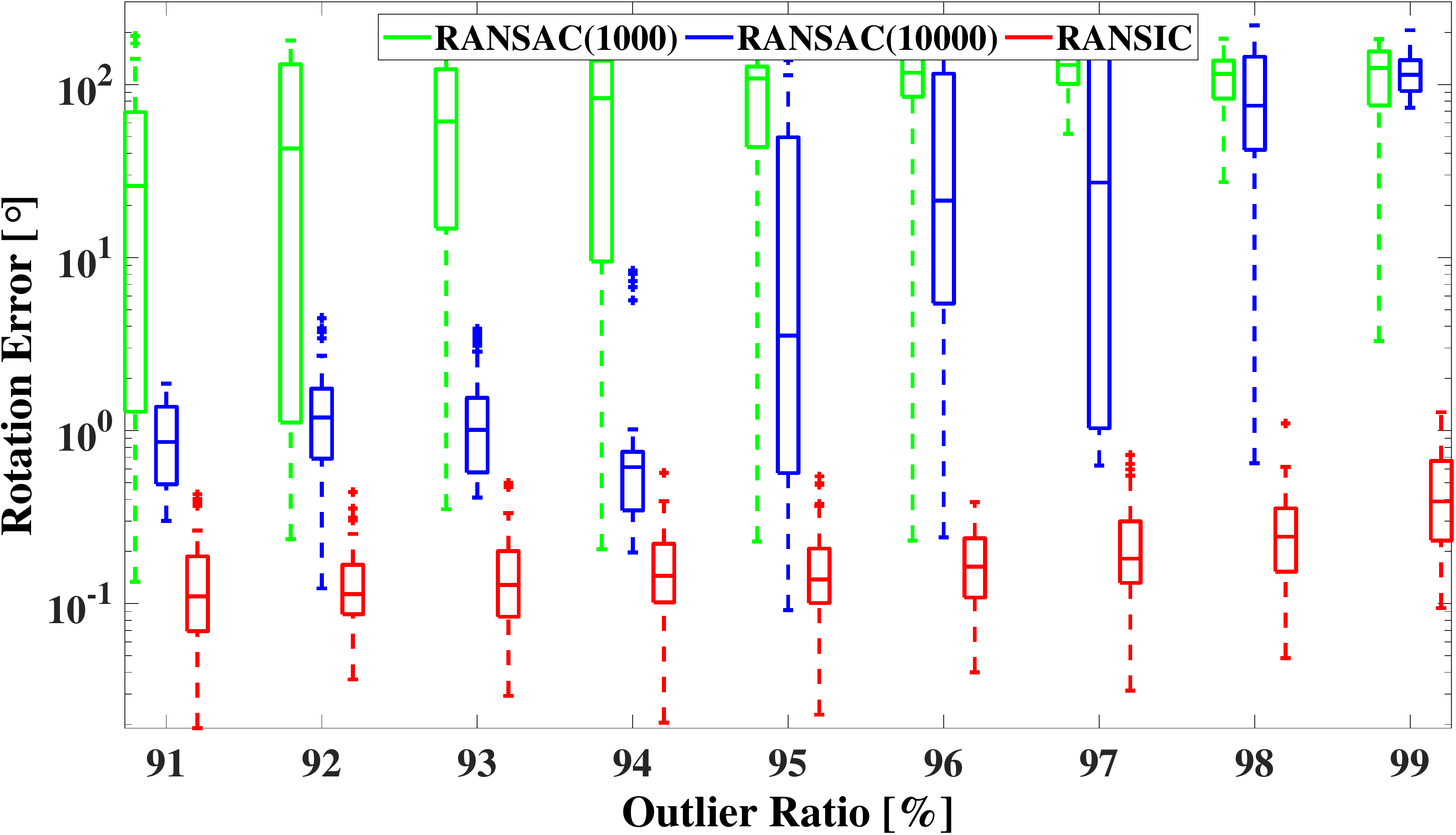}
\end{minipage}
}%

\subfigure{
\begin{minipage}[t]{1\linewidth}
\centering
\includegraphics[width=0.246\linewidth]{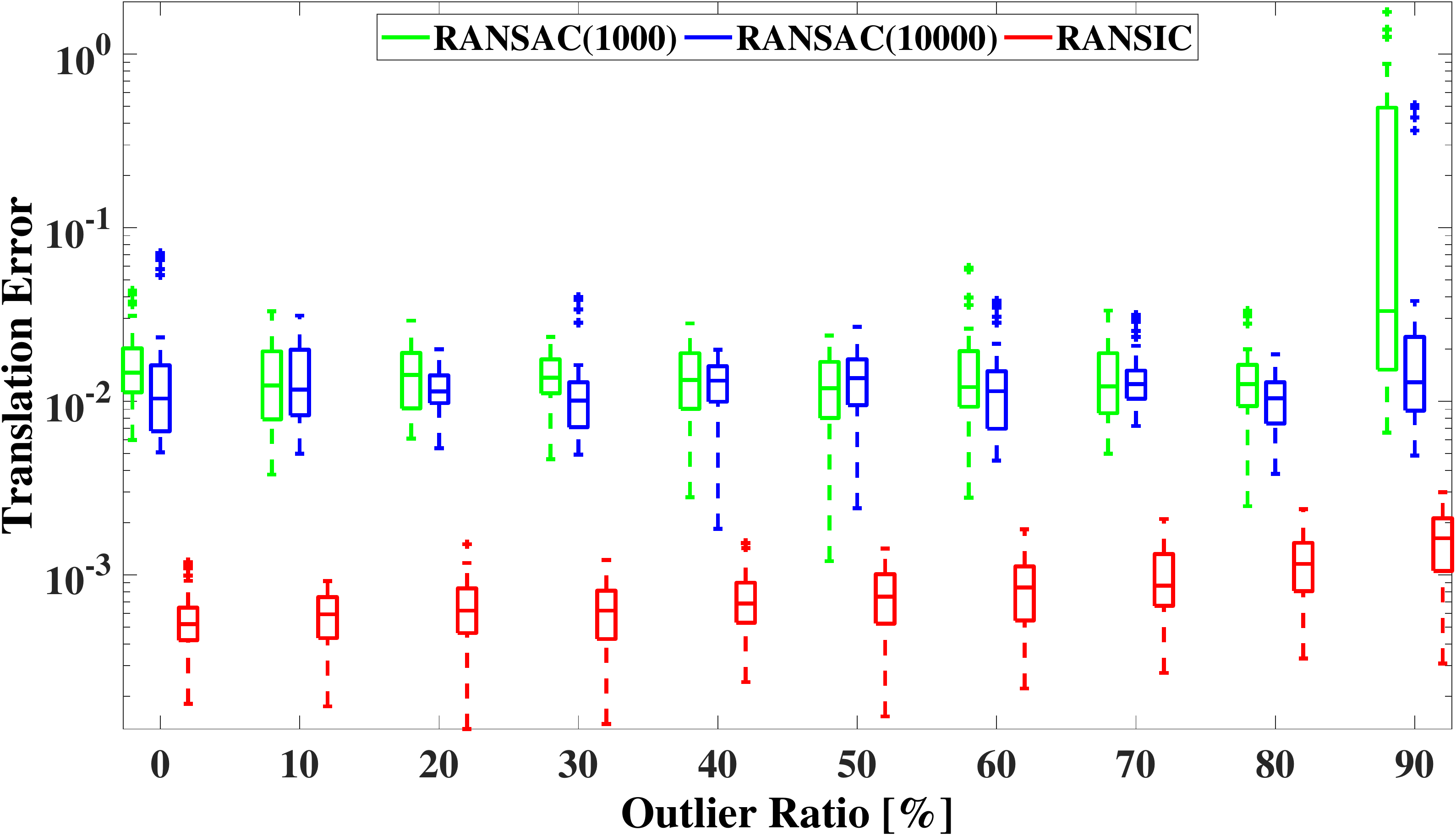}
\includegraphics[width=0.246\linewidth]{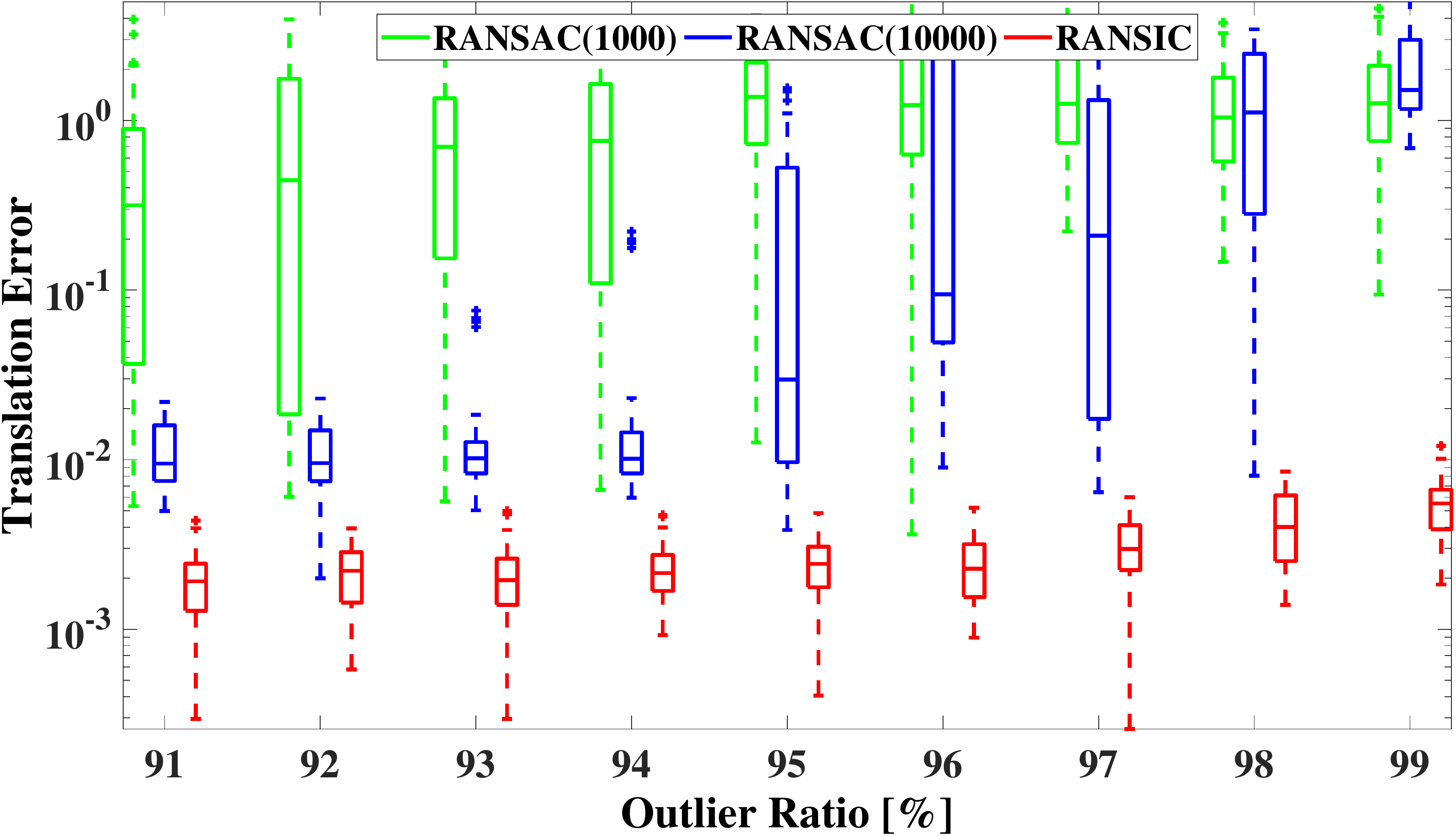}
\includegraphics[width=0.246\linewidth]{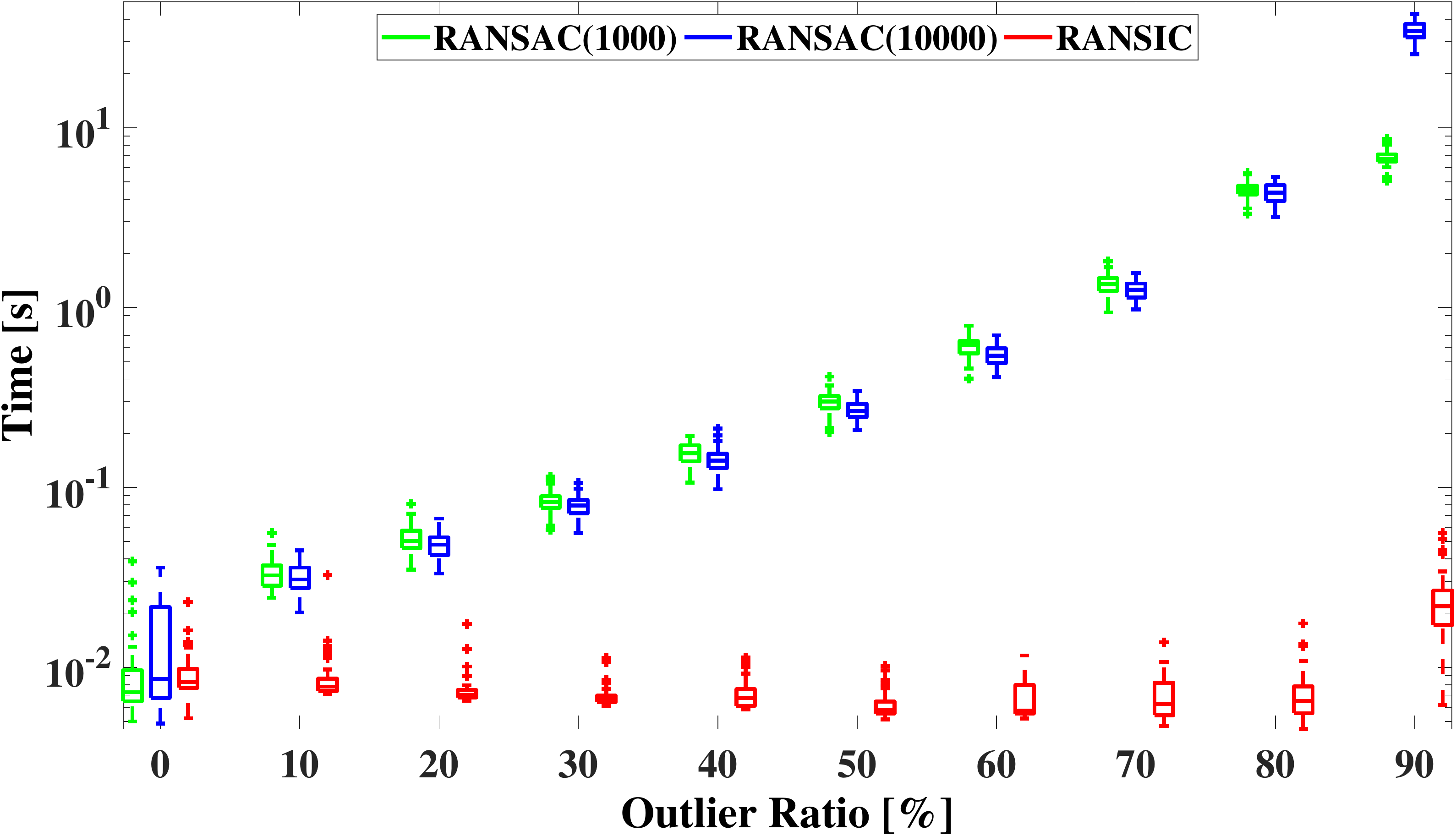}
\includegraphics[width=0.246\linewidth]{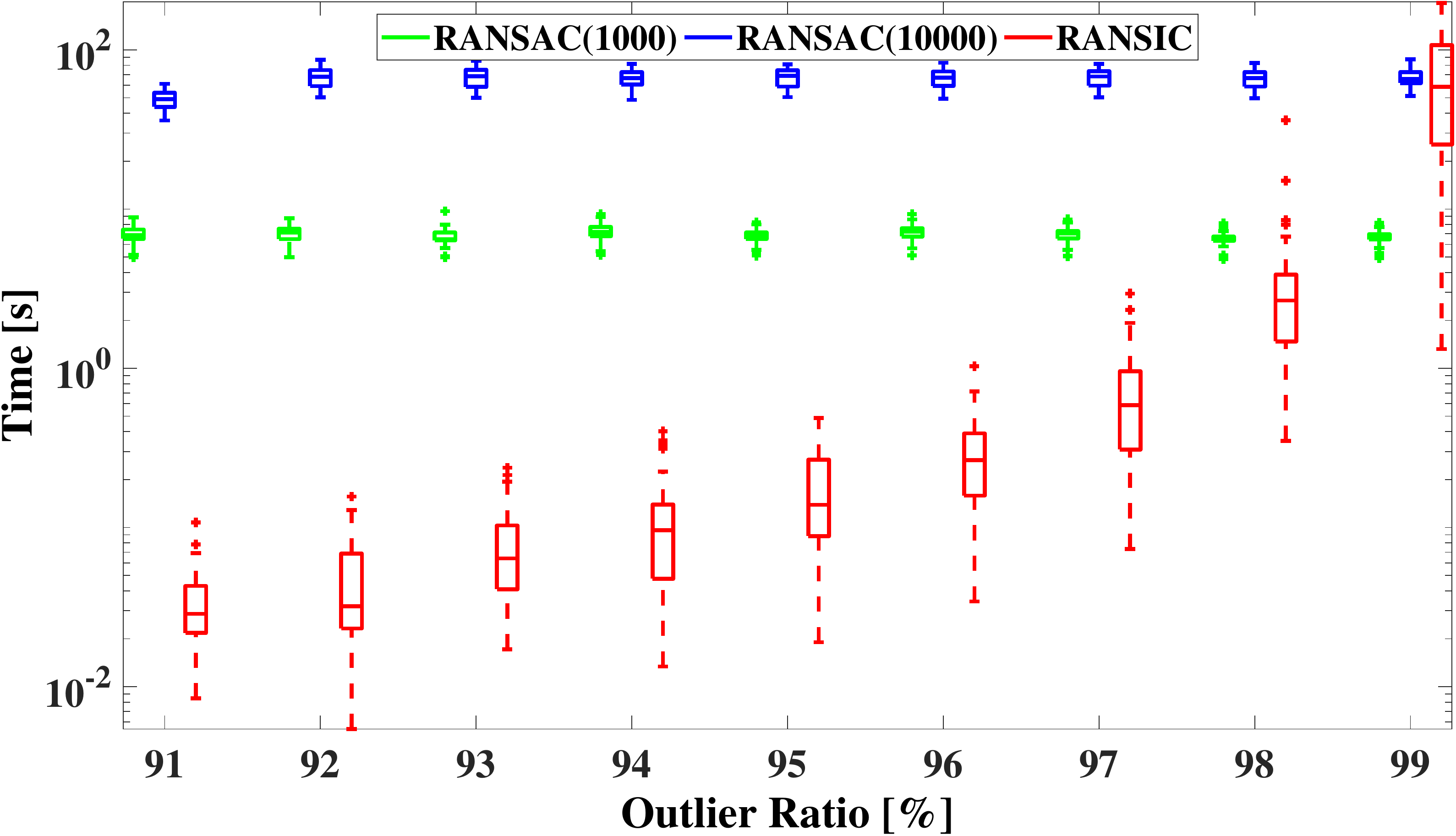}
\end{minipage}
}%

\vspace{-3mm}
\centering
\caption{Experimental results on point cloud registration over the `bunny' point cloud. Examples of a known-scale and an unknown-scale point cloud registration problem with 95\% outliers are shown in the top-left images. Parameter setup: $\gamma=10^{\circ}$, $\upsilon=3.2$, $\tau=9$, and for $itr=1$, $\alpha=3.6\sigma$, $\beta=5.4\sigma$, while for $itr=2$, $\alpha=3.2\sigma$, $\beta=4.8\sigma$. (a) Registration results with known scale. (b) Registration results with unknown scale.}
\label{Syn-PCR}
\end{figure*}

\subsection{Standard Experiments on Point Cloud Registration}
\label{Ex-B}

\begin{figure*}[t]
\centering

\subfigure[Scale Estimation]{
\begin{minipage}[t]{0.33\linewidth}
\centering
\includegraphics[width=0.49\linewidth]{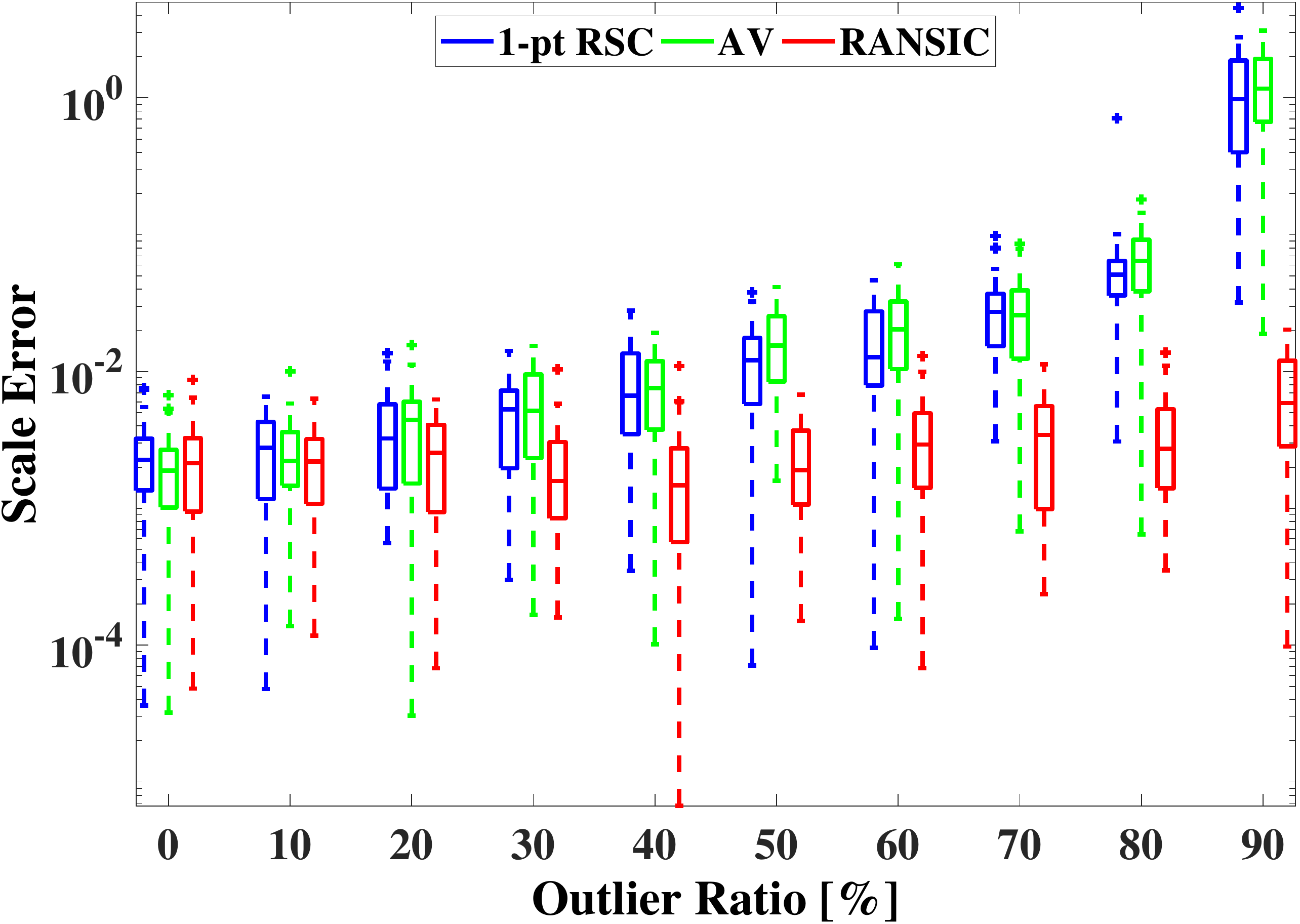}
\includegraphics[width=0.49\linewidth]{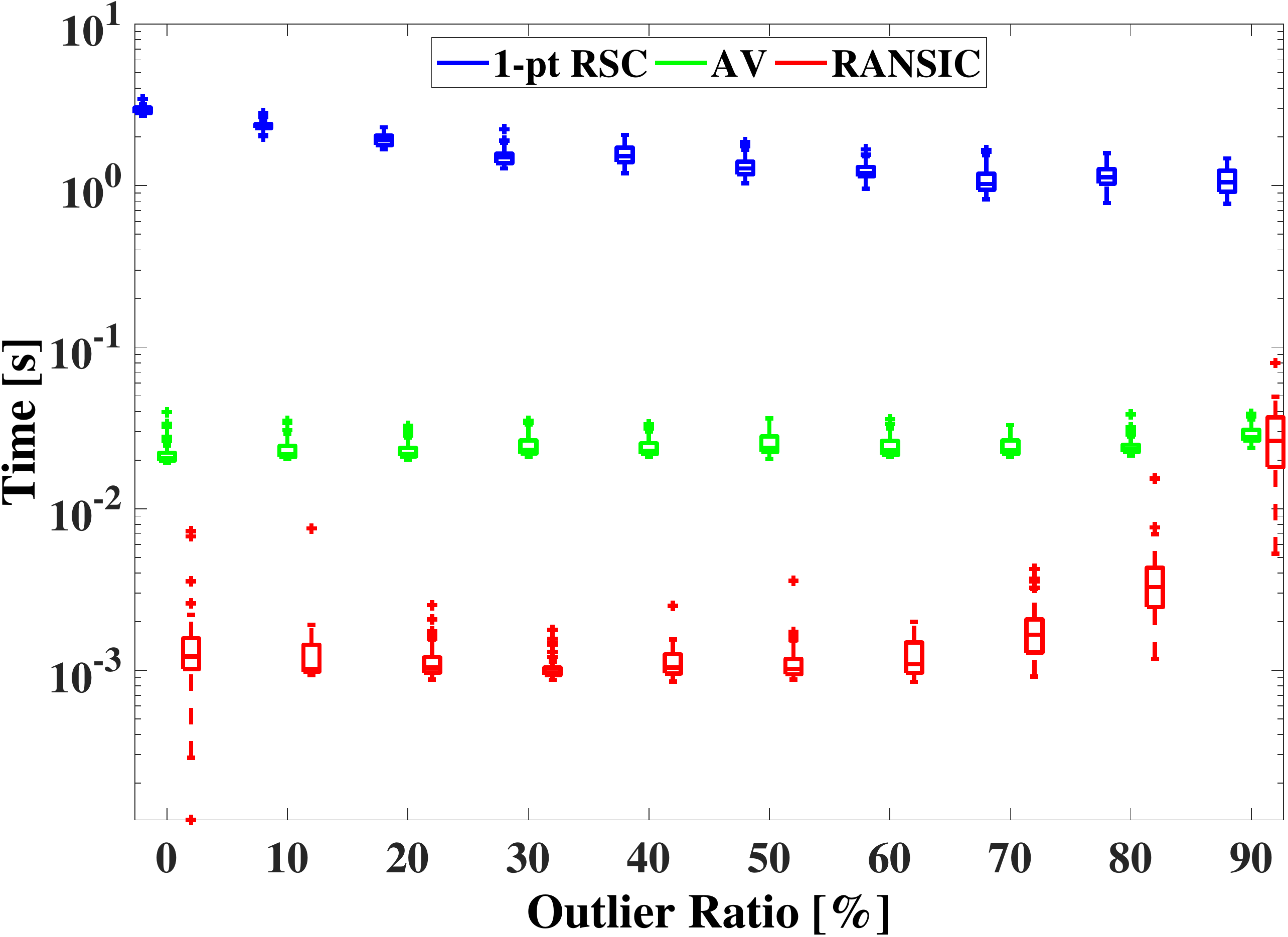}
\end{minipage}
}%
\subfigure[High Noise]{
\begin{minipage}[t]{0.33\linewidth}
\centering
\includegraphics[width=0.49\linewidth]{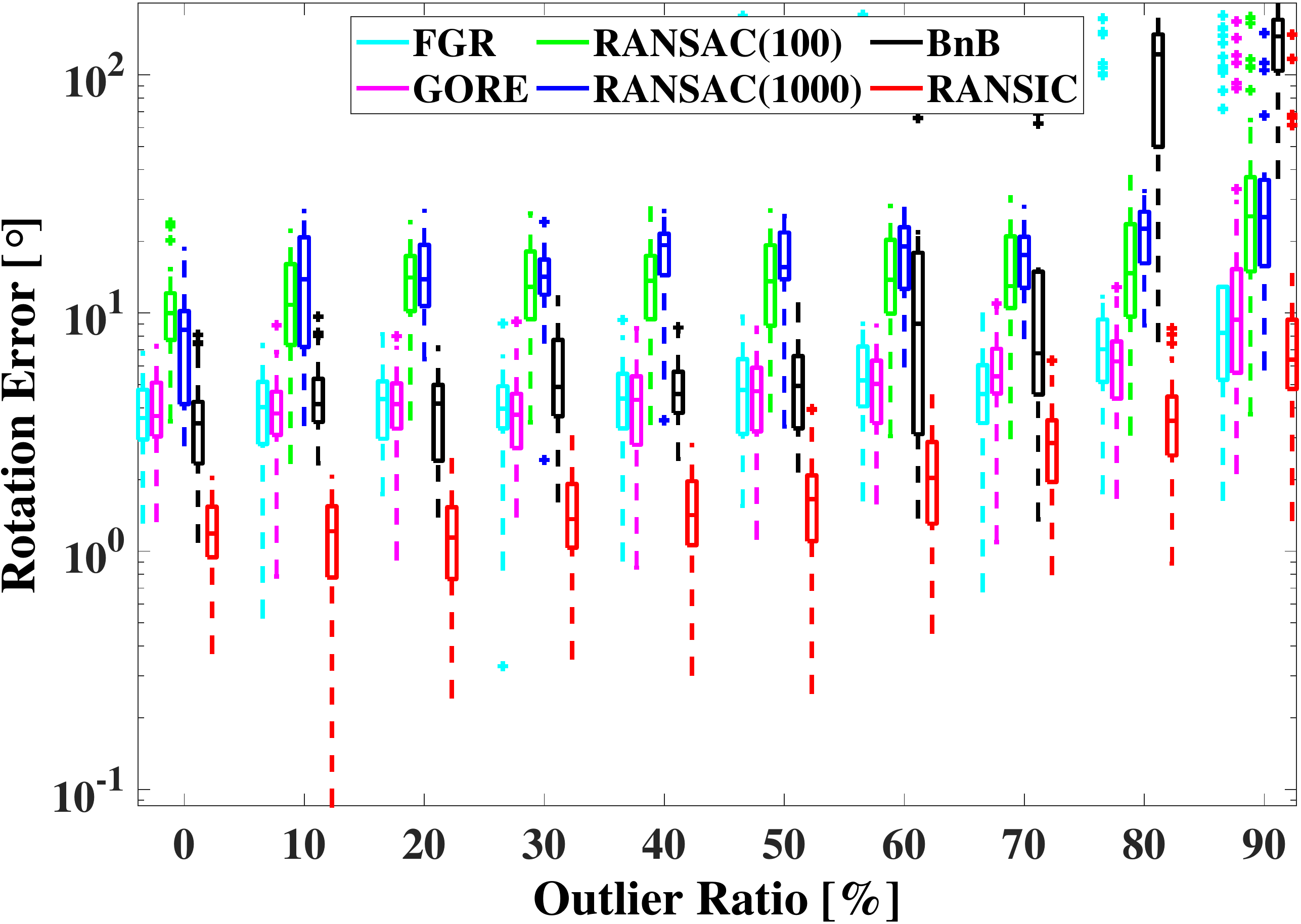}
\includegraphics[width=0.49\linewidth]{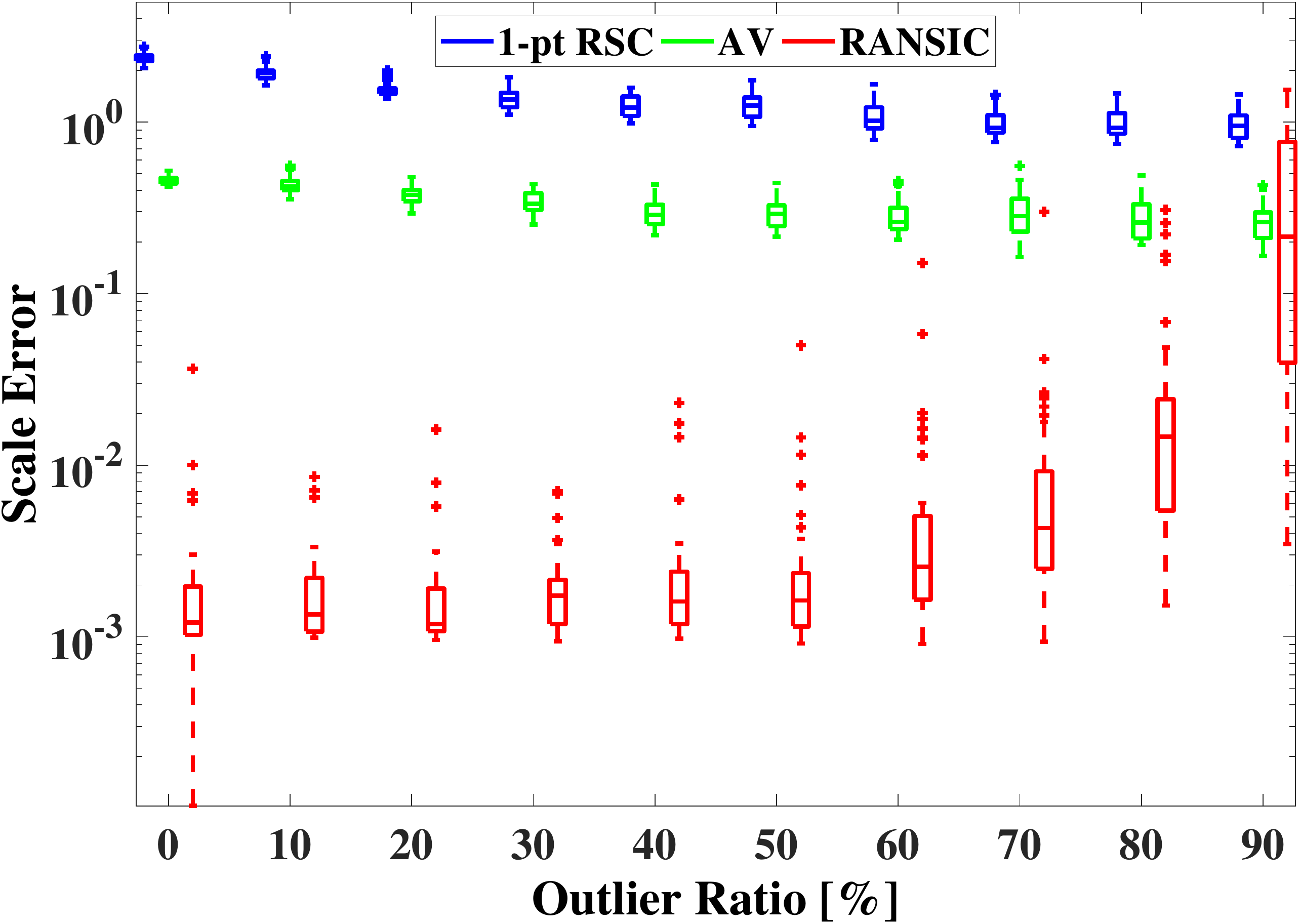}
\end{minipage}
}%
\subfigure[Inlier Recall Ratio]{
\begin{minipage}[t]{0.32\linewidth}
\centering
\includegraphics[width=1\linewidth]{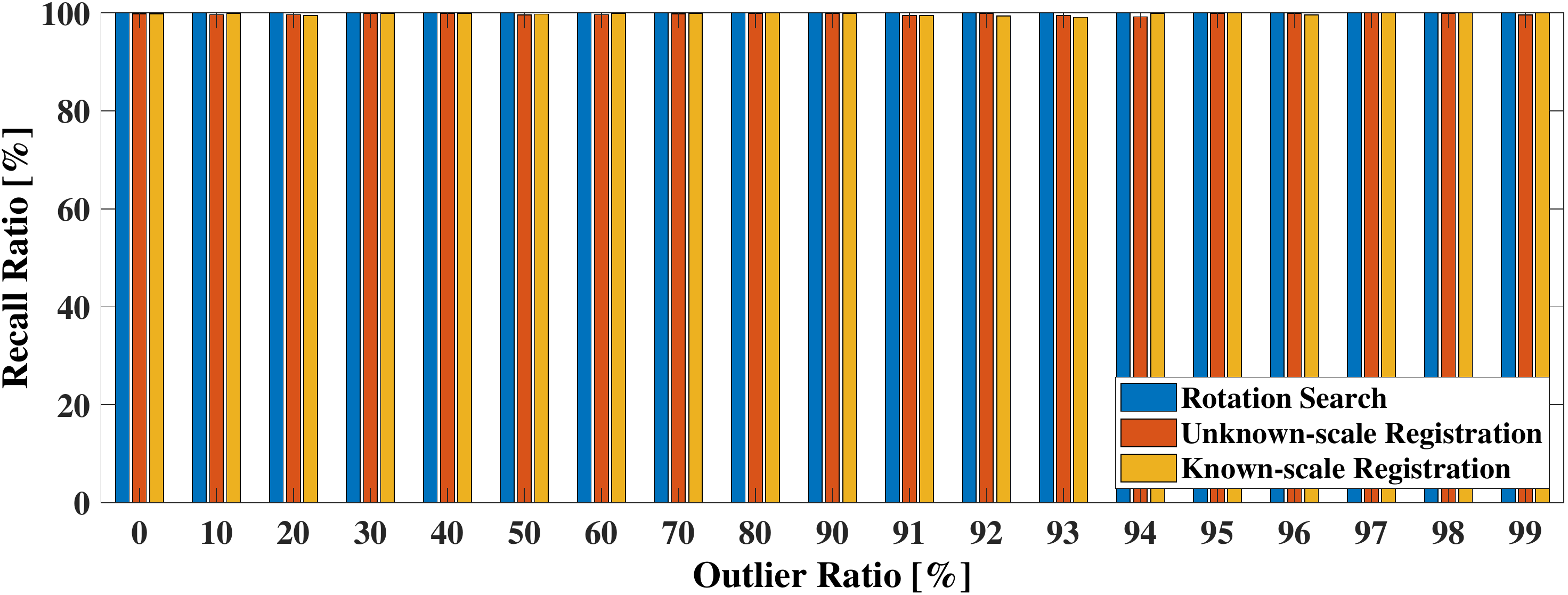}
\end{minipage}
}%

\vspace{-2mm}

\centering
\caption{Additional experimental results. (a) Scale estimation results of RANSIC compared with 1-pt RSC and AV. (b) Comparative results for handling high-noise problems. (c) Inlier recall ratio of RANSIC over rotation search and point cloud registration (unknown and known scale), all with $N=1000$.}
\label{A-Exp}
\end{figure*}

We evaluate RANSIC on point cloud registration, where we use the Stanford `bunny'~\cite{curless1996volumetric} as the point cloud. We downsample the bunny to $1000$ points  and resize it to fit in a $[-0.5,0.5]^3$ box as our first point set $\mathcal{P}=\{\mathbf{p}_i\}_{i=1}^{1000}$. Then we transform $\mathcal{P}$ with a random transformation ($\mathit{s}, \boldsymbol{R},\boldsymbol{t}$), where $\mathit{s}\in(1,5)$, $\boldsymbol{R}\in SO3$ and $||\boldsymbol{t}||\leq 3$. We add random zero-mean Gaussian noise with $\sigma=0.01$ to the transformed point set and take it as our second point set $\mathcal{Q}=\{\mathbf{q}_i\}_{i=1}^{1000}$. To create outliers in a cluttered pattern, a portion of the points in $\mathcal{Q}$ (0-99\%) are replaced by random points inside a 3D sphere of diameter $\mathit{s}\sqrt{3}$, as exemplified in Fig.~\ref{Syn-PCR}. All the results are obtained over 50 Monte Carlo runs.

For the benchmark on known-scale registration, we adopt FGR, GORE~\cite{bustos2017guaranteed}, RANSAC(1000) and RANSAC(10000), and TEASER~\cite{yang2019polynomial,yang2020teaser} (also as ROBIN's framework). TEASER is implemented as follows: (i) we solve the robust SDP with the GNC heuristic as in TEASER++~\cite{yang2020teaser}, (ii) we do not use parallelism programming to accelerate the maximal clique solver in order to guarantee fairness for the comparison of all solvers, and (iii) it is implemented in Matlab, not C++, with the fast maximal clique solver~\cite{eppstein2010listing}. For unknown-scale registration, only the two RANSAC solvers are adopted since (i) FGR and GORE are not designed for scale estimation, and (ii) scale estimator Adaptive Voting (AV) in TEASER runs in hours with $N=1000$ correspondences and fails to tolerate 90\% outliers, shown in Fig.~\ref{A-Exp}(a). The errors of $\mathit{s}$ and $\boldsymbol{t}$ are computed as: $|\mathit{s}^{\star}-\mathit{s}|$ and $||\boldsymbol{t}^{\star}-\boldsymbol{t}||$.

From Fig.~\ref{Syn-PCR}, we find that: (i) in known-scale registration, RANSAC(1000) yields wrong results at over 90\% and both RANSAC(10000) and FGR break at around 95\%, while both RANSIC and TEASER are robust against 99\%, (ii) in unknown-scale registration, RANSIC can still tolerate 99\% outliers, and (iii) RANSIC is the fastest when the outlier ratio is within 0-96\% and most often has the highest accuracy.

\subsection{Additional Experiments}

We supplement three additional experiments. In Fig.~\ref{A-Exp}(a), we test RANSIC as a scale estimator alone in comparison with AV in TEASER and 1-point RANSAC (1-pt RSC)~\cite{li2021point} with $N=100$. Both AV and 1-pt RSC break at 90\% and are slower than RANSIC, whereas, in fact, RANSIC can tolerate 99\% (Fig.~\ref{Syn-PCR}). Besides, we test RANSIC for rotation search and scale estimation with high noise $\sigma=0.1$ and $N=100$ in Fig.~\ref{A-Exp}(b). (Based on RANSIC's mechanism, as long as the scale is correctly estimated, meaning that the true inliers are found, optimal rotation and translation can be solved properly too.) RANSIC can tolerate 80\% outliers even with such high noise while most other solvers fail. Also, we evaluate the inlier recall ratio of RANSIC over three problems ($\sigma=0.01$) in Fig.~\ref{A-Exp}(c). RANSIC can recall approximately 100\% inliers.

\subsection{Real Application: Image Stitching}

To test RANSIC in real applications of rotation search, we use RANSIC to address the image stitching problem over the \textit{Lunch Room} of the PASSTA dataset~\cite{meneghetti2015image}. We first use SURF~\cite{bay2006surf} to detect and match the 2D keypoints across overlapped image pairs. Then we use $\mathbf{K}^{-1}$ ($\mathbf{K}$ is the known camera intrinsic matrix) to construct vector correspondences $\mathcal{A}=\{\mathbf{a}_i\}_{i=1}^N$ and $\mathcal{B}=\{\mathbf{b}_i\}_{i=1}^N$ w.r.t. the two camera views. We use Algorithm~\ref{Algo1-RS} to solve the rotation $\boldsymbol{R}^{\star}$ between the two views and compute the homography matrix using $\mathbf{H}=\mathbf{K}\boldsymbol{R}^{\star}\mathbf{K}^{-1}$, which is then used to stitch the images. The raw correspondences matched by SURF, the inliers found by RANSIC and the stitching results are shown in Fig.~\ref{Real-data}(a-b).

\subsection{Real Application: Object Localization}

\begin{figure}[h]
\centering

\footnotesize{(a) \textit{Image Stitching: 09\&15, 15\&22}}

\subfigure{
\begin{minipage}[t]{0.49\linewidth}
\centering
\includegraphics[width=0.38\linewidth]{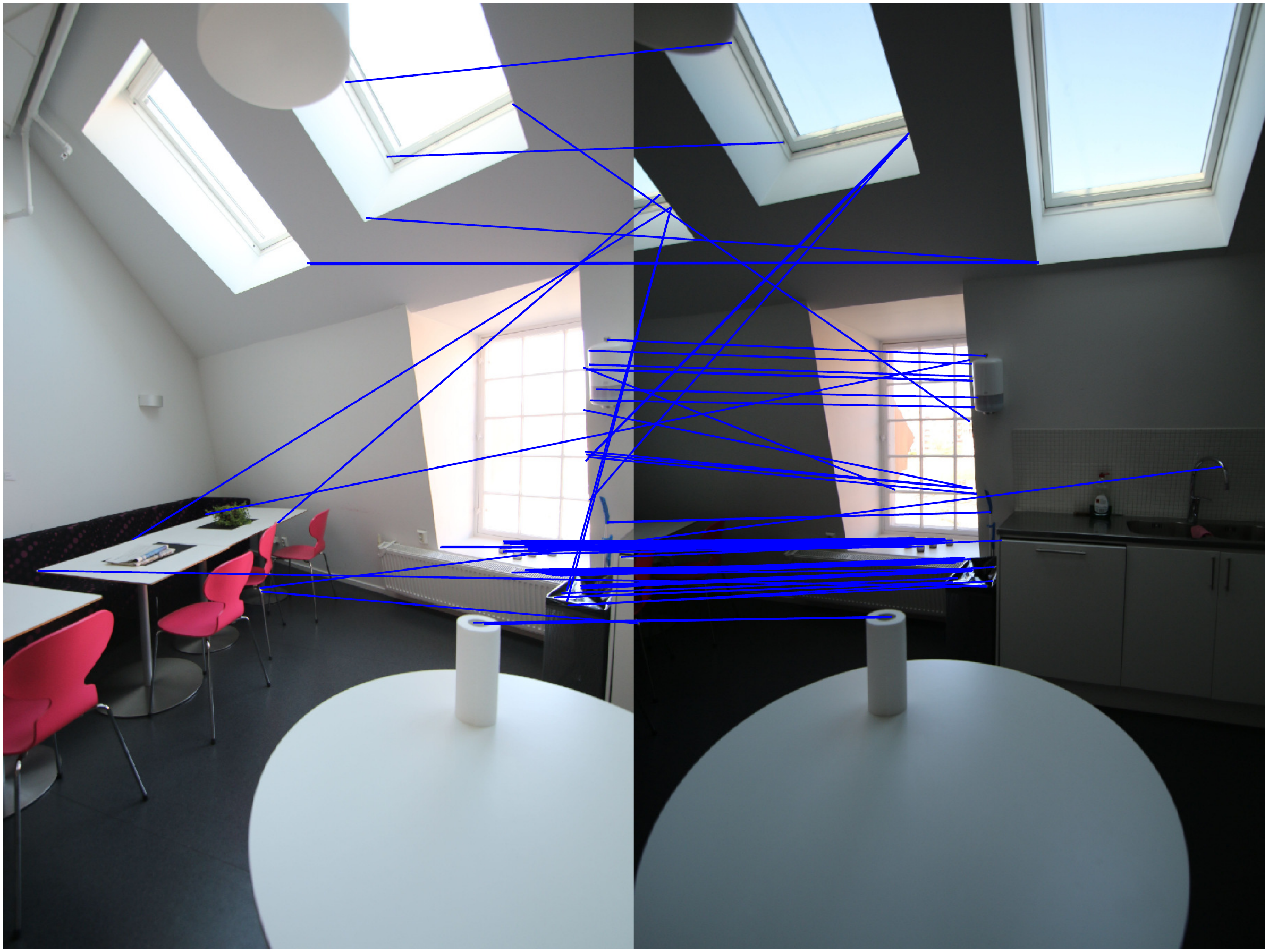}
\includegraphics[width=0.38\linewidth]{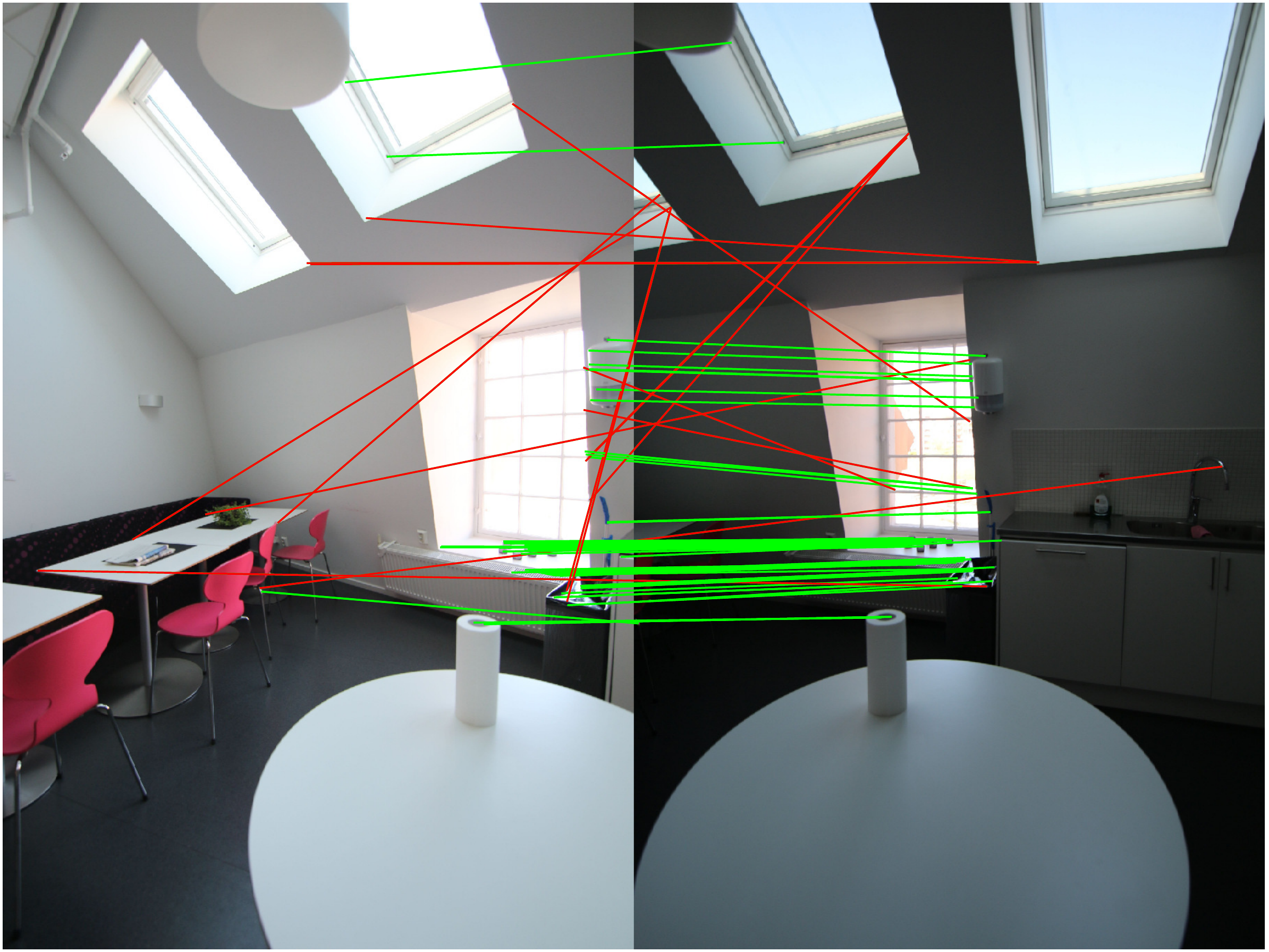}
\includegraphics[width=0.19\linewidth]{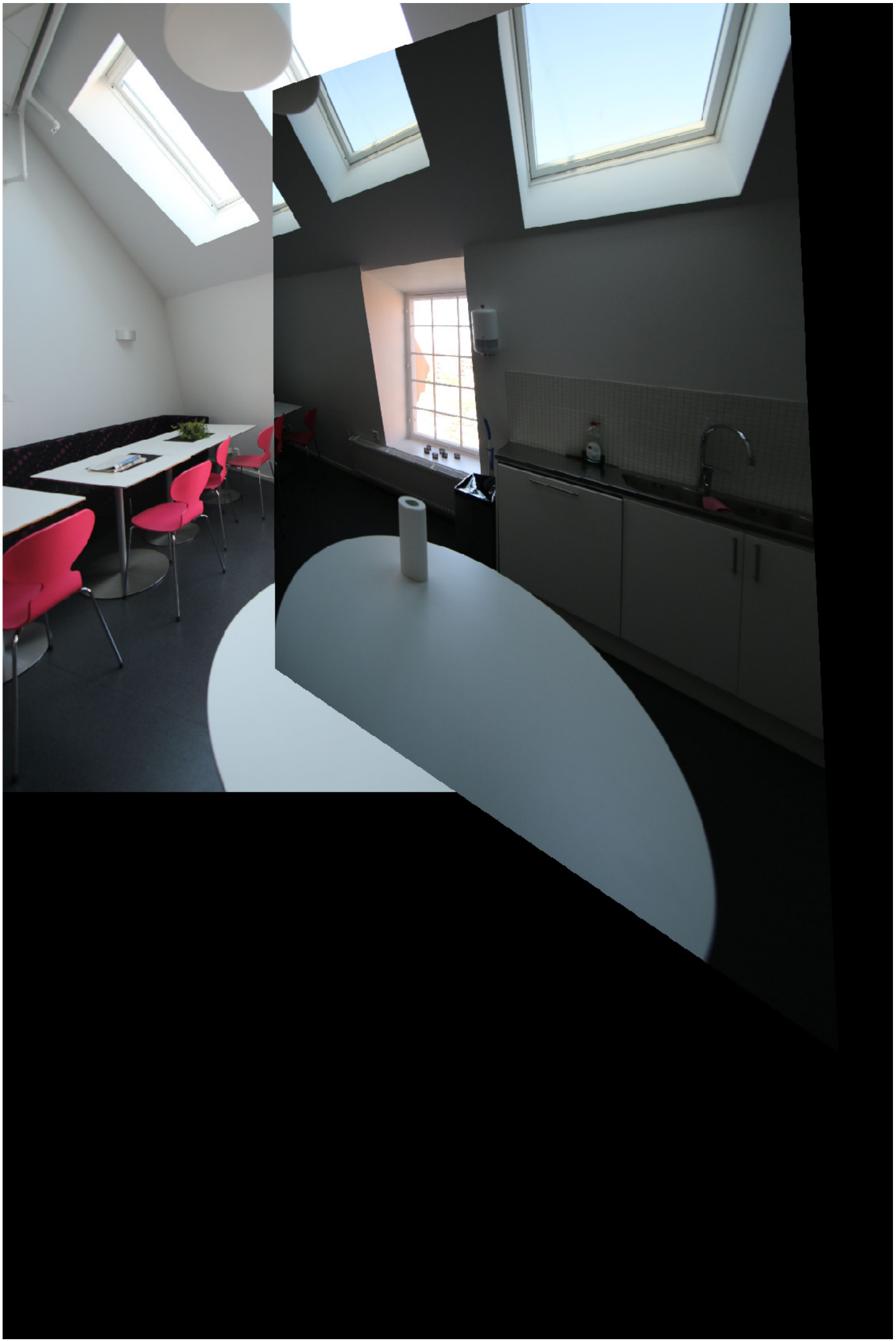}
\end{minipage}
\begin{minipage}[t]{0.49\linewidth}
\centering
\includegraphics[width=0.38\linewidth]{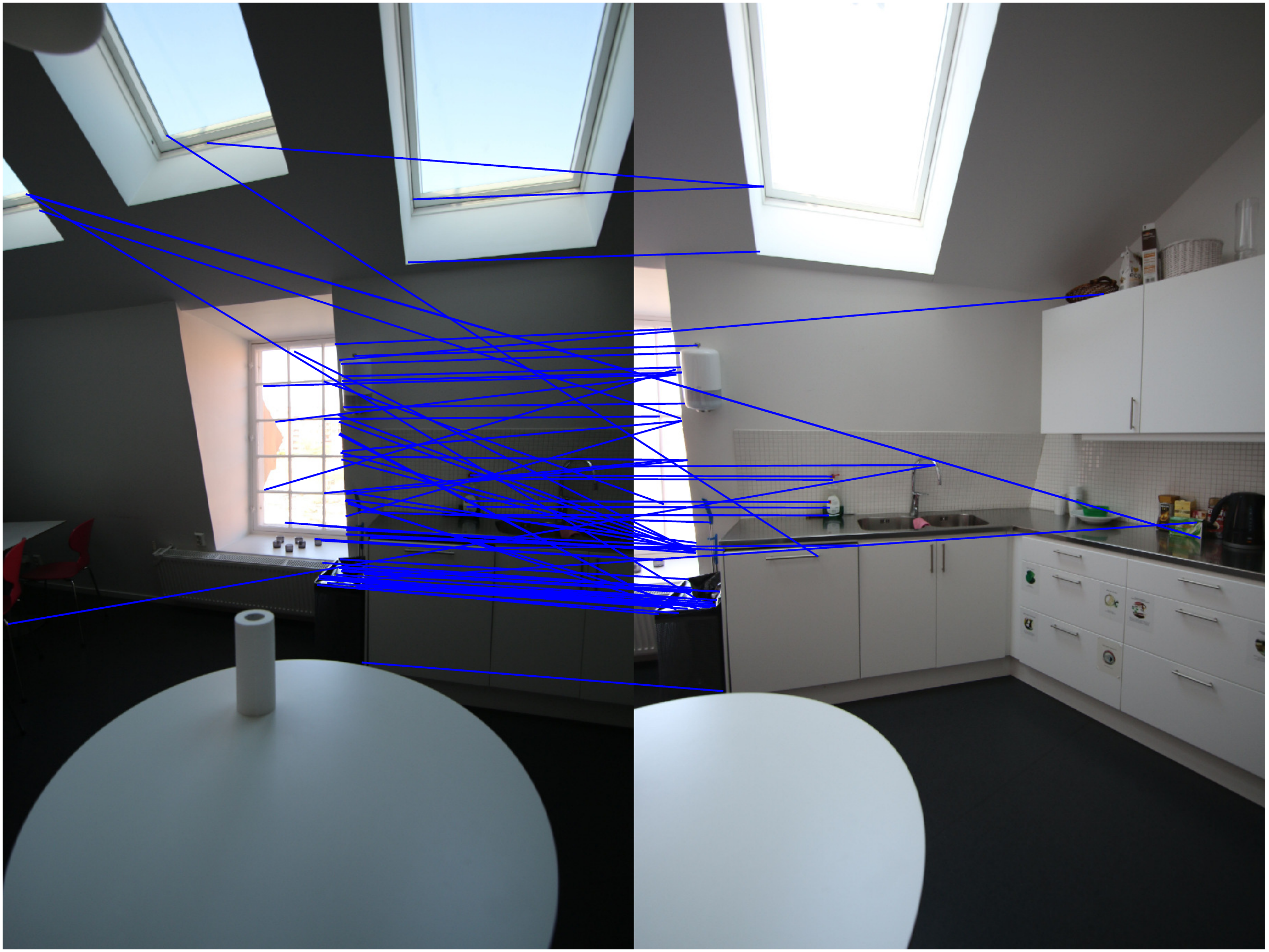}
\includegraphics[width=0.38\linewidth]{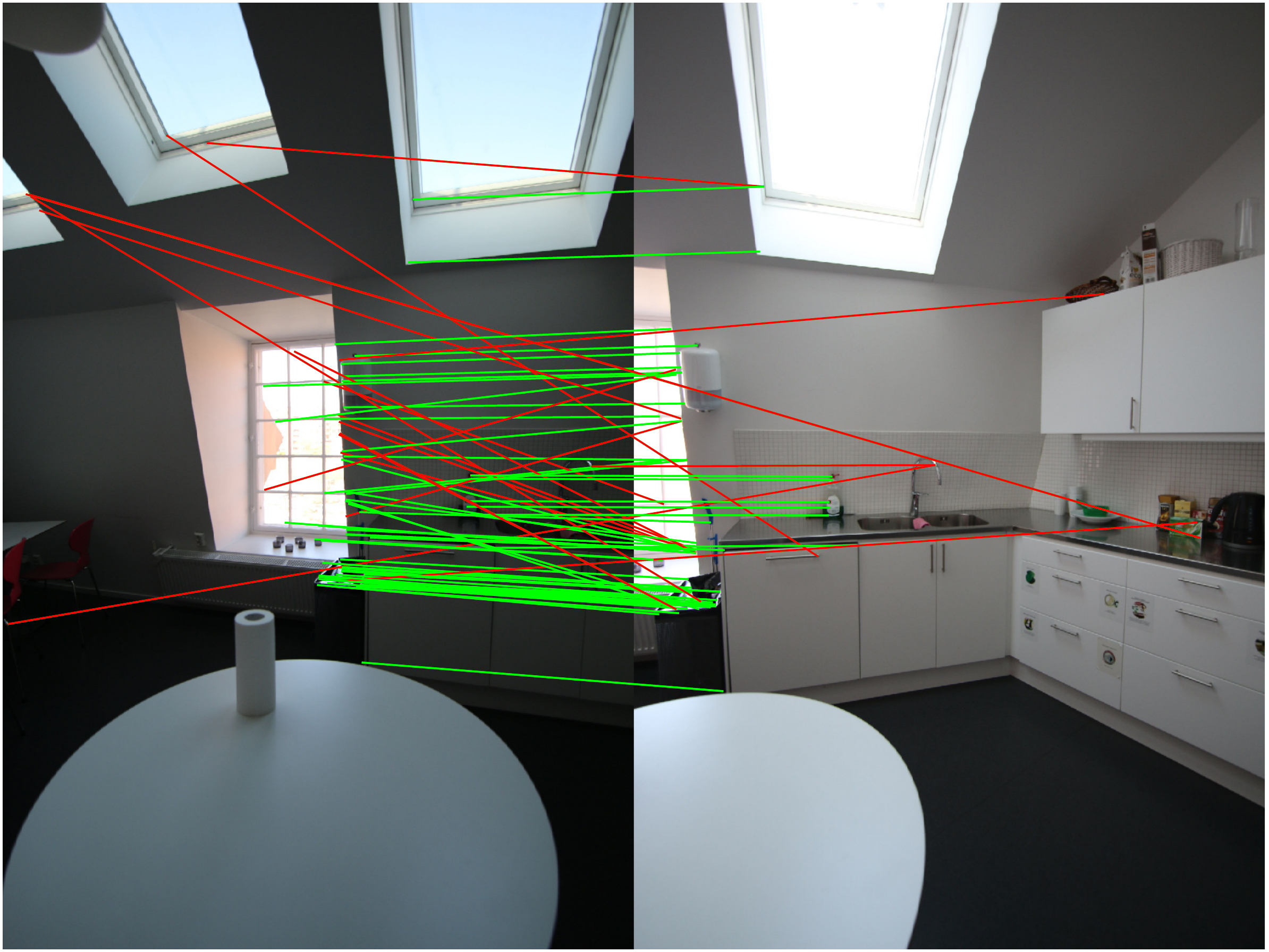}
\includegraphics[width=0.19\linewidth]{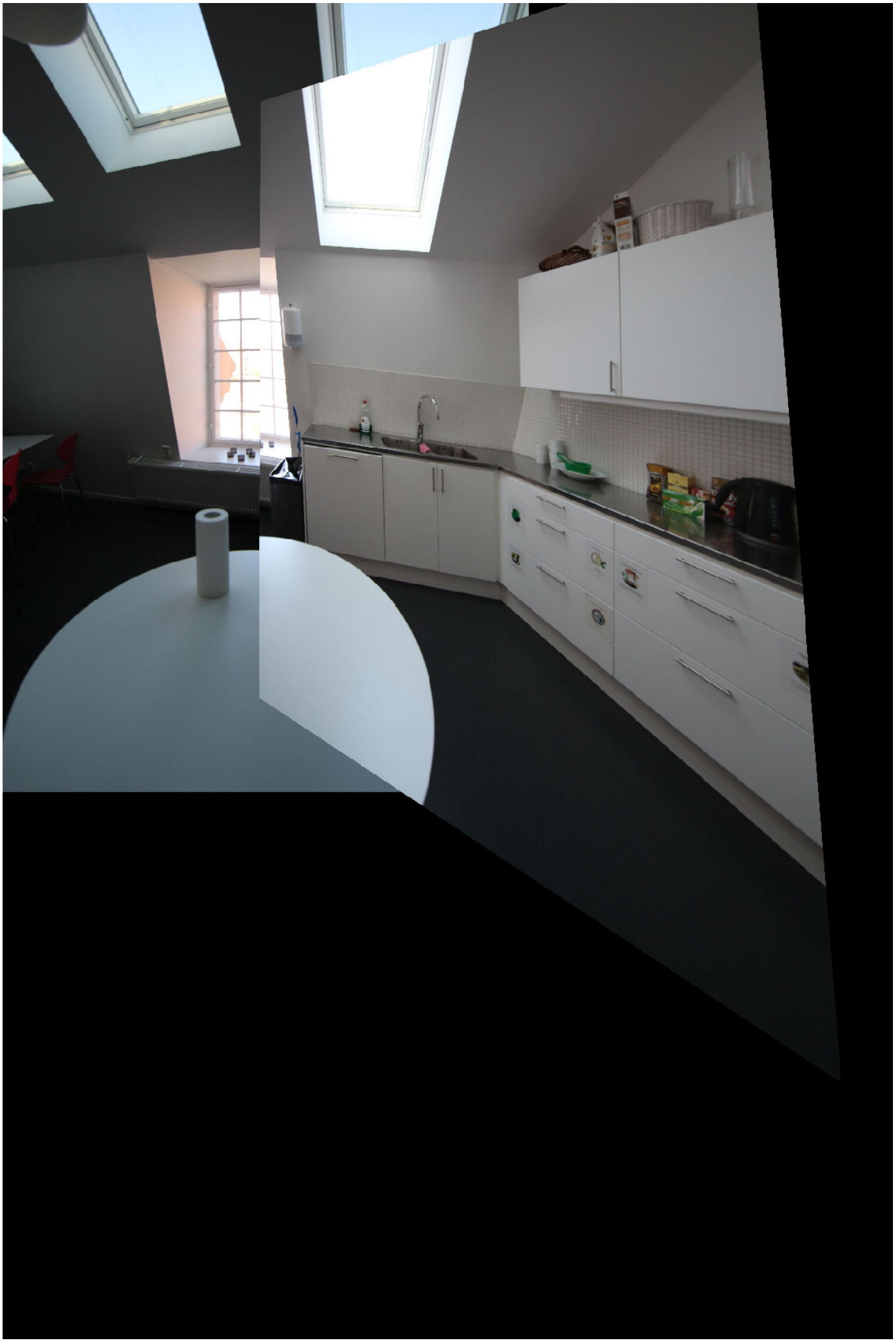}
\end{minipage}
}%

\footnotesize{(b) \textit{Image Stitching: 22\&30, 30\&41}}

\subfigure{
\begin{minipage}[t]{0.492\linewidth}
\centering
\includegraphics[width=0.38\linewidth]{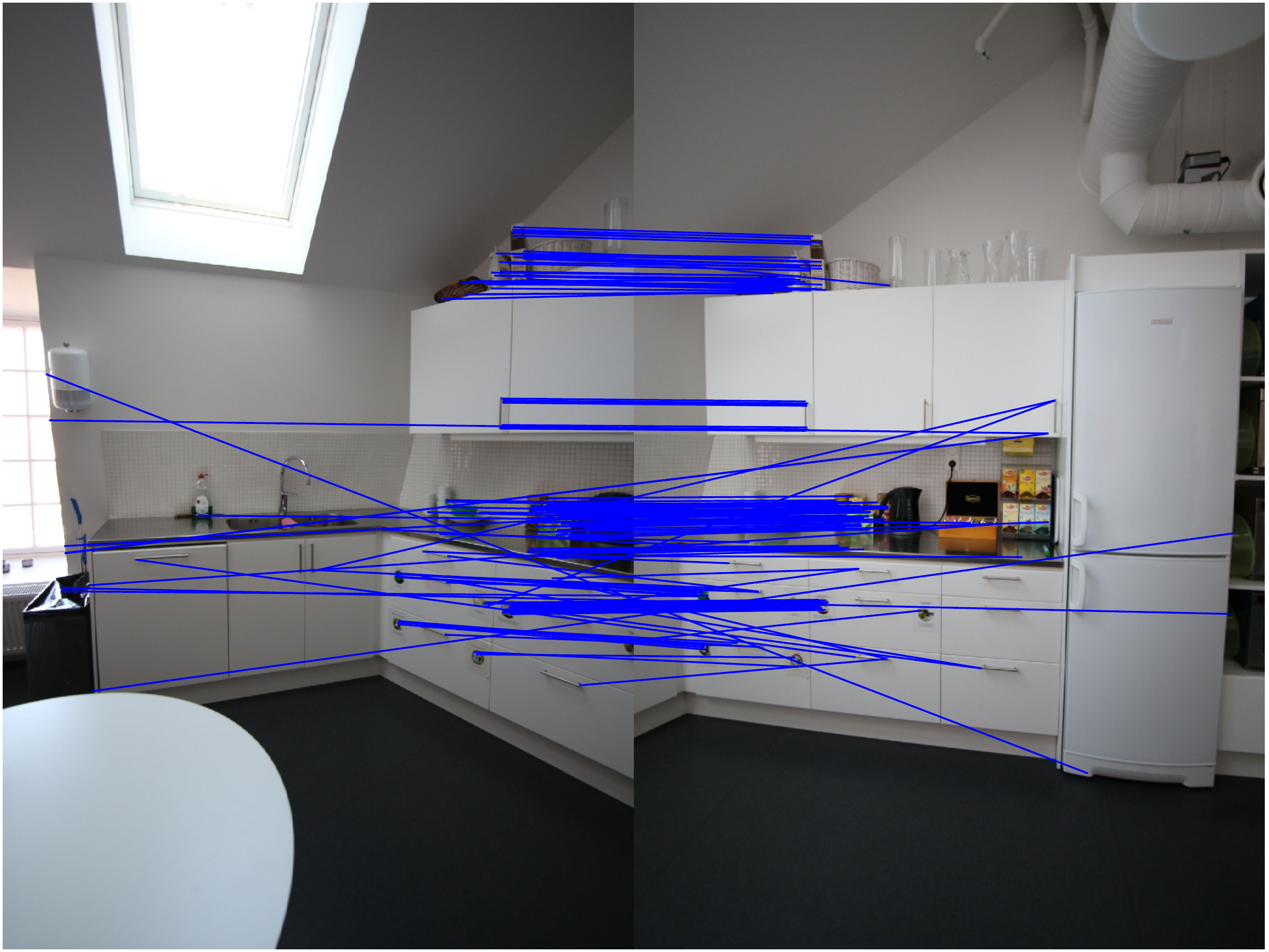}
\includegraphics[width=0.38\linewidth]{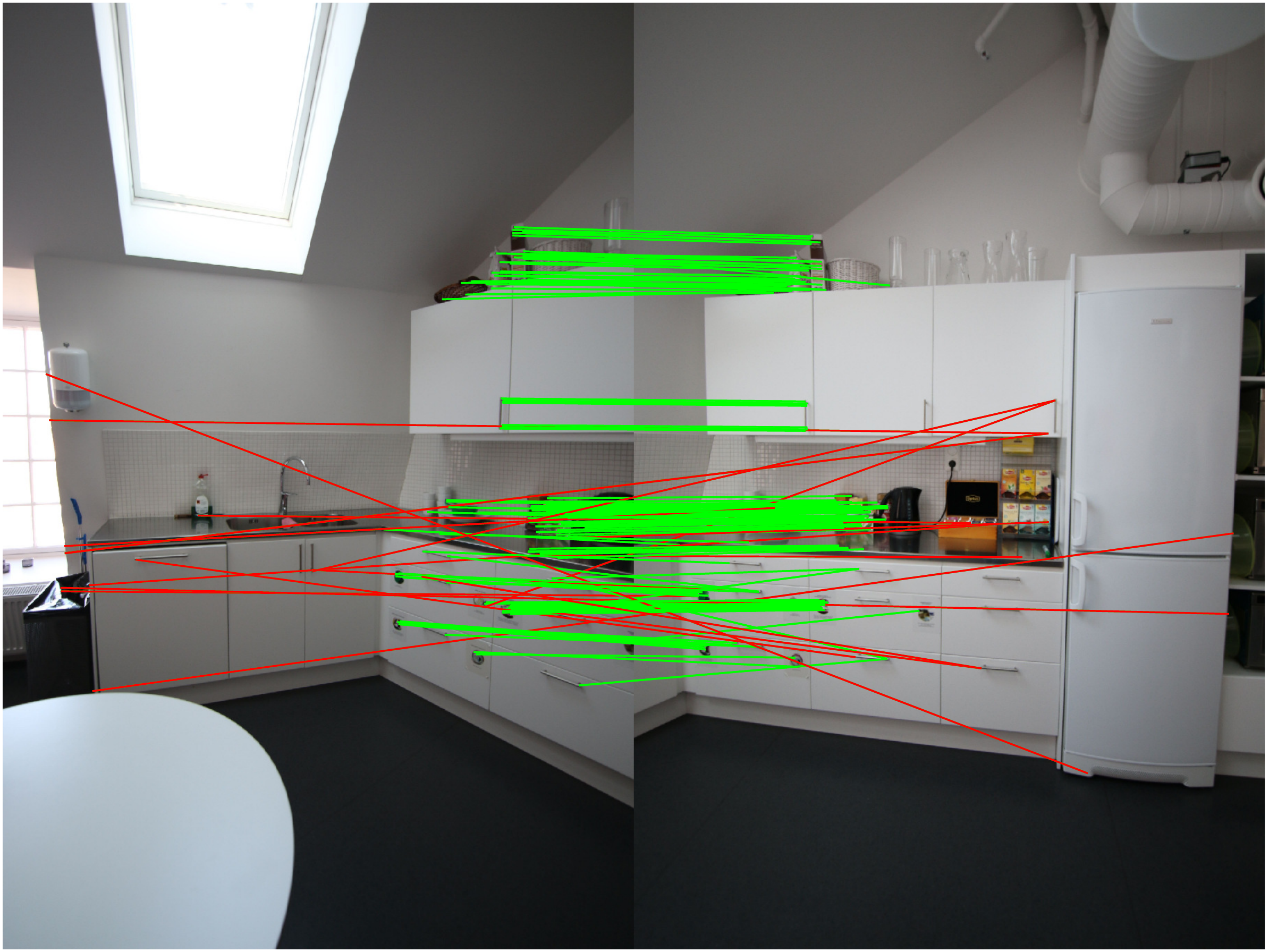}
\includegraphics[width=0.19\linewidth]{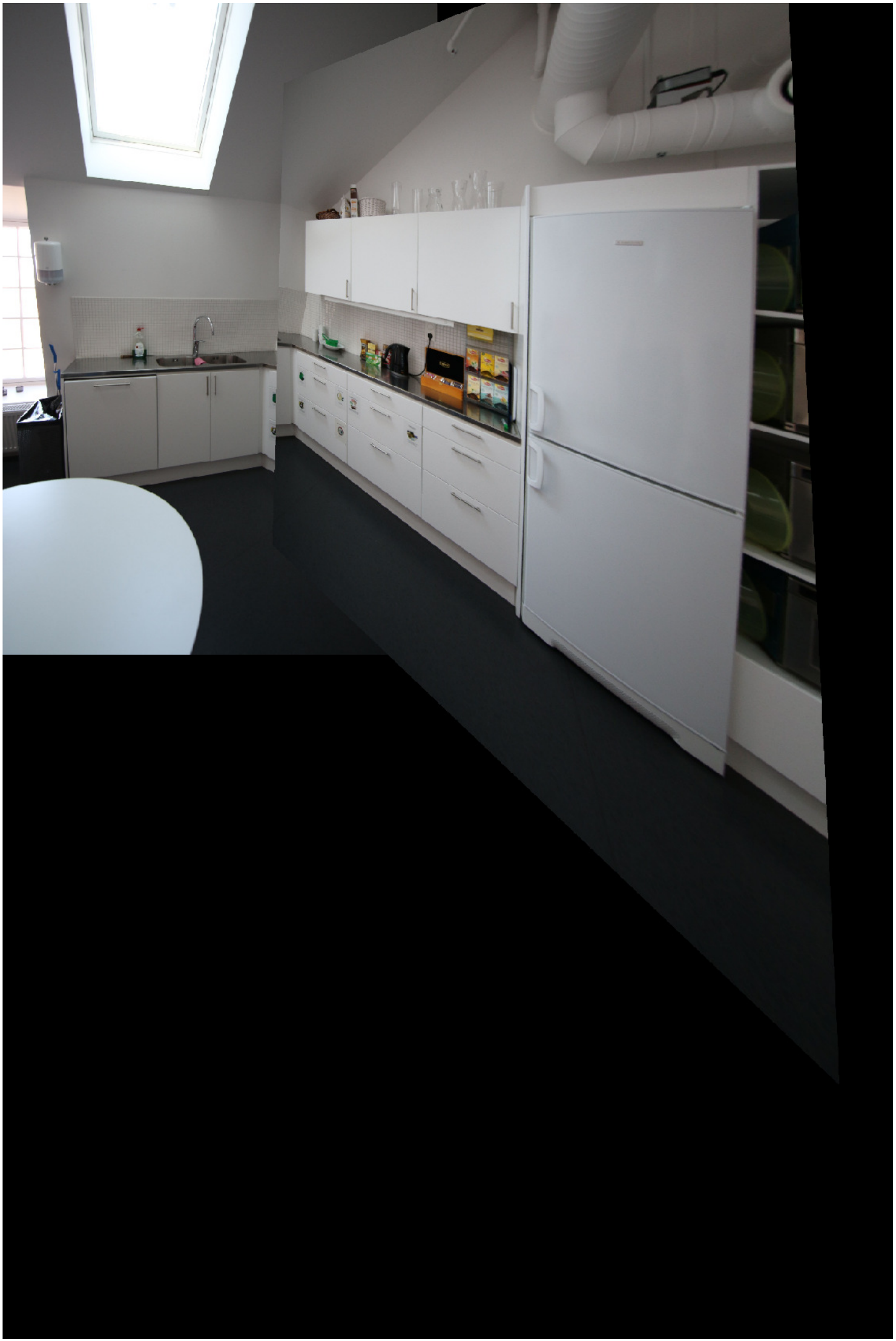}
\end{minipage}
\begin{minipage}[t]{0.492\linewidth}
\centering
\includegraphics[width=0.38\linewidth]{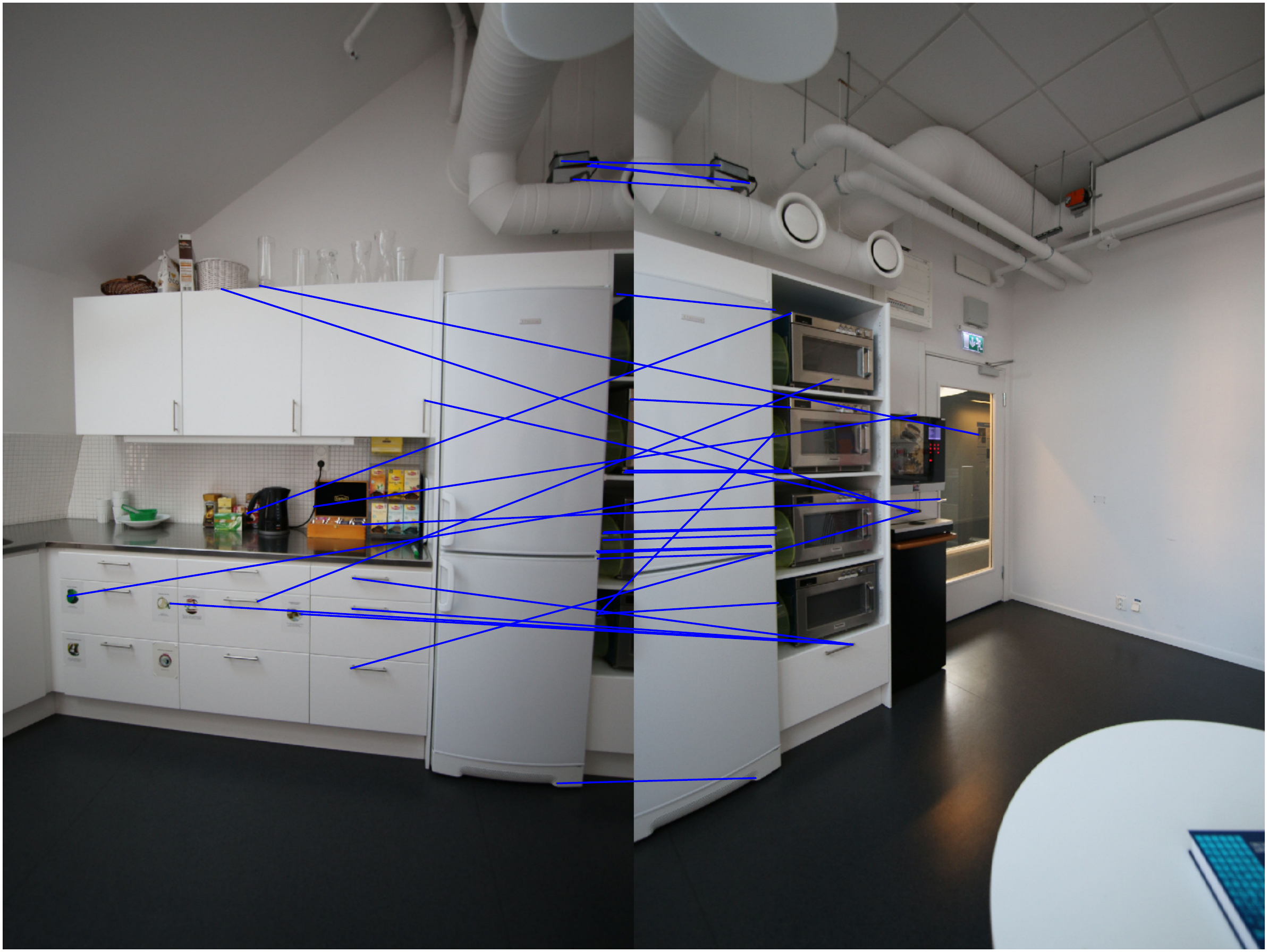}
\includegraphics[width=0.38\linewidth]{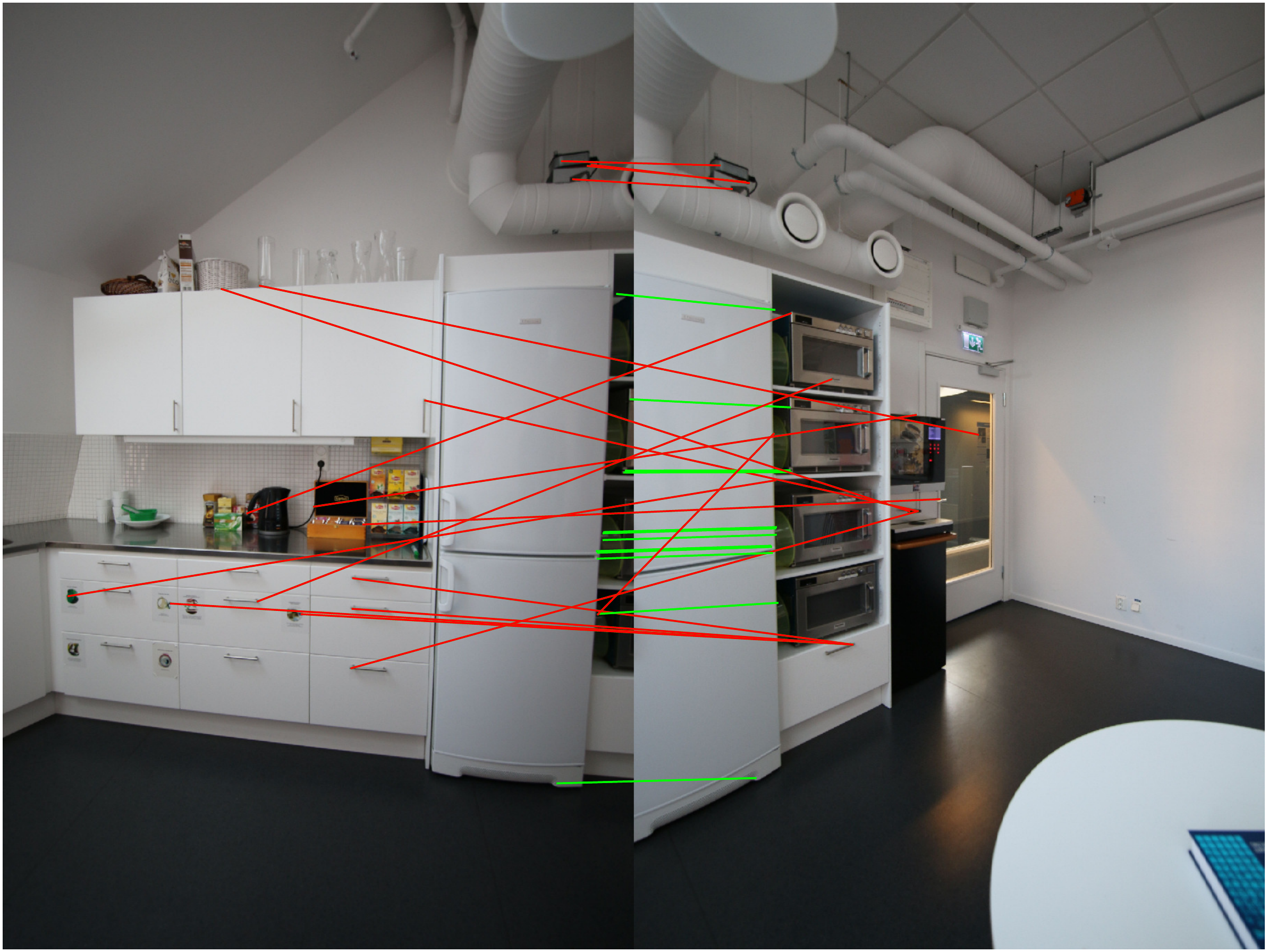}
\includegraphics[width=0.19\linewidth]{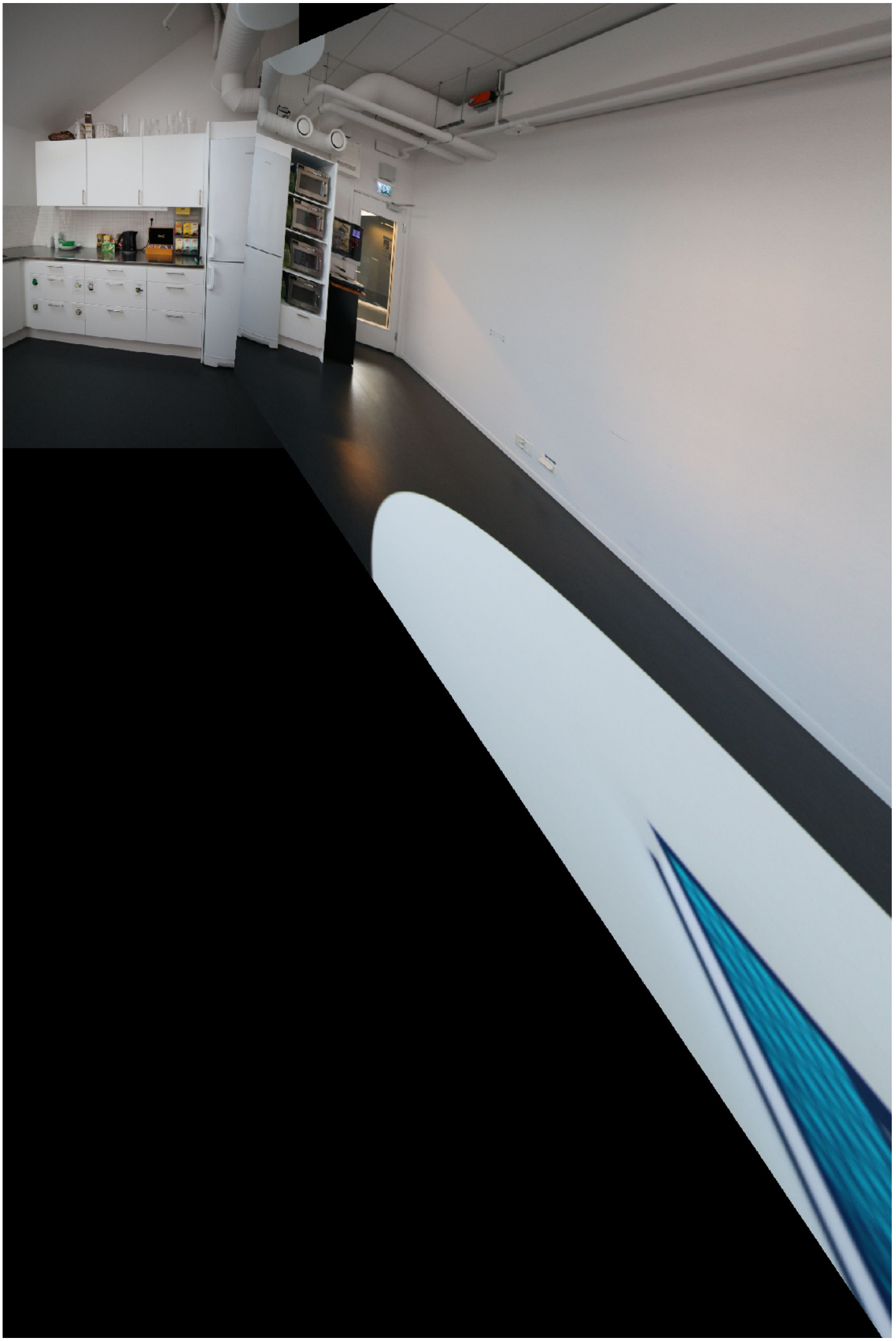}
\end{minipage}
}%

\footnotesize{(c) \textit{Object Localization: Scene-02}}

\subfigure{
\begin{minipage}[t]{0.49\linewidth}
\centering
\includegraphics[width=0.485\linewidth]{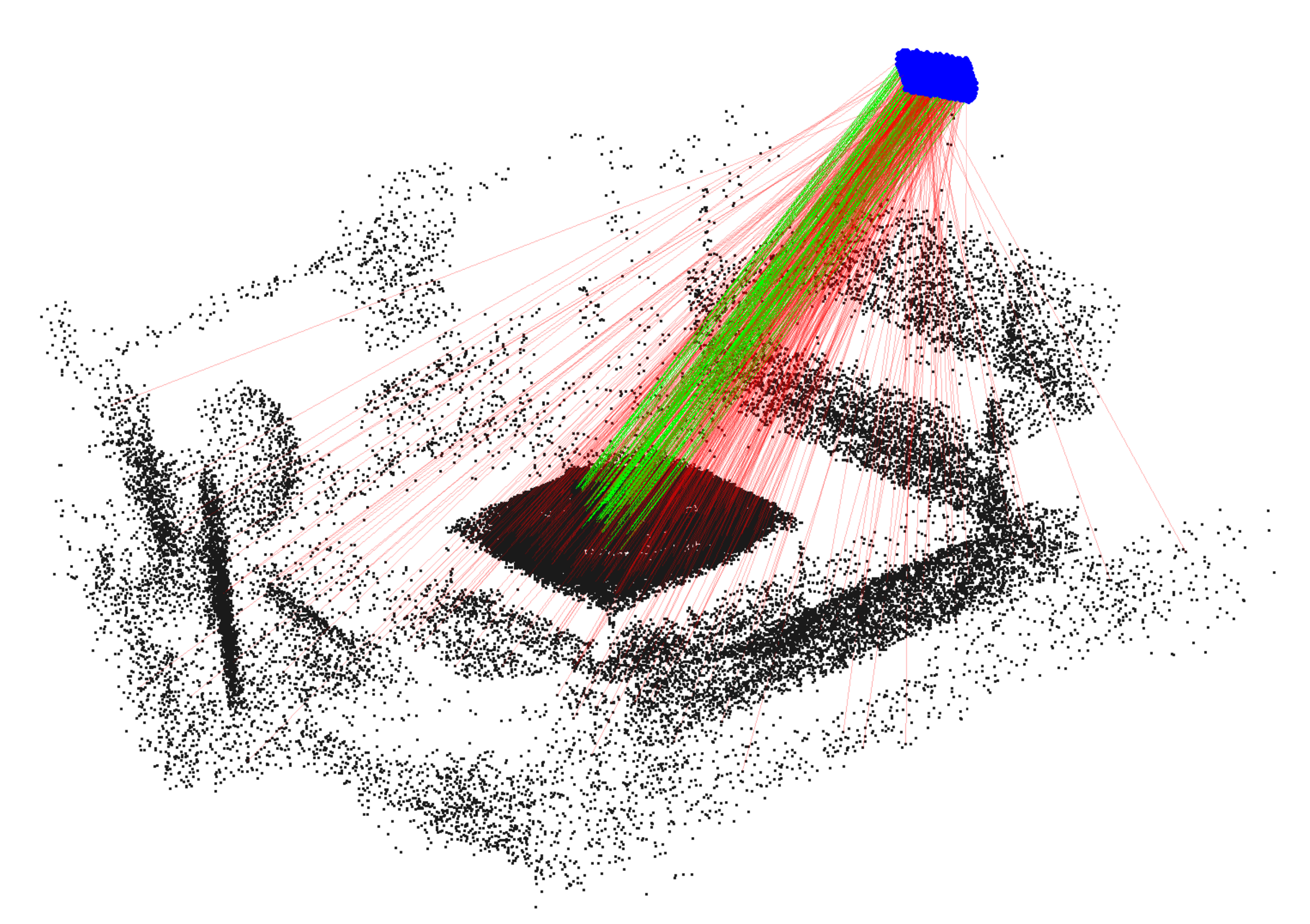}
\includegraphics[width=0.485\linewidth]{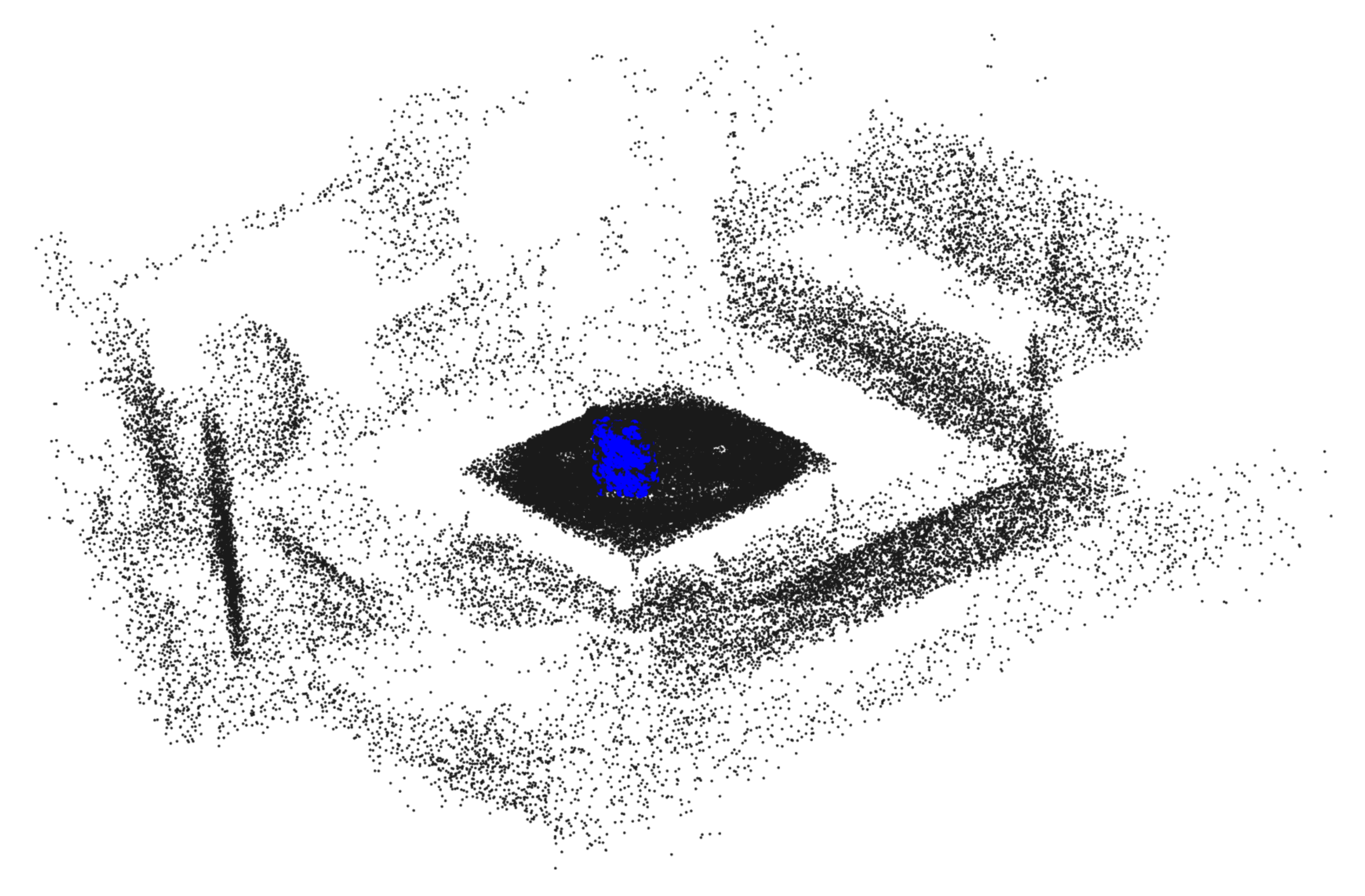}
\end{minipage}
\begin{minipage}[t]{0.49\linewidth}
\centering
\includegraphics[width=0.485\linewidth]{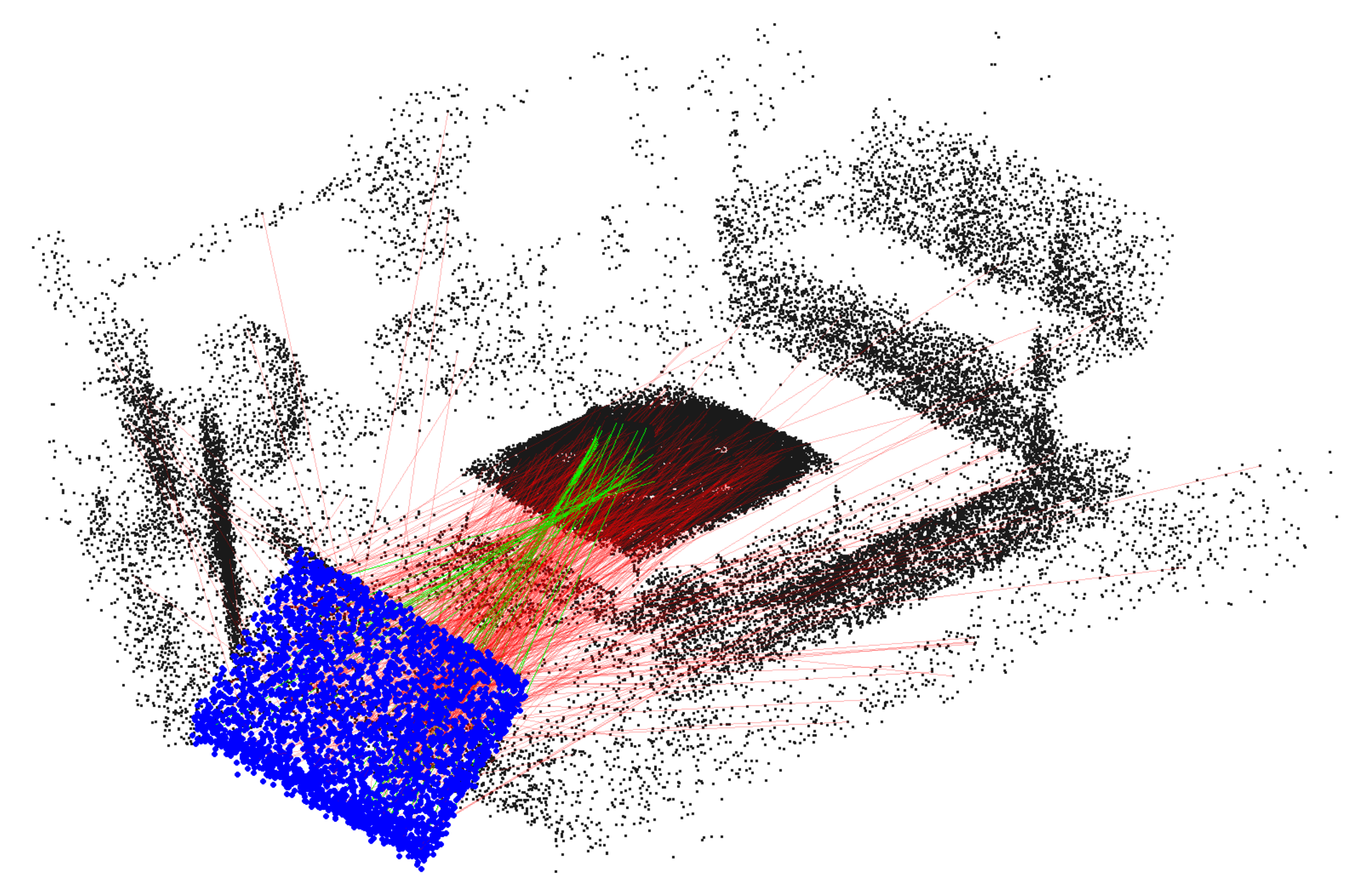}
\includegraphics[width=0.485\linewidth]{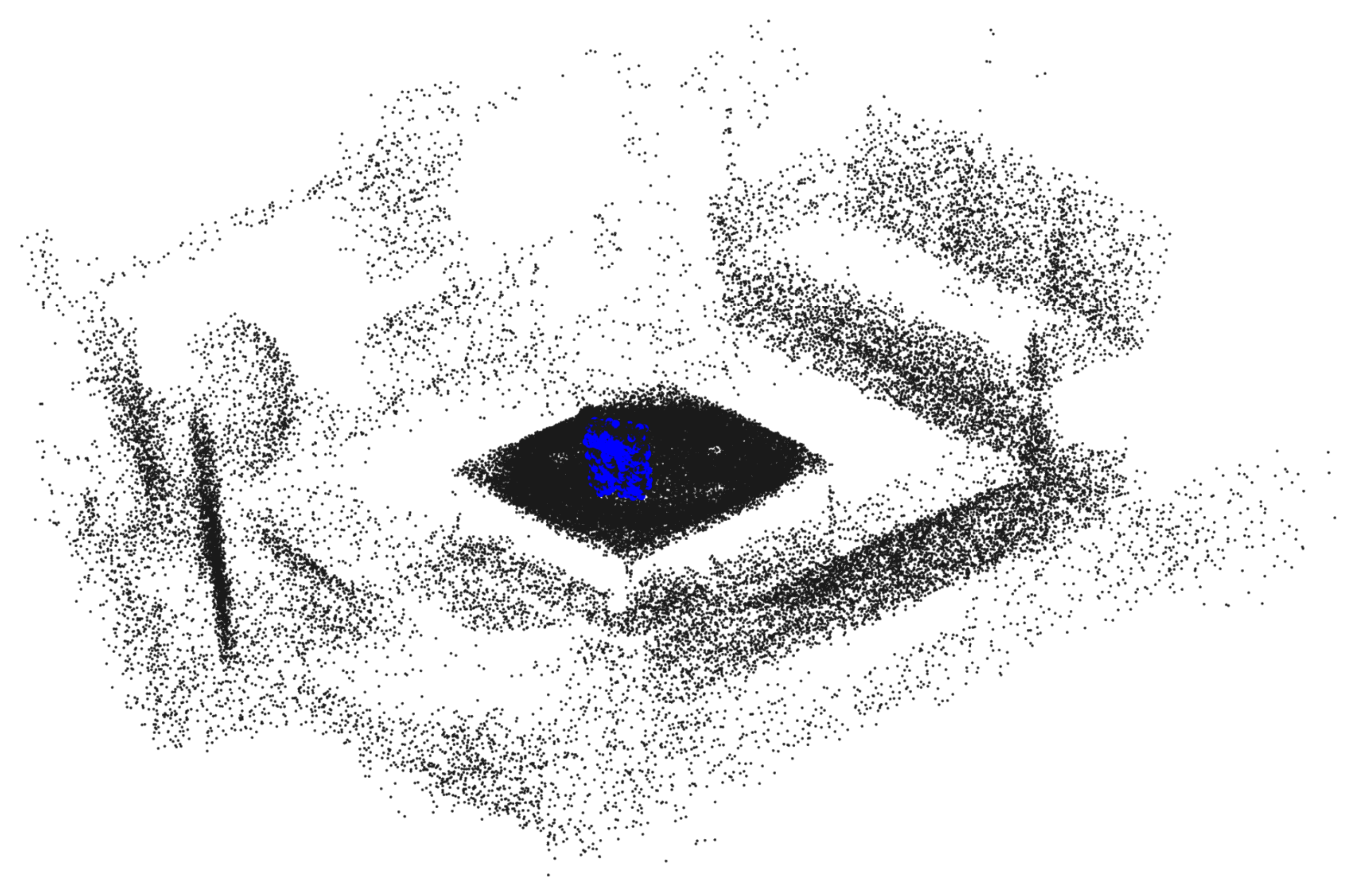}
\end{minipage}
}%

\footnotesize{(d) \textit{Object Localization: Scene-10}}

\subfigure{
\begin{minipage}[t]{0.49\linewidth}
\centering
\includegraphics[width=0.485\linewidth]{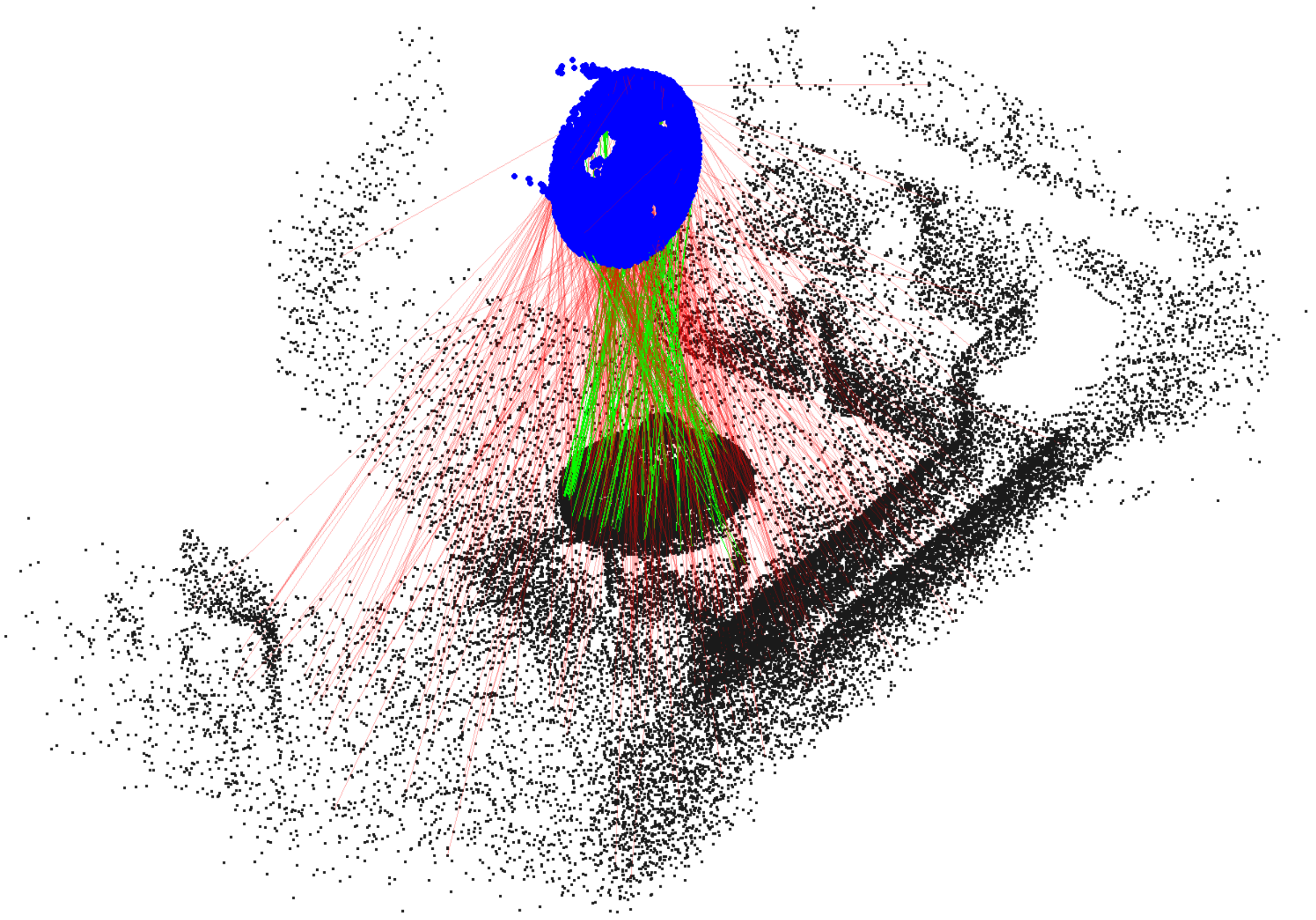}
\includegraphics[width=0.485\linewidth]{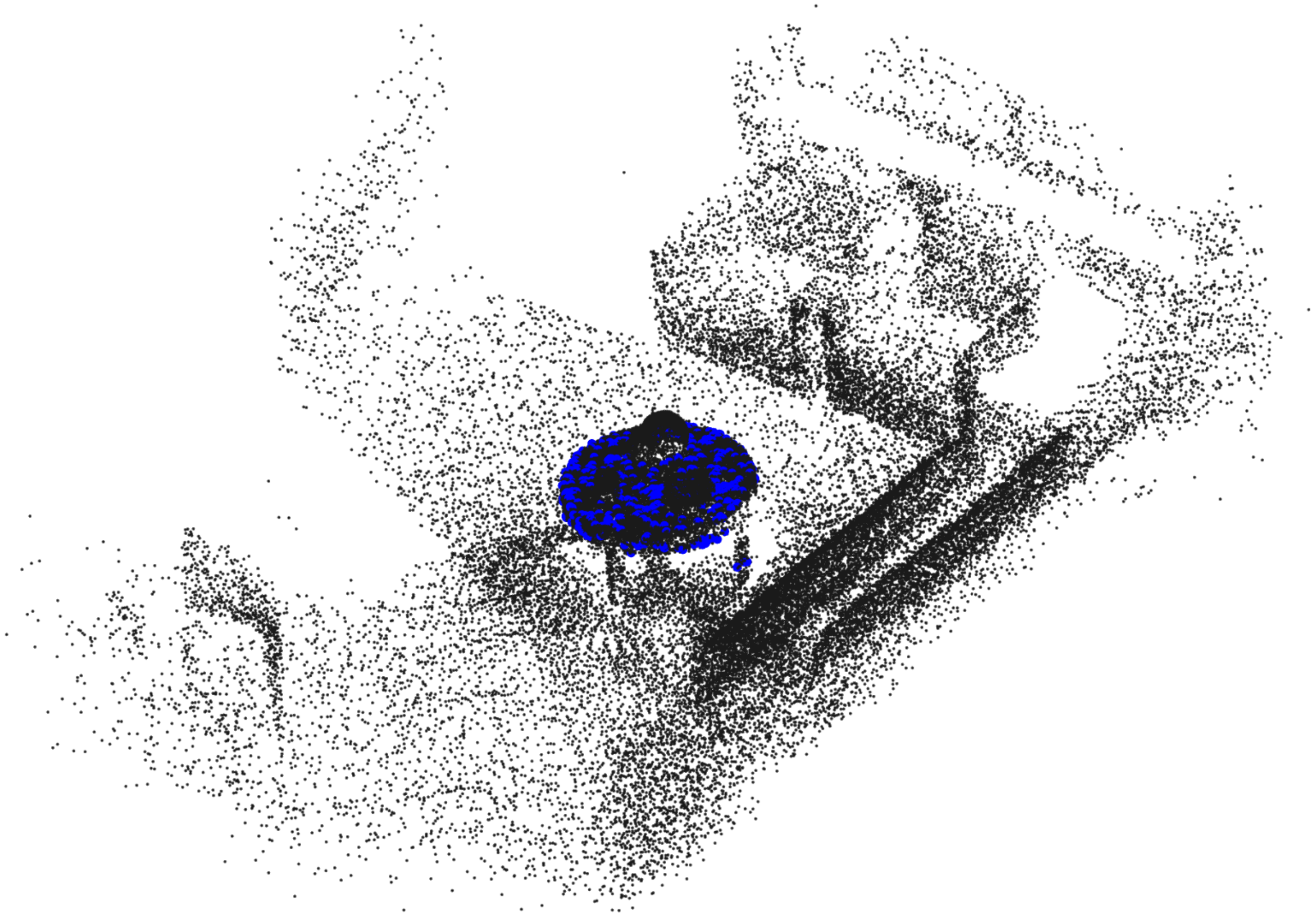}
\end{minipage}
\begin{minipage}[t]{0.49\linewidth}
\centering
\includegraphics[width=0.485\linewidth]{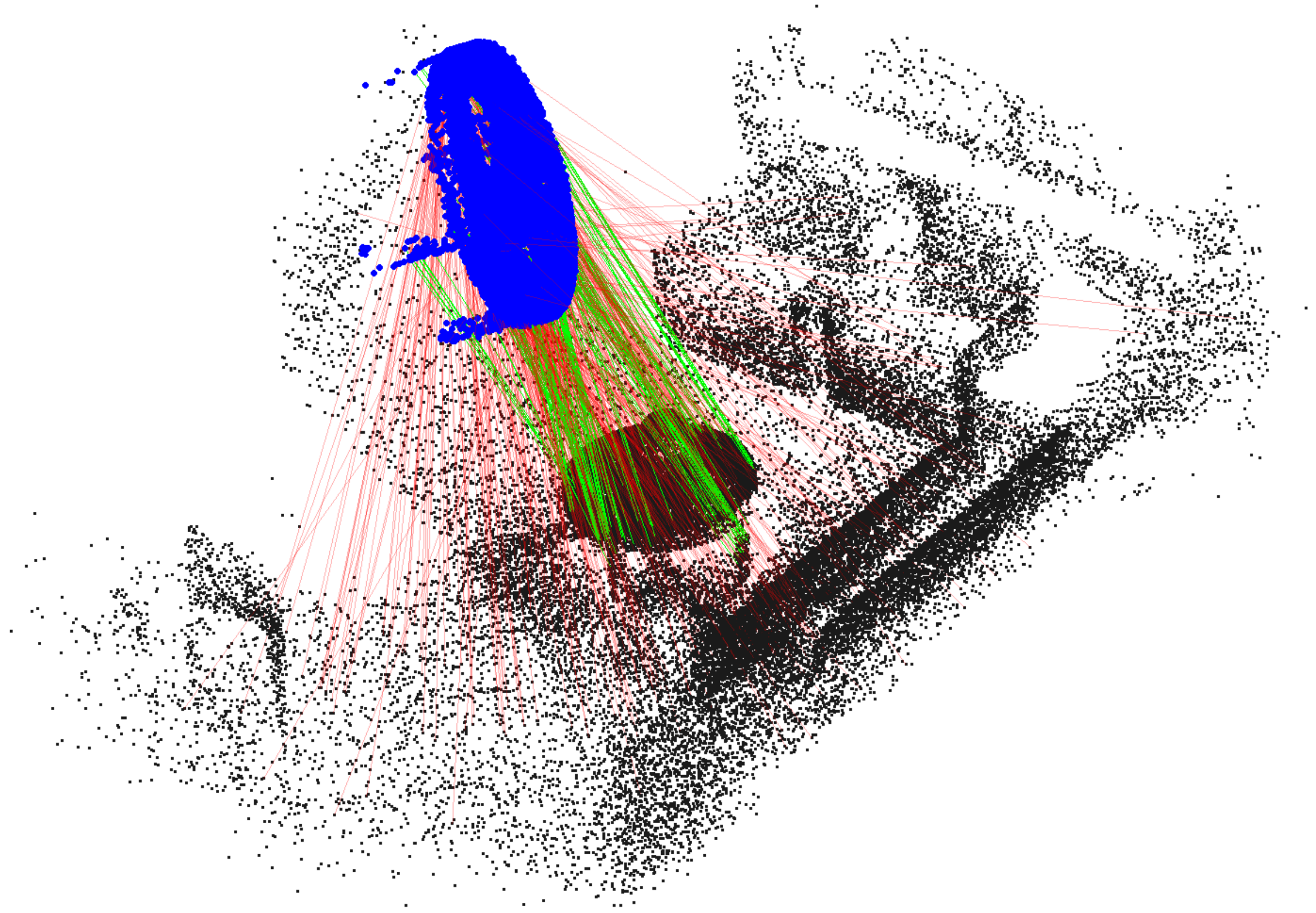}
\includegraphics[width=0.485\linewidth]{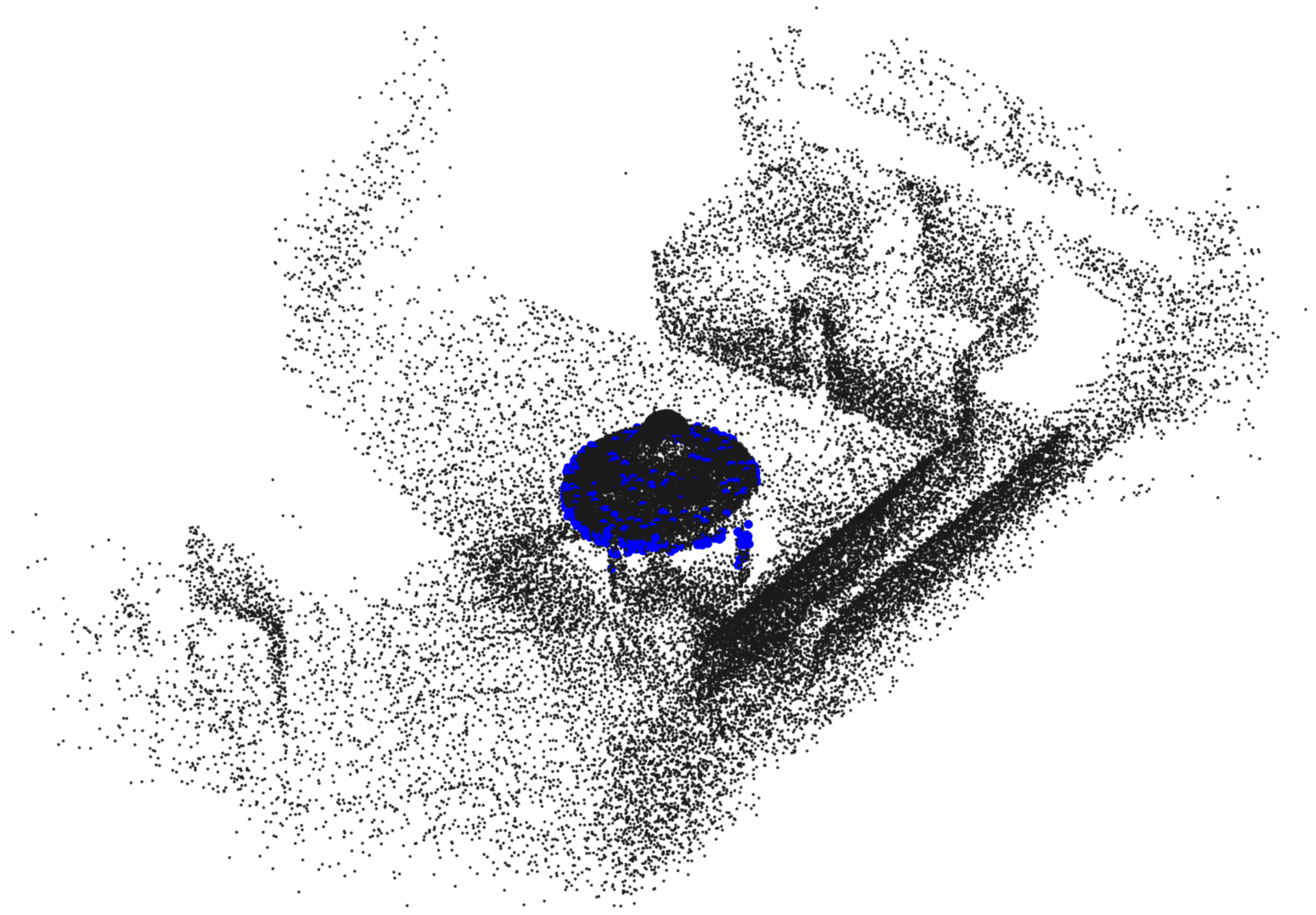}
\end{minipage}
}%

\vspace{-2mm}

\centering
\caption{Real experiments using RANSIC. (a-b) Image stitching. Setting is similar to Fig.~\ref{demo}(a). (c-d) Object localization. Column 1-2: Known-scale registration. Column 3-4: Unknown-scale registration. Column 1\&3: Inliers are in green lines and outliers are in red lines. Column 2\&4: The object is reprojected back to the scene with the transformation estimated by RANSIC.}
\label{Real-data}
\end{figure}

We test RANSIC for 3D object localization over the RGB-D scenes datasets~\cite{lai2011large}. We first extract the point cloud of the target object (\textit{cereal box, table}) from the scene using the labels and transform it with a random transformation. We then apply the FPFH feature descriptor~\cite{rusu2009fast} to construct correspondences between the scene and the transformed object (outlier ratio  always $>$75\%). We use Algorithm~\ref{Algo2-PCR} to estimate the transformation (in both unknown-scale and known-scale cases), and display the qualitative results in Fig.~\ref{Real-data}(c-d). The errors of ($\mathit{s}, \boldsymbol{R}, \boldsymbol{t}$) are all lower than ($0.001, 0.02^{\circ}, 0.004$).

\section{Conclusion}

We present a novel approach, named RANSIC, for the correspondence-based rotation search and point cloud registration problems by applying the compatibility of invariants to seek inliers from random samples. RANSIC has a well-designed mechanism that can automatically break from random sampling when a sufficient number of possible inliers are collected and the termination condition on residual errors is fulfilled. In the experiments, RANSIC is efficient in speed, robust against more than 95\% outliers (up to 99\% in some cases), accurate under noise disturbance, and also capable of recalling almost 100\% of the inliers, manifesting the most, or at least one of the most, state-of-the-art performance. Future efforts may focus on applying RANSIC to more problems.

\clearpage

\newpage






{\small


\begin{thebibliography}{10}\itemsep=-1pt

\bibitem{andrew2001multiple}
Alex~M Andrew.
\newblock Multiple view geometry in computer vision.
\newblock {\em Kybernetes}, 2001.

\bibitem{antonante2020outlier}
Pasquale Antonante, Vasileios Tzoumas, Heng Yang, and Luca Carlone.
\newblock Outlier-robust estimation: Hardness, minimally-tuned algorithms, and
  applications.
\newblock {\em arXiv preprint arXiv:2007.15109}, 2020.

\bibitem{arun1987least}
K~Somani Arun, Thomas~S Huang, and Steven~D Blostein.
\newblock Least-squares fitting of two 3-d point sets.
\newblock {\em IEEE Transactions on pattern analysis and machine intelligence},
  (5):698--700, 1987.

\bibitem{bailey2000data}
Tim Bailey, Eduardo~Mario Nebot, JK Rosenblatt, and Hugh~F Durrant-Whyte.
\newblock Data association for mobile robot navigation: A graph theoretic
  approach.
\newblock In {\em Proceedings 2000 ICRA. Millennium Conference. IEEE
  International Conference on Robotics and Automation. Symposia Proceedings
  (Cat. No. 00CH37065)}, volume~3, pages 2512--2517. IEEE, 2000.

\bibitem{barath2018graph}
Daniel Barath and Ji{\v{r}}{\'\i} Matas.
\newblock Graph-cut ransac.
\newblock In {\em Proceedings of the IEEE conference on computer vision and
  pattern recognition}, pages 6733--6741, 2018.

\bibitem{barfoot2017state}
Timothy~D Barfoot.
\newblock {\em State estimation for robotics}.
\newblock Cambridge University Press, 2017.

\bibitem{bay2006surf}
Herbert Bay, Tinne Tuytelaars, and Luc Van~Gool.
\newblock Surf: Speeded up robust features.
\newblock In {\em European conference on computer vision}, pages 404--417.
  Springer, 2006.

\bibitem{blais1995registering}
G{\'e}rard Blais and Martin~D. Levine.
\newblock Registering multiview range data to create 3d computer objects.
\newblock {\em IEEE Transactions on Pattern Analysis and Machine Intelligence},
  17(8):820--824, 1995.

\bibitem{bray2006posecut}
Matthieu Bray, Pushmeet Kohli, and Philip~HS Torr.
\newblock Posecut: Simultaneous segmentation and 3d pose estimation of humans
  using dynamic graph-cuts.
\newblock In {\em European conference on computer vision}, pages 642--655.
  Springer, 2006.

\bibitem{bustos2017guaranteed}
Alvaro~Parra Bustos and Tat-Jun Chin.
\newblock Guaranteed outlier removal for point cloud registration with
  correspondences.
\newblock {\em IEEE transactions on pattern analysis and machine intelligence},
  40(12):2868--2882, 2017.

\bibitem{chin2017maximum}
Tat-Jun Chin and David Suter.
\newblock The maximum consensus problem: recent algorithmic advances.
\newblock {\em Synthesis Lectures on Computer Vision}, 7(2):1--194, 2017.

\bibitem{choi2015robust}
Sungjoon Choi, Qian-Yi Zhou, and Vladlen Koltun.
\newblock Robust reconstruction of indoor scenes.
\newblock In {\em Proceedings of the IEEE Conference on Computer Vision and
  Pattern Recognition}, pages 5556--5565, 2015.

\bibitem{curless1996volumetric}
Brian Curless and Marc Levoy.
\newblock A volumetric method for building complex models from range images.
\newblock In {\em Proceedings of the 23rd annual conference on Computer
  graphics and interactive techniques}, pages 303--312, 1996.

\bibitem{dasari2014park}
Naga~Shailaja Dasari, Ranjan Desh, and Mohammad Zubair.
\newblock Park: An efficient algorithm for k-core decomposition on multicore
  processors.
\newblock In {\em 2014 IEEE International Conference on Big Data (Big Data)},
  pages 9--16. IEEE, 2014.

\bibitem{drost2010model}
Bertram Drost, Markus Ulrich, Nassir Navab, and Slobodan Ilic.
\newblock Model globally, match locally: Efficient and robust 3d object
  recognition.
\newblock In {\em 2010 IEEE computer society conference on computer vision and
  pattern recognition}, pages 998--1005. Ieee, 2010.

\bibitem{enqvist2009optimal}
Olof Enqvist, Klas Josephson, and Fredrik Kahl.
\newblock Optimal correspondences from pairwise constraints.
\newblock In {\em 2009 IEEE 12th international conference on computer vision},
  pages 1295--1302. IEEE, 2009.

\bibitem{eppstein2010listing}
David Eppstein, Maarten L{\"o}ffler, and Darren Strash.
\newblock Listing all maximal cliques in sparse graphs in near-optimal time.
\newblock In {\em International Symposium on Algorithms and Computation}, pages
  403--414. Springer, 2010.

\bibitem{fischler1981random}
Martin~A Fischler and Robert~C Bolles.
\newblock Random sample consensus: a paradigm for model fitting with
  applications to image analysis and automated cartography.
\newblock {\em Communications of the ACM}, 24(6):381--395, 1981.

\bibitem{guo20143d}
Yulan Guo, Mohammed Bennamoun, Ferdous Sohel, Min Lu, and Jianwei Wan.
\newblock 3d object recognition in cluttered scenes with local surface
  features: A survey.
\newblock {\em IEEE Transactions on Pattern Analysis and Machine Intelligence},
  36(11):2270--2287, 2014.

\bibitem{hartley2013rotation}
Richard Hartley, Jochen Trumpf, Yuchao Dai, and Hongdong Li.
\newblock Rotation averaging.
\newblock {\em International journal of computer vision}, 103(3):267--305,
  2013.

\bibitem{hartley2009global}
Richard~I Hartley and Fredrik Kahl.
\newblock Global optimization through rotation space search.
\newblock {\em International Journal of Computer Vision}, 82(1):64--79, 2009.

\bibitem{henry2012rgb}
Peter Henry, Michael Krainin, Evan Herbst, Xiaofeng Ren, and Dieter Fox.
\newblock Rgb-d mapping: Using kinect-style depth cameras for dense 3d modeling
  of indoor environments.
\newblock {\em The International Journal of Robotics Research}, 31(5):647--663,
  2012.

\bibitem{horn1987closed}
Berthold~KP Horn.
\newblock Closed-form solution of absolute orientation using unit quaternions.
\newblock {\em Josa a}, 4(4):629--642, 1987.

\bibitem{horn1988closed}
Berthold~KP Horn, Hugh~M Hilden, and Shahriar Negahdaripour.
\newblock Closed-form solution of absolute orientation using orthonormal
  matrices.
\newblock {\em JOSA A}, 5(7):1127--1135, 1988.

\bibitem{kummerle2011g}
Rainer K{\"u}mmerle, Giorgio Grisetti, Hauke Strasdat, Kurt Konolige, and
  Wolfram Burgard.
\newblock g 2 o: A general framework for graph optimization.
\newblock In {\em 2011 IEEE International Conference on Robotics and
  Automation}, pages 3607--3613. IEEE, 2011.

\bibitem{lai2011large}
Kevin Lai, Liefeng Bo, Xiaofeng Ren, and Dieter Fox.
\newblock A large-scale hierarchical multi-view rgb-d object dataset.
\newblock In {\em 2011 IEEE international conference on robotics and
  automation}, pages 1817--1824. IEEE, 2011.

\bibitem{li2021point}
Jiayuan Li, Qingwu Hu, and Mingyao Ai.
\newblock Point cloud registration based on one-point ransac and
  scale-annealing biweight estimation.
\newblock {\em IEEE Transactions on Geoscience and Remote Sensing}, 2021.

\bibitem{lowe2004distinctive}
David~G Lowe.
\newblock Distinctive image features from scale-invariant keypoints.
\newblock {\em International journal of computer vision}, 60(2):91--110, 2004.

\bibitem{markley1988attitude}
F~Landis Markley.
\newblock Attitude determination using vector observations and the singular
  value decomposition.
\newblock {\em Journal of the Astronautical Sciences}, 36(3):245--258, 1988.

\bibitem{meneghetti2015image}
Giulia Meneghetti, Martin Danelljan, Michael Felsberg, and Klas Nordberg.
\newblock Image alignment for panorama stitching in sparsely structured
  environments.
\newblock In {\em Scandinavian Conference on Image Analysis}, pages 428--439.
  Springer, 2015.

\bibitem{olsson2008branch}
Carl Olsson, Fredrik Kahl, and Magnus Oskarsson.
\newblock Branch-and-bound methods for euclidean registration problems.
\newblock {\em IEEE Transactions on Pattern Analysis and Machine Intelligence},
  31(5):783--794, 2008.

\bibitem{papazov2012rigid}
Chavdar Papazov, Sami Haddadin, Sven Parusel, Kai Krieger, and Darius Burschka.
\newblock Rigid 3d geometry matching for grasping of known objects in cluttered
  scenes.
\newblock {\em The International Journal of Robotics Research}, 31(4):538--553,
  2012.

\bibitem{parra2014fast}
Alvaro Parra~Bustos, Tat-Jun Chin, and David Suter.
\newblock Fast rotation search with stereographic projections for 3d
  registration.
\newblock In {\em Proceedings of the IEEE conference on computer vision and
  pattern recognition}, pages 3930--3937, 2014.

\bibitem{perera2012maximal}
Samunda Perera and Nick Barnes.
\newblock Maximal cliques based rigid body motion segmentation with a rgb-d
  camera.
\newblock In {\em Asian Conference on Computer Vision}, pages 120--133.
  Springer, 2012.

\bibitem{rusu2009fast}
Radu~Bogdan Rusu, Nico Blodow, and Michael Beetz.
\newblock Fast point feature histograms (fpfh) for 3d registration.
\newblock In {\em 2009 IEEE international conference on robotics and
  automation}, pages 3212--3217. IEEE, 2009.

\bibitem{rusu2008aligning}
Radu~Bogdan Rusu, Nico Blodow, Zoltan~Csaba Marton, and Michael Beetz.
\newblock Aligning point cloud views using persistent feature histograms.
\newblock In {\em 2008 IEEE/RSJ international conference on intelligent robots
  and systems}, pages 3384--3391. IEEE, 2008.

\bibitem{shi2020robin}
Jingnan Shi, Heng Yang, and Luca Carlone.
\newblock Robin: a graph-theoretic approach to reject outliers in robust
  estimation using invariants.
\newblock {\em arXiv preprint arXiv:2011.03659}, 2020.

\bibitem{sun2021iron}
Lei Sun.
\newblock Iron: Invariant-based highly robust point cloud registration.
\newblock {\em arXiv preprint arXiv:2103.04357}, 2021.

\bibitem{sunderhauf2012towards}
Niko S{\"u}nderhauf and Peter Protzel.
\newblock Towards a robust back-end for pose graph slam.
\newblock In {\em 2012 IEEE international conference on robotics and
  automation}, pages 1254--1261. IEEE, 2012.

\bibitem{wahba1965least}
Grace Wahba.
\newblock A least squares estimate of satellite attitude.
\newblock {\em SIAM review}, 7(3):409--409, 1965.

\bibitem{yang2020graduated}
Heng Yang, Pasquale Antonante, Vasileios Tzoumas, and Luca Carlone.
\newblock Graduated non-convexity for robust spatial perception: From
  non-minimal solvers to global outlier rejection.
\newblock {\em IEEE Robotics and Automation Letters}, 5(2):1127--1134, 2020.

\bibitem{yang2019polynomial}
Heng Yang and Luca Carlone.
\newblock A polynomial-time solution for robust registration with extreme
  outlier rates.
\newblock In {\em Robotics: Science and Systems}, 2019.

\bibitem{yang2019quaternion}
Heng Yang and Luca Carlone.
\newblock A quaternion-based certifiably optimal solution to the wahba problem
  with outliers.
\newblock In {\em Proceedings of the IEEE/CVF International Conference on
  Computer Vision}, pages 1665--1674, 2019.

\bibitem{yang2020teaser}
Heng Yang, Jingnan Shi, and Luca Carlone.
\newblock Teaser: Fast and certifiable point cloud registration.
\newblock {\em IEEE Transactions on Robotics}, 2020.

\bibitem{zhang2014loam}
Ji Zhang and Sanjiv Singh.
\newblock Loam: Lidar odometry and mapping in real-time.
\newblock In {\em Robotics: Science and Systems}, volume~2, 2014.

\bibitem{zhou2016fast}
Qian-Yi Zhou, Jaesik Park, and Vladlen Koltun.
\newblock Fast global registration.
\newblock In {\em European Conference on Computer Vision}, pages 766--782.
  Springer, 2016.

\end{thebibliography}
}

\end{document}